\documentclass{article} 
\usepackage{iclr2026_conference,times}

\usepackage{amsmath,amsfonts,bm}

\def\eqref#1{equation~\ref{#1}}

\def\1{\bm{1}}

\DeclareMathAlphabet{\mathsfit}{\encodingdefault}{\sfdefault}{m}{sl}
\SetMathAlphabet{\mathsfit}{bold}{\encodingdefault}{\sfdefault}{bx}{n}

\usepackage{hyperref}
\usepackage{url}
\usepackage{booktabs}
\usepackage{multirow}
\usepackage[table]{xcolor}
\usepackage{bbding}
\usepackage{subfig}
\usepackage{authblk}

\usepackage{setspace}
\usepackage{tcolorbox}
\usepackage{enumitem}
\setlist[itemize]{leftmargin=9pt}
\usepackage{subcaption}
\usepackage{float}
\usepackage{wrapfig}

\definecolor{deepred}{rgb}{0.631,0.102,0.102}
\definecolor{amethyst}{rgb}{0.6, 0.4, 0.8}
\definecolor{darkgreen}{rgb}{0.3,0.7,0.3}
\definecolor{salmon}{RGB}{241, 150, 141}
\definecolor{mildyellow}{HTML}{FFF2CC}
\newcommand{\finding}[2]{
\vspace{3pt}
\noindent\fcolorbox{deepred}{mildyellow}{\begin{minipage}{0.98\columnwidth}
    \textcolor{deepred}{\textbf{\textit{Finding} #1.} #2}
\end{minipage}}
\vspace{-3pt}
}

\newcommand{\model}{\text{VideoNSA}}
\newcommand{\pagehead}{\model: Native Sparse Attention Scales Video Understanding}
\renewcommand{\cite}{\citep}

\usepackage{tcolorbox}
\tcbuselibrary{skins,breakable}   
\usetikzlibrary{calc}

\definecolor{navy}{HTML}{182B49}
\definecolor{blue}{HTML}{00629B}
\definecolor{gold}{HTML}{C69214}
\definecolor{turquoise}{HTML}{00C6D7}
\definecolor{yellow}{HTML}{FFCD00}

\newcounter{researchquestion}
\newcommand{\researchquestion}[2][]{%
  \vspace{0.6em}\refstepcounter{researchquestion}%
  \begin{tcolorbox}[
    enhanced, colback=blue!10, colframe=navy,
    boxrule=1.5pt, arc=2mm, 
    left=7pt,right=7pt,top=6pt,bottom=2pt,
    overlay={
      \node[anchor=west, draw=none,
            fill=gold!20, text opacity=1,
            rounded corners=1mm,
            inner xsep=5pt, inner ysep=1pt,
            font=\bfseries\fontsize{12pt}{14pt}\selectfont, inner ysep=4pt, text=black] (rqtag)
        at ([xshift=1.0em,yshift=1.5pt]frame.north west)
        {Question \textcolor{black}{\theresearchquestion}};
      \fill[colback] ([yshift=.9pt]rqtag.south west)
                    rectangle
                    ([yshift=.9pt]rqtag.south east);
    }
  ]
  \normalsize #2
  \end{tcolorbox}%
  \ifx\\#1\\\else\label{rq:#1}\fi\vspace{0.4em}%
}

\author[1]{Enxin Song}
\author[2]{Wenhao Chai}
\author[3]{Shusheng Yang}
\author[1]{Ethan Armand}
\author[1]{Xiaojun Shan}
\author[1]{Haiyang Xu}
\author[4]{Jianwen Xie}
\author[1]{Zhuowen Tu}

\affil[ ]{$^1$University of California, San Diego \quad
          $^2$Princeton University \quad
          $^3$New York University \quad
          $^4$Lambda, Inc}

\title{\model: Native Sparse Attention Scales Video Understanding}

\iclrfinalcopy 
\begin{document}

\maketitle

\begin{abstract}

Video understanding in multimodal language models remains limited by context length: models often miss key transition frames and struggle to maintain coherence across long time scales. To address this, we adapt Native Sparse Attention (NSA) to video-language models.  \textbf{Our method, ~\model, adapts Qwen2.5-VL through end-to-end training on a 216K video instruction dataset. We employ a hardware-aware hybrid approach to attention, preserving dense attention for text, while employing NSA for video.} Compared to token-compression and training-free sparse baselines,~\model~achieves competitive performance on long-video understanding, temporal reasoning, and spatial benchmarks. Further ablation analysis reveals four key findings: (1) reliable scaling to 128K tokens; (2) an optimal global–local attention allocation at a fixed budget; (3) task-dependent branch usage patterns; and (4) the learnable combined sparse attention help induce dynamic attention sinks.

\vspace{2mm}
\textbf{Project page:} \href{https://enxinsong.com/VideoNSA-web/}{https://enxinsong.com/VideoNSA-web/}

\textbf{Code:} \href{https://github.com/Espere-1119-Song/VideoNSA}{https://github.com/Espere-1119-Song/VideoNSA}

\textbf{Model:} \href{https://huggingface.co/Enxin/VideoNSA}{https://huggingface.co/Enxin/VideoNSA}

\end{abstract}
\section{Introduction}

Key moments of a video can occur at any time, exemplified by soccer where game deciding moments typically span seconds of a 90 minute game. Within those game deciding moments split second actions define the outcome: an assist, a missed tackle, the movement of the keeper. Multimodal large language models (MLLMs)\cite{kwaikeyeteam2025kwaikeyevltechnicalreport,team2025kimi,coreteam2025mimovltechnicalreport,qiu2025intentvcnet,yuan2025datedynamicabsolutetime,Gui2025CharacterCentricUO,Chung2025VideoPS,Li2025VideoProAP} have achieved substantial progress in vision-language perception and reasoning,  but still cannot match humans ability to extract and reason about salient moments in videos. While humans naturally sample color visuals around 60hz, \citep{Kalloniatis2007TemporalResolution} across large contexts, existing VLMs  often sample a single frame per second. 
Intuitively, increasing the context for these models by sampling more frames improves accuracy~\cite{cai2024temporalbench,wu2024longvideobench}, particularly for long videos and complex reasoning tasks. However, this approach pays for improvement with additional tokens, increasing computational complexity and pushing against fundamental limits of model context.

To address these challenges, many approaches~\cite{wang2024retake,li2024videochat,jin2024chat,wang2025internvideo2,yang2024pvc} adopt token compression to reduce redundancy and increase informative context. However, when applied to complex reasoning tasks, these compression-based models perform worse compared to full-token methods~\cite{song2025video}. Moreover, compression strategies often limit generalization through reduced perception and reasoning capacity~\cite{wen2025token}. In contrast, sparse attention mechanisms preserve tokens, but focus the models capabilities on relevant dependencies between tokens. Numerous sparse attention methods have already been employed in large language models (LLMs), but most are inadequate for video complexity (detailed in Appendix~\ref{related}). Therefore, we present VideoNSA, which adopts Native Sparse Attention~\cite{yuan2025native}, a learnable hardware-aware sparse attention mechanism proven to be effective in long-context modeling. VideoNSA is the first learnable and hardware-aware sparse attention framework tailored for video understanding, effectively scaling to ultra-long vision-text context. We apply the learnable sparse attention to video token sequences, while preserving grouped-query attention for text tokens. Following this pattern, our experiments show that using only 3.6\% of the attention budget on 128K context length while improving performance on various tasks

We further conduct massive experiments and analyses of~\model~, revealing several important findings: (1) ~\model~ extrapolates effectively to contexts beyond its training length, and the optimal balance between temporal density and spatial resolution is highly task dependent. (2)~\model~is also sensitive by attention scaling, with results remaining strongest near the training configuration. (3) The gating distribution evolves dynamically across layers, and the selection and sliding-window branches gradually lose importance in deeper layers. (4) The compression branch emerges as the main computational bottleneck. (5)Moreover, the learned sparse attention weights remain beneficial even under dense attention settings. (6) Learnable sparse attention induces distinctive attention sink behaviors across branches, with very few sinks in the selection branch and periodic sink formation in the compression branch.



In particular, our paper makes the following contributions:
\begin{itemize}
    \item We propose~\model, a hardware-aware native sparse attention mechanism, and systematically investigate its effectiveness for video understanding, scaling up to a 128K vision context length.
    \item We introduce hybrid sparse attention in~\model, enabling flexible allocation of information and attention budgets to achieve optimal performance across diverse task.
    \item We dynamically combine global and local attention through three complementary branches, which effectively reduce attention sinks in long vision contexts.
\end{itemize}

\section{\model}
\subsection{Preliminaries}

\paragraph{Native sparse attention. }
Existing training-free sparse attention methods are rarely hardware aligned, and typically don't increase training efficiency.
Native Sparse Attention~\cite{yuan2025native} (NSA) avoids computing attention between all key-value pairs $(\mathbf{K}_t, \mathbf{V}_t)$, instead, for each query $\mathbf{q}_t$, NSA dynamically constructs an information-dense KV cache subset.  NSA combines three complementary cache branches with a learnable gate $g_t^c$ adaptively weighting each branch yielding $\mathbf{o}_t$:
\begin{equation}
\mathbf{o}_t=\sum_{c\in\{\mathrm{cmp},\,\mathrm{slc},\,\mathrm{win}\}} 
g_t^c \cdot \mathrm{Attn}\!\big(q_t,\tilde{\mathbf{K}}^{\,c}_t,\tilde{\mathbf{V}}^{\,c}_t\big).
\end{equation}

    \noindent\text{Token Compression (CMP)} branch aggregates sequential blocks of keys into more coarse-grained, single block-level representations $\tilde{\mathbf{K}}^{\text{cmp}}_t$ via a learnable MLP $\varphi$:
    \begin{equation}
        \tilde{\mathbf{K}}^{\text{cmp}}_t = \{ \varphi(\mathbf{K}_{[id+1:id+m]}) \mid 0 \le i < \lfloor\frac{t-m}{d}\rfloor \},
    \end{equation}
    where $m$ is the block length, $d$ is the stride.

    \noindent\text{Token Selection (SLC)} branch preserves the most salient key-value blocks by computing importance scores $p^{\text{slc'}}_t$ and selecting the indices of the top-n blocks:
    \begin{equation}
        I_t = \{i \mid \text{rank}(p^{\text{slc'}}_t[i]) \le n \}.
    \end{equation}
    The final set of selected keys is formed by concatenating these top-ranked blocks:
    \begin{equation}
        \tilde{\mathbf{K}}^{\text{slc}}_t = \text{Cat}(\{ \mathbf{K}_{[im'+1:(i+1)m']} \mid i \in I_t \}),
    \end{equation}
    where $I_t$ is the set of selected indices, $n$ is the number of blocks to retain.

    \noindent\text{Sliding Window (SWA)} branch simply applies the standard sliding window attention, which retains the fixed $w$ most recent key-value pairs:
    \begin{equation}
        \tilde{\mathbf{K}}^{\text{swa}}_t = \mathbf{K}_{t-w+1:t}, \quad \tilde{\mathbf{V}}^{\text{swa}}_t = \mathbf{V}_{t-w+1:t}.
    \end{equation}

\paragraph{Grouped query attention. } 
In Multi-Head Attention (MHA), each query head has dedicated key--value (KV) projections, which makes the KV cache scale with the number of heads and increases inference cost. 
Grouped-Query Attention (GQA)~\cite{ainslie2023gqa} mitigates this by letting multiple query heads share fewer KV heads. 
For each input $\{x_i\}_{i=1}^{L}$, GQA partitions the $h$ query heads into $g$ groups $(1\le g\le h)$. At a given timestep $t$, the output $o_t^{(s)}$ for the $s$-th query head with group index $m(s)= \lceil sg/h \rceil$ is computed by applying attention to the shared keys and values as:
\begin{align}
    o_t^{(s)} &= \text{Attention}(q_t^{(s)}, K_{\le t}^{(m(s))}, V_{\le t}^{(m(s))}) = \text{softmax}\left(\frac{(q_t^{(s)})^{\top} K_{\le t}^{(m(s))}}{\sqrt{d_k}}\right) V_{\le t}^{(m(s))},
\end{align}
where $q_i^{(s)} = x_i W_q^{(s)}$, $k_i^{(m(s))} = x_i W_k^{(m(s))}$, $v_i^{(m(s))} = x_i W_v^{(m(s))}$. The outputs $o_t$ from all heads are concatenated by $o_t = [o_t^{(1)}, o_t^{(2)}, \dots, o_t^{(h)}]$.
\model~utilizes Qwen2.5-VL-7B~\cite{bai2025qwen2} as the backbone, with Qwen2.5-7B~\cite{qwen2025qwen25technicalreport} as the LLM decoder, which employs GQA for efficient KV cache utilization using 28 query heads and 4 shared key/value heads.

\begin{figure}[t]
    \centering
    \includegraphics[width=\linewidth]{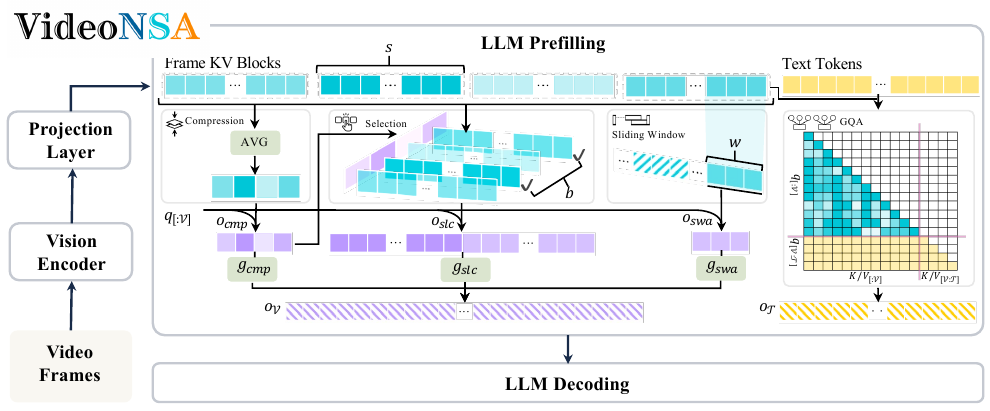}
    \caption{ Overview of~\model. Video frames are encoded into frame-level KV blocks.~\model~utilizes three sparse attention branches during prefilling stage: \textbf{compression branch} reduces redundancy via token averaging, \textbf{selection branch} identifies top-k important tokens, and \textbf{sliding window branch} enforces local temporal coverage. The outputs are combined through dynamic gating before integration with text tokens for LLM decoding.}
    \label{fig:framework}
\end{figure}

\subsection{Architecture}


Existing token compression methods~\cite{yang2025vflowopt,zhang2025infinite,hyun2025multi,zhang2025flexselect} suffer from irreversible information loss on complex tasks and don't address computational and latency bottlenecks in LLM video understanding. From the perspective of attention as a message passing in a  Graph Neural Network~\cite{joshi2025transformers, pappone2025attentionsinks}, it's clear this bottleneck is fundamental. Standard attention propagates information between nodes (tokens) through edges (attention weights), with each token being updated by aggregating features from its neighbors, weighted by attention scores.
Training-free sparse attention often imposes a static adjacency matrix whose fixed subgraph connectivity restricts information flow. Conversely, NSA~\cite{yuan2025native} provides data-dependent sparsity that preserves edges necessary for a particular task.

We build~\model~upon Qwen2.5-VL-7B~\cite{qwen2025qwen25technicalreport}, which incorporates a vision encoder and adopts Qwen2.5-7B~\cite{bai2025qwen2} as the LLM. As illustrated in Figure~\ref{fig:framework},~\model~introduces a hybrid attention mechanism in the LLM across different modalities. At each layer $l$, we split the input tokens $\mathbf{X}^{(l-1)}$ into vision tokens $\mathbf{X}^{(l-1)}_{\mathcal V}$ and text tokens $\mathbf{X}^{(l-1)}_{\mathcal T}$ according to their position IDs. 
For vision tokens,~\model~applies NSA~\cite{yuan2025native} with a dedicated gate $g^{c}_{t}$ on each head. We set the block size $s$ equal to the token number per frame, and obtain the block-level representation by averaging all tokens within the block. The vision attention output $o_{\mathcal{V}}$ is dynamically weighted by the compression, selection, and sliding window branches as:
\[
\mathbf{o}^{(l)}_{\mathcal V} = 
\sum_{c\in\{\text{cmp}, \text{slc}, \text{win}\}} g^{c}_{t}\, 
\text{Attn}\!\big(q_t, \tilde{\mathbf{K}}^{c}_{t}, \tilde{\mathbf{V}}^{c}_{t}\big),
\]
where $g^{c}_{t}$ is implemented as a two-layer MLP with a sigmoid activation. 

The text attention output $\mathbf{o}^{(l)}_{\mathcal T}$ is computed using standard GQA~\cite{ainslie2023gqa} to preserve instruction following capabilities.
We obtain the final output $\mathbf{o}^{(l)}$ of the layer $l$ by concatenating:
\[
\mathbf{o}^{(l)} = [\,\mathbf{o}^{(l)}_{\mathcal V};\, \mathbf{o}^{(l)}_{\mathcal T}\,].
\]


\subsection{Training Recipe}

%

We conduct end-to-end training to adapt vision features for data-dependent sparse connectivity in the language model. The training dataset of~\model~is constructed from LLaVA-Video-178K~\cite{zhang2024video} by filtering for  question answer pairs at 4 fps and retaining videos with 350–550 frames, for a subset of 216K pairs. To emphasize sparse attention for temporal redundancy, we constrain the maximum pixels per frame to 50,176, and the maximum context length per training instance to 36K tokens. In~\model, block size $s$ is set to 64, block  $b$ is set to 32, and sliding window size $w$ is set to 256. We trained using SWIFT~\cite{zhao2024swiftascalablelightweightinfrastructure}, adapting the NSA~\cite{yuan2025native} implementation from FLA~\cite{yang2024fla} and~\cite{pai2025sparsity}. The complete training process requires 4600 H100 GPU hours.
More training details including hyper-parameters selection can be found in Appendix~\ref{train_detail}.


\section{Experiments}

\subsection{Effectiveness on Video Understanding}

\paragraph{Baselines}
Our primary baseline is Qwen2.5-VL-7B~\cite{qwen2025qwen25technicalreport} with dense FlashAttention~\cite{dao2023flashattention}. We compare~\model~against several strong baselines, including the quantization model AWQ~\cite{qwen2.5-vl-7b-instruct-awq}, training-free token compression models~\cite{yang2025visionzip,zhang2025vscan,chen2024image}, and training-free sparse attention methods~\cite{jiang2024minference,xu2025xattention,lai2025flexprefill,li2024scbench}.
All methods employ their official configuration without additional training and using Qwen2.5-VL-7B~\cite{qwen2025qwen25technicalreport} as a base. For token compression baselines, we use the token kept ratio and sampling fps from the original papers that yield the best accuracy, while for sparse attention baselines, we use the same configuration as~\model. 
In addition, we fine-tune Qwen2.5-VL-7B~\cite{qwen2025qwen25technicalreport} using the same training dataset as~\model~to serve as a competitive baseline. We also include models with different backbones for a broad comparison.

\definecolor{sand}{HTML}{f5f0E6}   
\begin{table*}[t]
\renewcommand{\arraystretch}{1.05}
\centering
\caption{Results on long video understanding, temporal reasoning and spatial understanding tasks. LVB, LTS for LongVideobench~\cite{wu2024longvideobench} and LongTimeScope~\cite{zohar2025apollo2}.} 
\addtolength\tabcolsep{-2.4pt} 
\resizebox{1\linewidth}{!}{
\begin{tabular}{lcccccc}
\toprule
\multicolumn{1}{c}{\multirow{2}{*}{Model}} & \multicolumn{4}{c}{Long-form Video}  & \multicolumn{1}{c}{Temporal} & \multicolumn{1}{c}{Spatial} \\
\cmidrule(lr){2-5} \cmidrule(lr){6-6} \cmidrule(lr){7-7}
\multicolumn{1}{c}{} & ~~~LVB~~~ & ~~MLVU$_{test}$ & TimeScope & ~~~LTS~~~ & Tomato & VSIBench \\ 
\midrule
LLaVA-OneVision-7B~\cite{li2024llava} & 56.3 & -- & -- & -- & \underline{25.5} & 32.4 \\
LLaVA-Video-7B~\cite{zhang2024llavanextvideo} &  58.2 & -- & 74.1 & 34.0 & -- & 35.6\\
VideoLLaMA3-8B~\cite{zhang2025videollama} & 59.8 & 47.7 & 69.5 & -- & -- & -- \\
InternVL2.5-8B~\cite{chen2024expanding} &  \underline{60.0} & -- & 55.8 & -- & -- & --\\
Video-XL-2~\cite{qin2025video} & \textbf{61.0} & \textbf{52.2} & -- & -- & -- & --\\
\midrule
Qwen2.5-VL-7B~\cite{qwen2025qwen25technicalreport} & 58.7 & 51.2  & 81.0  & 40.7 & 22.6  & 29.7 \\
Qwen2.5-VL-7B-AWQ~\cite{qwen2.5-vl-7b-instruct-awq} & 59.0 & 46.0 & --    & --   & --    & 35.0 \\
Qwen2.5-VL-7B-SFT & 57.8 & 51.2  & 76.8  & 40.2 & 21.7  & 30.5  \\
\multicolumn{7}{l}{\textit{Token Compression Methods}} \\ 
+ FastV~\cite{chen2024image} & 57.3 & 41.8 & 46.5  & 35.6 & 21.6  & 32.0 \\
+ VScan~\cite{zhang2025vscan} & 58.7 & 48.1 & 80.3  & 31.1 & 19.1  & 34.4 \\
+ VisionZip~\cite{yang2025visionzip} & 52.4 & 33.1 & 43.5  & 40.4 & 23.6 & 32.1 \\
\multicolumn{7}{l}{\textit{Sparse Attention Methods}} \\ 
+ Tri-Shape~\cite{li2024scbench} & 59.5 & 49.2  & 82.7  & 28.4 & 22.1  & 34.9  \\
+ MInference~\cite{jiang2024minference} & 59.2 & 49.2  & 82.7  & \textbf{44.4} & 23.0  & 36.5 \\
+ FlexPrefill~\cite{lai2025flexprefill} & 58.4 & 46.0  & 83.0  & 39.1 & 23.7  & 34.0 \\
+ XAttention~\cite{xu2025xattention}& 59.1 & 50.2  & \underline{83.1}  & \underline{41.1} & 21.4  & \textbf{36.6} \\
\midrule
\textbf{\model} & \underline{60.0} & \underline{51.8} & \textbf{83.7} & \textbf{44.4} & \textbf{26.5} & \underline{36.1} \\ 
\bottomrule
\end{tabular}
}

\label{tab:baseline} 
\end{table*}

We evaluate~\model~across three domains including \textbf{long video understanding}, \textbf{temporal reasoning}, and \textbf{spatial understanding} using LMMs-Eval~\cite{zhang2024lmms} and VLMEvalKit~\cite{duan2024vlmevalkit}. Table~\ref{tab:baseline} indicates that sparse attention methods consistently outperform token compression approaches.
We empirically evaluate the effectiveness of~\model~based on several popular long video understanding benchmarks, including \textbf{LongVideoBench}~\cite{wu2024longvideobench}, \textbf{MLVU}~\cite{zhou2024mlvu}, \textbf{TimeScope}~\cite{zohar2025apollo2} and \textbf{LongTimeScope}~\cite{zohar2025apollo2}.~\model~achieves competitive results, narrowing the gap with state-of-the-art methods. We observe that~\model~shows clear advantages on tasks involving order-sensitive temporal reasoning and ultra-long video settings (\textbf{10 hours} in LongTimeScope~\cite{zohar2025apollo2}).
To evaluate the visual temporal reasoning capbility of~\model, we evaluate~\model~on \textbf{Tomato}~\cite{shangguan2024tomato}, a benchmark spanning six reasoning types and three video scenarios.~\model~attains the highest accuracy on Tomato~\cite{shangguan2024tomato}, substantially outperforming compression-based methods, underscoring their limitations in fine-grained temporal inference. 
\textbf{VSIBench}~\cite{yang2025thinkingspacemultimodallarge} focuses on spatial reasoning allowing us to test whether efficient models can preserve local fidelity while achieving efficiency. ~\model~matches the strongest sparse attention baselines and significantly surpasses token compression methods in spatial understanding, confirming that it preserves spatial fidelity. All detailed evaluation settings and subset results can be found in Appendix~\ref{sec:evaluation_datasets_and_settings}, Appendix~\ref{base_long_results}, Appendix~\ref{base_reasoning_results}, and Appendix~\ref{base_spatial_results}.

\subsection{Ablation Study}

To further analyze the components of~\model, we visualize attention pattern in each branch in Appendix~\ref{vis_attn_branch} and assess the effectiveness of different branches.
Table~\ref{tab:ablation} shows that single-branch models suffer significant degradation, and even two-branch combinations remain inferior to the full~\model, highlighting the necessity of integrating all three branches with dynamic gating. Detailed results of different branch combination can be found in Appendix~\ref{branch_comb}.

\definecolor{sand}{HTML}{f5f0E6}   
\definecolor{blue}{HTML}{00629B}   

\begin{table*}[t]
\centering
\caption{Ablation study on branch selection across different tasks. LVB, LTS for LongVideobench~\cite{wu2024longvideobench} and LongTimeScope~\cite{zohar2025apollo2}.} 
\renewcommand{\arraystretch}{0.95}
\addtolength\tabcolsep{-2.4pt} 
\resizebox{1\linewidth}{!}{
\begin{tabular}{ccccccccc}
\toprule
\multicolumn{3}{c}{Branch} & \multicolumn{4}{c}{Long Video Understanding}  & \multicolumn{1}{c}{Temporal Reasoning} & \multicolumn{1}{c}{Spatial Understanding} \\
\cmidrule(lr){1-3} \cmidrule(lr){4-7} \cmidrule(lr){8-8} \cmidrule(lr){9-9}
\multicolumn{1}{c}{CMP} & \multicolumn{1}{c}{SLC} & \multicolumn{1}{c}{SWD} & LVB & MLVU$_{test}$ & TimeScope & LTS & Tomato & VSIBench \\ 

\midrule

\Checkmark & & & 48.1 & 43.9  & 41.5 & 25.1 & 23.3  & 29.2 \\
 & \Checkmark & & 48.4 & \underline{47.7}  & \underline{63.7} & \underline{37.1} & 24.0  & 27.6 \\
 & & \Checkmark & 49.1 & 40.2  & 59.3 & 29.8 & 24.0  & 29.8\\
 \midrule
\Checkmark & \Checkmark& & \underline{49.4} & 42.7  & 57.3  & 32.4 & 23.5  & 29.4 \\
\Checkmark & & \Checkmark& 49.3 & 42.4  & 65.2  & 34.4 & 23.0  & 29.1 \\
 & \Checkmark& \Checkmark& 48.8 & 43.4  & 57.3  & 31.6 & \underline{24.5}  & \underline{30.3} \\
 \midrule
\Checkmark & \Checkmark& \Checkmark  & \textbf{60.0} & \textbf{51.8} & \textbf{83.7} & \textbf{44.4} & \textbf{26.5} & \textbf{36.1}  \\

\bottomrule
\end{tabular}}

\vspace{-4mm}
\label{tab:ablation} 
\end{table*}

\begin{figure}[!h]
    \captionsetup{skip=2pt}
    \centering
    \begin{tabular}{ccc}
         \subfloat[Information Scaling of LongVideoBench]{
             \includegraphics[width=0.48\textwidth]{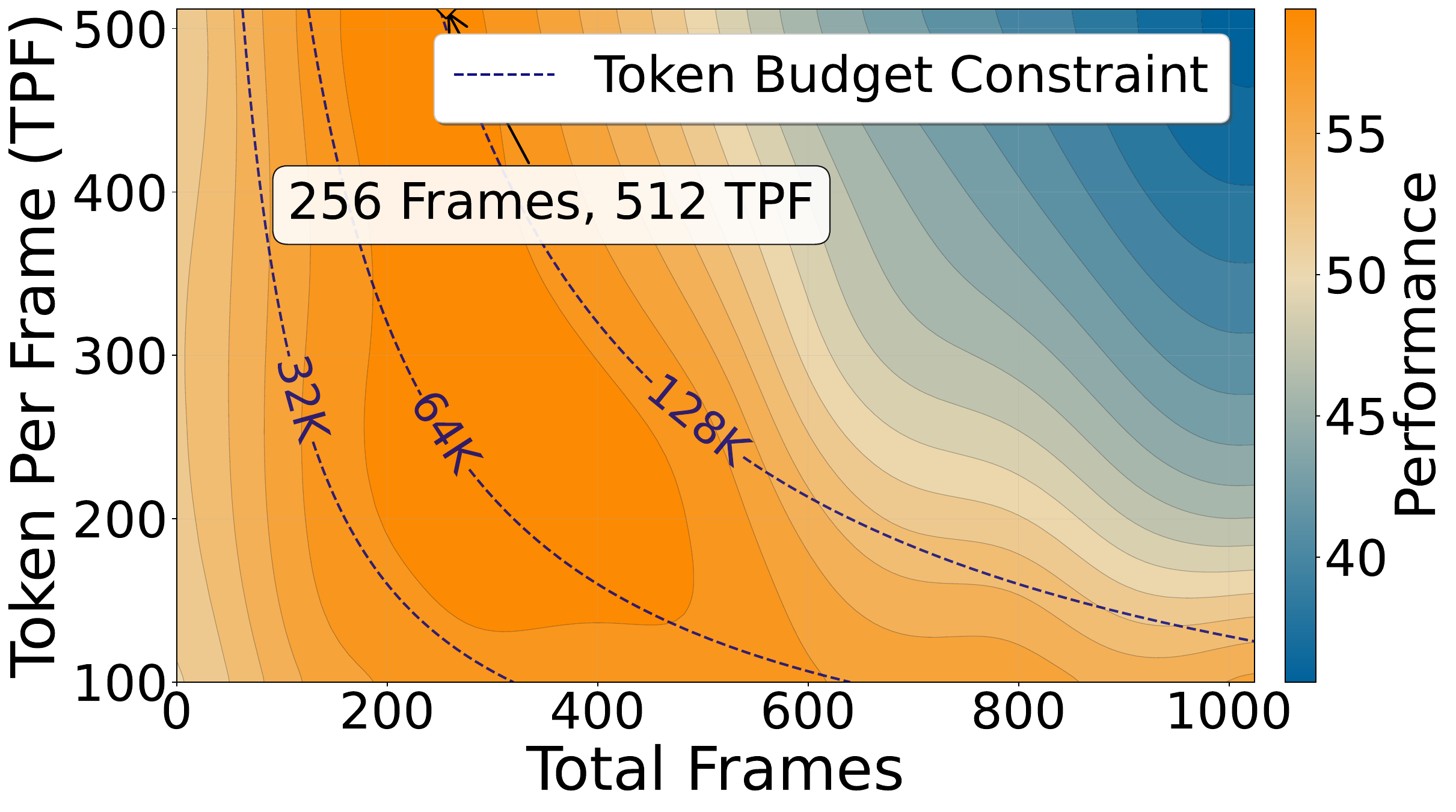}
         } &
         \subfloat[Information Scaling of TimeScope]{\includegraphics[width=0.48\textwidth]{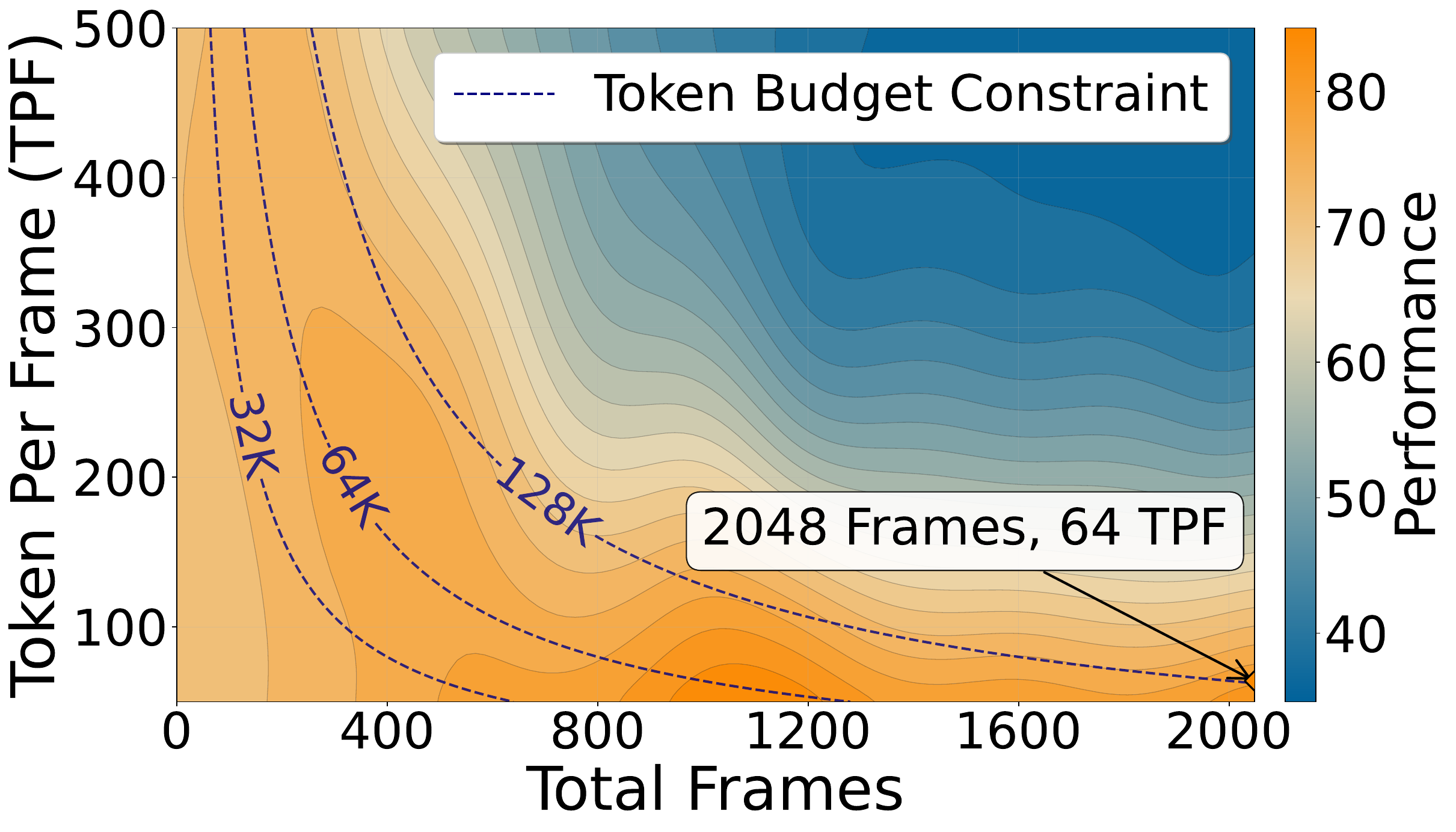}} 
         \\
         \subfloat[Information Scaling of Tomato]{\includegraphics[width=0.48\textwidth]{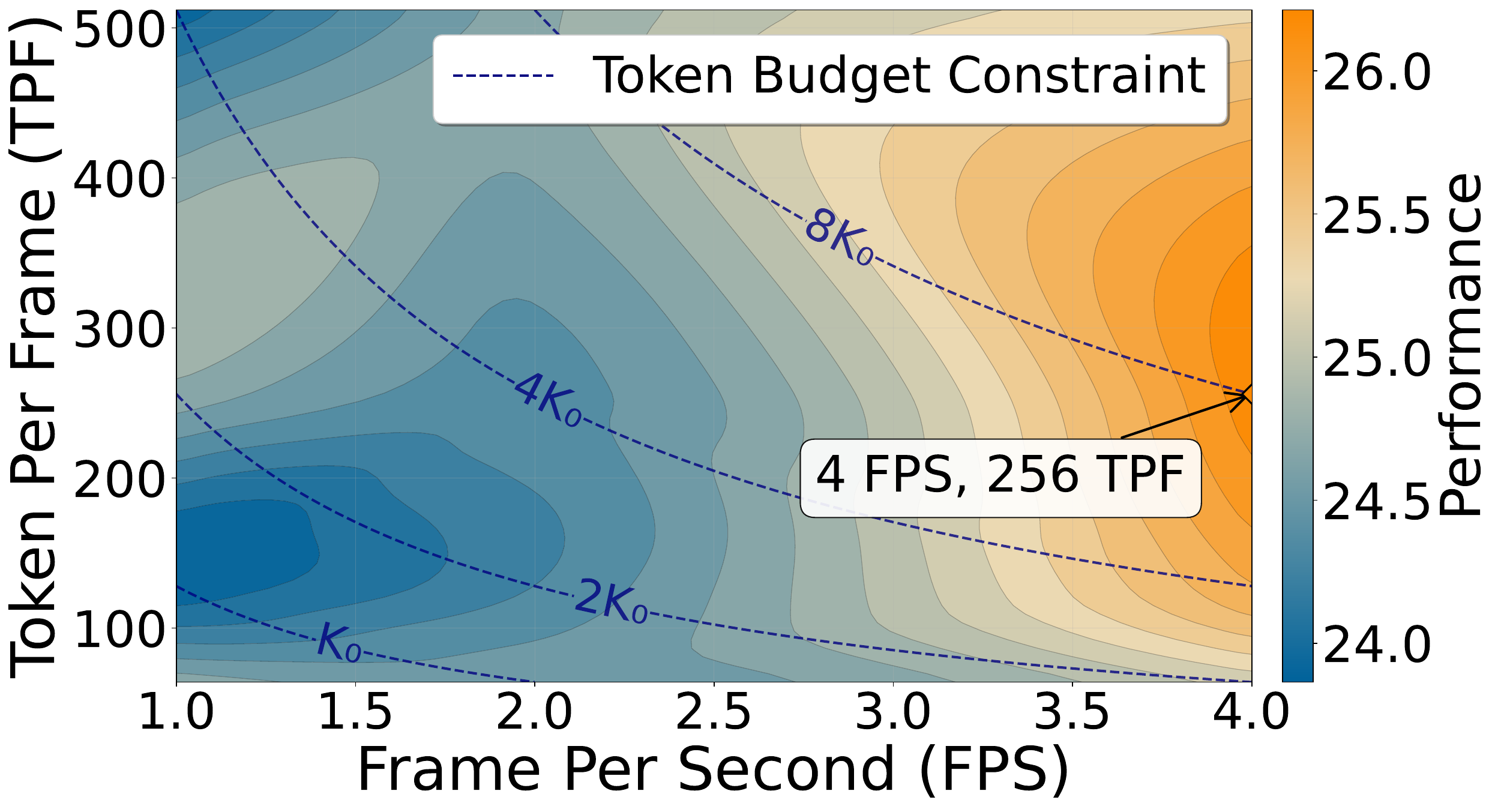}} &
         \subfloat[Information Scaling of VSIBench]{\includegraphics[width=0.48\textwidth]{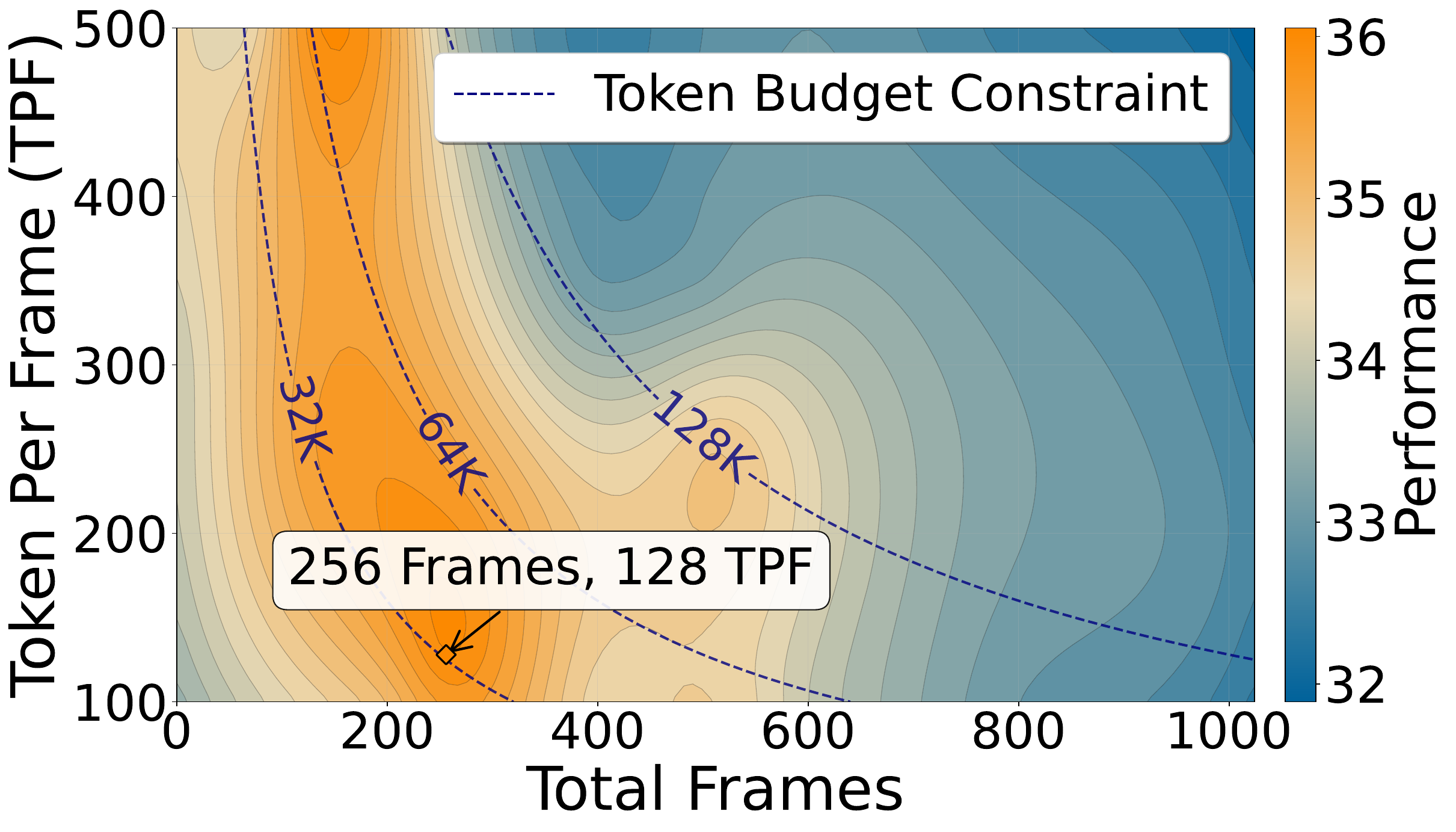}} \\
    \end{tabular}
    \caption{Scaling Performance of~\model~under Different Context Allocation Strategies. 
We highlight the Token Budget Constraint to indicate settings with equal context length, 
and annotate the best-performing configuration under each benchmark. Since videos in Tomato~\cite{shangguan2024tomato}, we vary FPS instead of total frames, with FPS~$\times$~TPF~=~128 denoted as $K_0$.}
\label{attention_budget}
\end{figure}


\definecolor{sand}{HTML}{f5f0E6}   
\definecolor{blue}{HTML}{00629B}   

\section{Scaling Analysis and Findings}
\label{sec:findings}

\finding{1}{Do learned sparse attention weights remain beneficial in dense attention settings?}


\definecolor{sand}{HTML}{f5f0E6}   
\definecolor{blue}{HTML}{00629B}   

\begin{table*}[h]
\centering
\vspace{-10pt}
\caption{Ablation study on transferring sparse attention weights to dense attention across tasks.} 
\addtolength\tabcolsep{-2.4pt} 
\resizebox{1\linewidth}{!}{
\begin{tabular}{ccccccc}
\toprule
\multicolumn{1}{c}{\multirow{2}{*}{Model}} & \multicolumn{4}{c}{Long Video Understanding}  & \multicolumn{1}{c}{Temporal Reasoning} & \multicolumn{1}{c}{Spatial Understanding} \\
\cmidrule(lr){2-5} \cmidrule(lr){6-6} \cmidrule(lr){7-7}
\multicolumn{1}{c}{}& LongVideoBench & MLVU$_{Test}$ & TimeScope & LongTimeScope & Tomato & VSIBench \\ 

\midrule
\textcolor{gray}{Qwen2.5-VL-7B} & \textcolor{gray}{58.7} & \textcolor{gray}{51.2} & \textcolor{gray}{81.0} & \textcolor{gray}{40.7} & \textcolor{gray}{22.6} & \textcolor{gray}{29.7}\\
Dense-SFT & 57.8 (-1.5\%) & 51.2 (+0.0\%) & 76.8 (-5.2\%) & 40.2 (-1.2\%) & 21.7 (-4.0\%) & 30.6 (+2.1\%)\\
Dense-NSA & 56.1 (-4.4\%) & 51.6 (+0.8\%) & \textbf{83.0 (+2.5\%)} & 40.9 (+0.5\%)& 23.4 (+3.5\%) & 33.1 (+10.7\%)\\
\model & \textbf{59.4 (+1.1\%)} & \textbf{51.8 (+1.2\%)} & 82.7 (+2.1\%) & \textbf{44.4 (+9.1\%)} & \textbf{26.2 (+15.9\%)} & \textbf{36.1 (+20.3\%)} \\

\bottomrule
\end{tabular}}

\vspace{-4mm}
\label{tab:nsa2full} 
\end{table*}

We further examine whether the learned QKV weights of~\model~ can imrpove performance in dense attention inference.
Table~\ref{tab:nsa2full} reports the relative performance change over the Qwen2.5-VL-7B~\cite{qwen2025qwen25technicalreport}. Due to the limited quality of the training data, our fine-tuned Qwen2.5-VL-7B (Dense-SFT) exhibits slight performance drops on most benchmarks.
We observe that the transferred model (Dense-NSA) allows the dense variant to recover and surpass the baseline on several benchmarks suggesting that sparse-trained weights provides inductive bias towards more effective attention distributions. However, the effect remains limited on LongVideoBench~\cite{wu2024longvideobench}.
\model~significantly outperforms Dense-NSA on most tasks, highlighting the importance of runtime sparsity and dynamic gating.

\finding{2}{How far can \model~scale in context length?}



The effective vision context length $L$ is jointly determined by the number of vision tokens per frame $T$ and the total number of input frames $F$.~\model~is trained with a maximum context length of $L=36K$ tokens, corresponding to $T=64$ tokens per frame. We conduct an information budget study under a fixed context length, by varying tokens per frame and frame rate. We then scale up the context length beyond the training budget, evaluating up to the maximum 128K tokens supported by the language model. As observed in Figure~\ref{attention_budget}, the model consistently achieves higher performance when scaled to longer contexts beyond its training length across benchmarks. However, the ideal allocation of same token budget is highly task-dependent. LongVideoBench~\cite{wu2024longvideobench} favors allocating more tokens per frame, while Tomato~\cite{shangguan2024tomato} and TimeScope~\cite{zohar2025apollo2} benefit more from increasing the number of frames, emphasizing temporal coverage. VSIBench~\cite{yang2025thinkingspacemultimodallarge} shows mixed preferences depending on context length, reflecting a balance between spatial and temporal sampling. Additional results on information scaling are reported in Appendix~\ref{information_scaling}.



\finding{3}{How to allocate the attention budget?}

\begin{figure}[!ht]
    \captionsetup{skip=2pt}
    \centering
    \begin{tabular}{ccc}
         \subfloat[Attention Scaling of MLVU]{
             \includegraphics[width=0.48\textwidth]{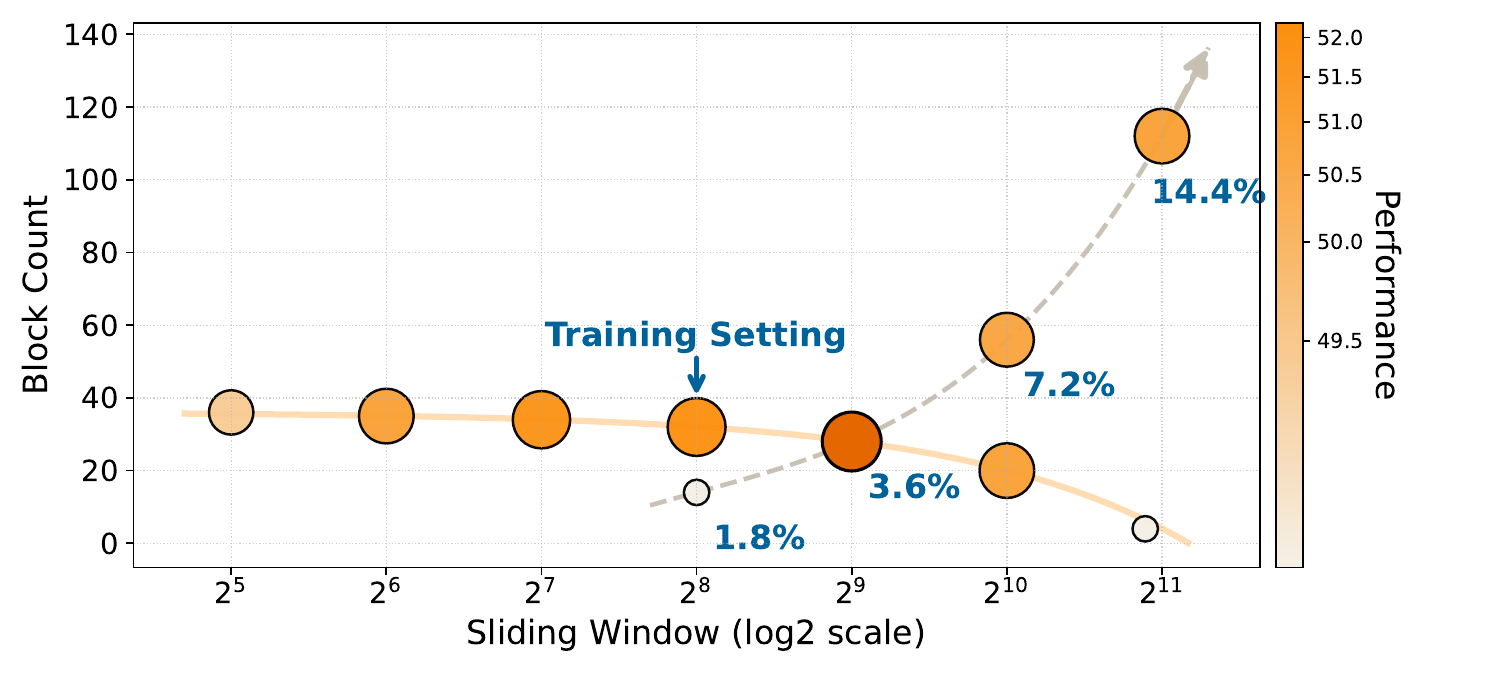}
         } &
         \subfloat[Attention Scaling of LongTimeScope]{\includegraphics[width=0.48\textwidth]{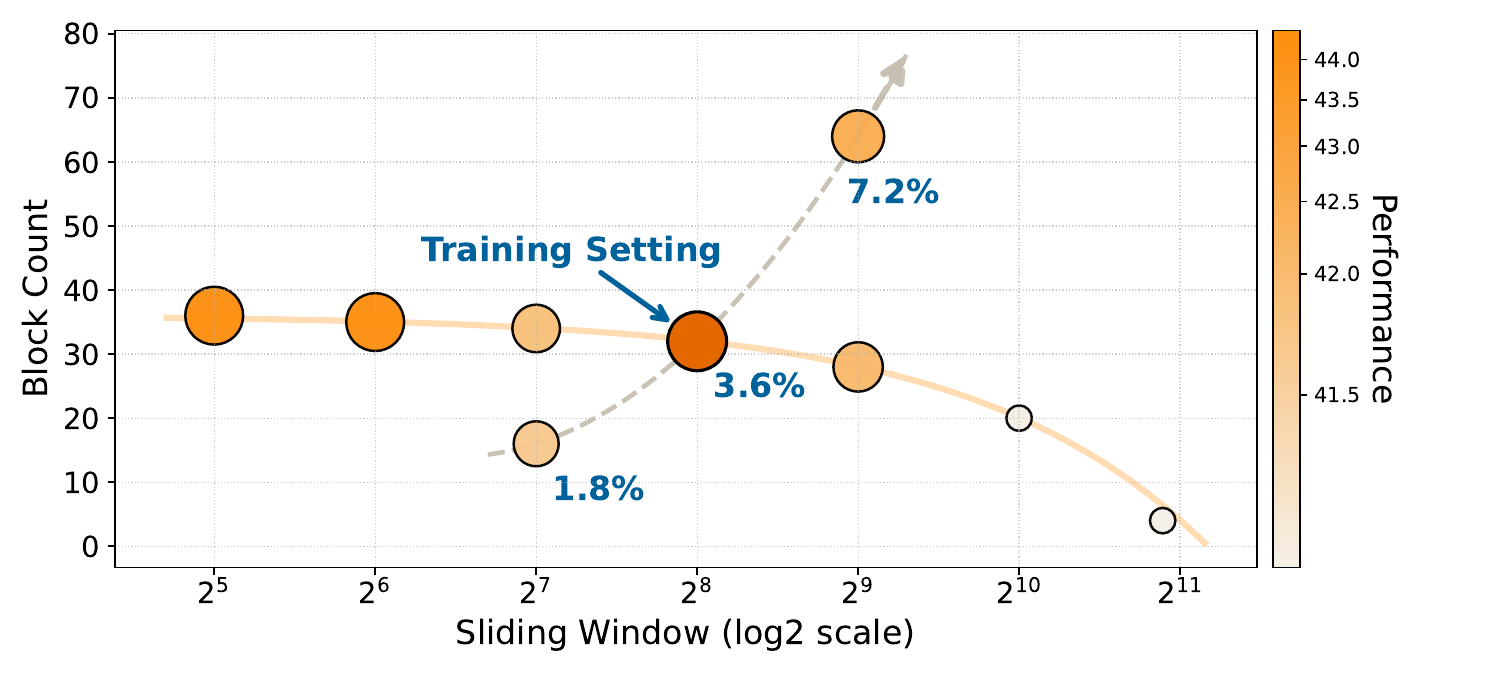}} 
         \\
         \subfloat[Attention Scaling of Tomato]{\includegraphics[width=0.48\textwidth]{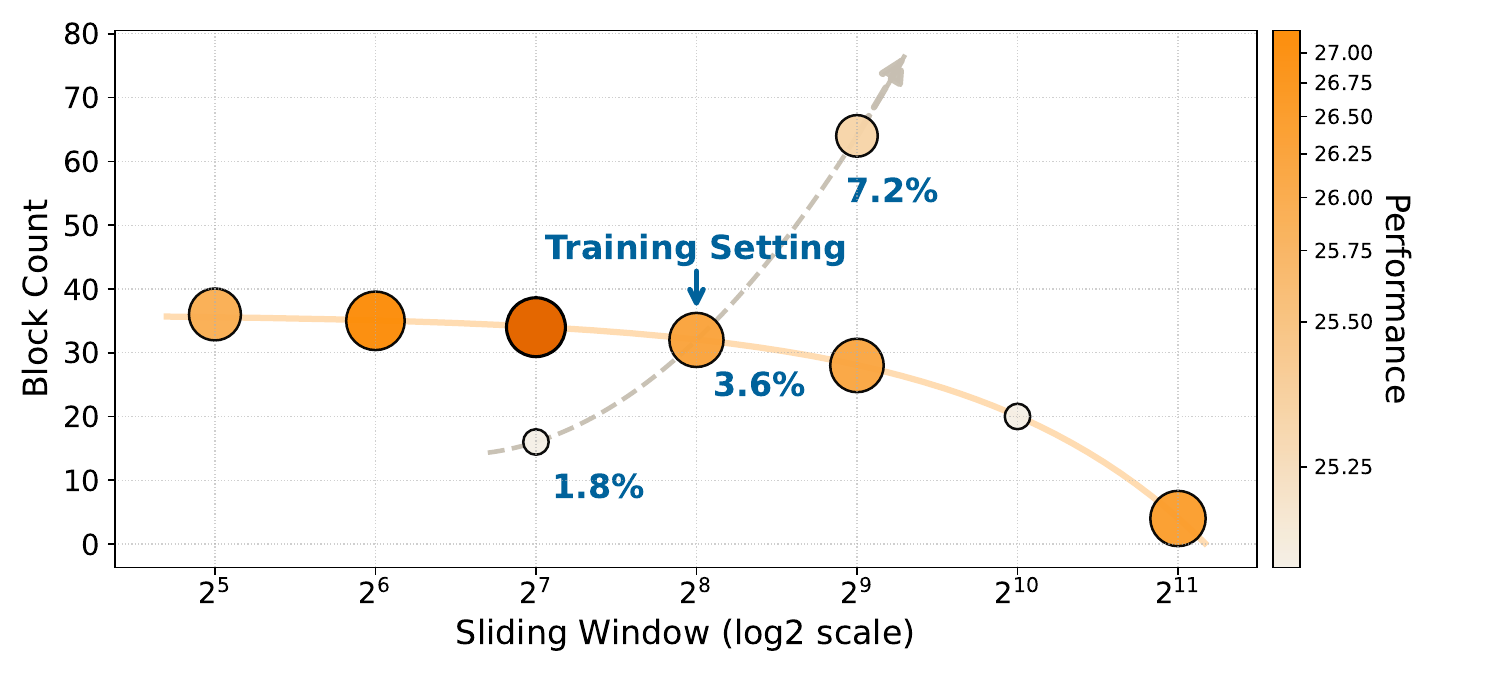}} &
         \subfloat[Attention Scaling of VSIBench]{\includegraphics[width=0.48\textwidth]{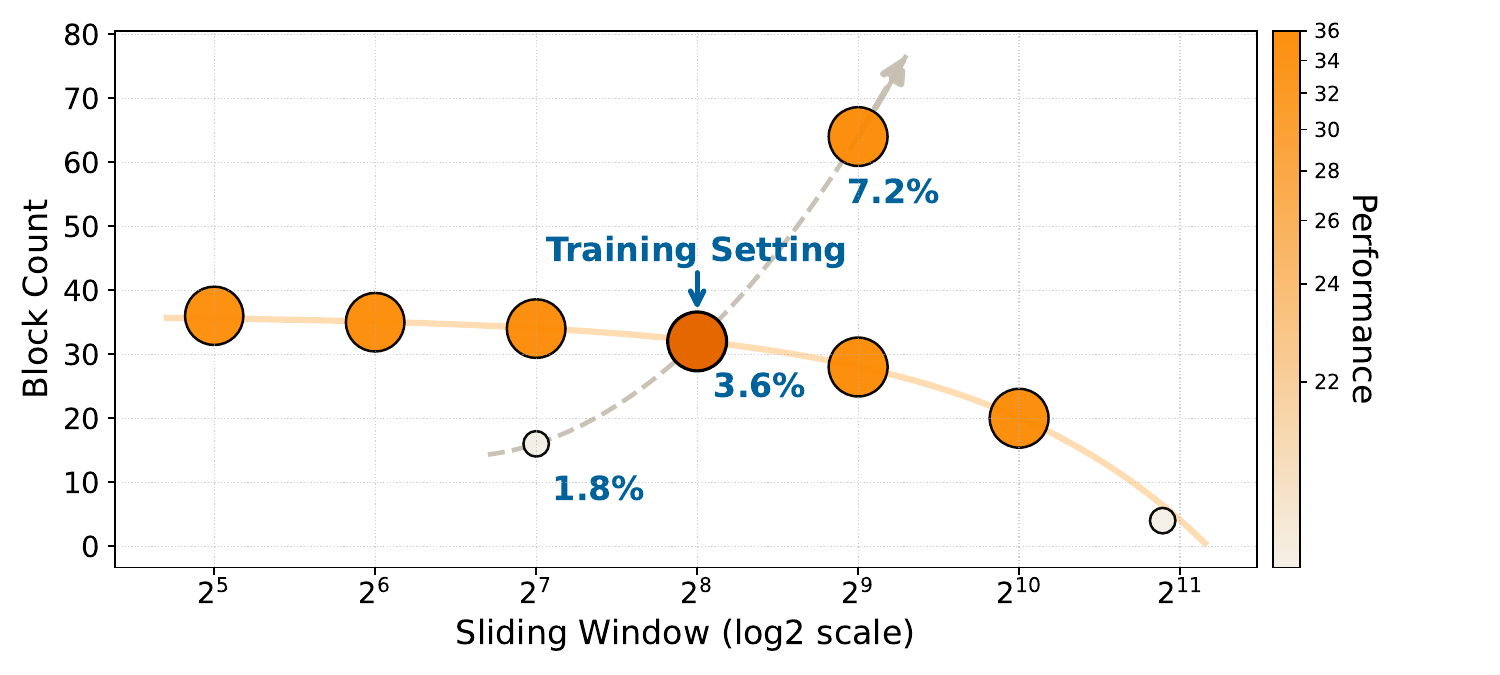}} \\
    \end{tabular}
    \caption{Scaling Performance of~\model~under Different Attention Allocation Strategies. Scatter points from small to large and from light to dark indicate increasing performance. We annotate the point corresponding to the same attention allocation strategy as used during training and connect configurations with equal attention budgets using solid orange lines. We further scale the best configuration using dashed lines. Percentages show attention relative to full attention.}
    \label{fig:attention_budget}
\end{figure}

We define the \textit{Attention Budget} as the total number of key-value pairs visible to each query, denoted by $K_{vis}$. 
It is composed of a global sparse component and a local sliding-window component as:
$
K_{attn} = b \times s + w,
$
where $b$ and $s$ denote the number and size of global blocks, and $w$ is the sliding-window width. 
With context length $L$, compared to causal dense attention with $\tfrac{L(L-1)}{2}$ edges, the fraction of attention used $\gamma$ is
\[
\gamma
= \frac{L(cS+w)}{\tfrac{L(L-1)}{2}}
= \frac{2(cS+w)}{L-1},
\]
To determine the optimal attention allocation, we first fix the total sequence length $L$, the attention budget $K_{\text{vis}}$, and the block size $S=64$, while systematically varying the local attention ratio $\alpha = \frac{w}{K_{\text{attn}}}$.
We then employ the optimal allocation ratio $\alpha^\star$ for attention budget scaling. 
As shown in Figure~\ref{fig:attention_budget}, scatter points denote different allocation strategies, with their size and color reflecting performance. 
We highlight the point corresponding to the training configuration, connect equal-budget settings with solid orange lines, 
and extend the best-performing configuration with dashed lines, where the annotated values indicate the fraction of attention used $\gamma$. 
Results show that model performance is highly sensitive to attention allocation. 
Although the optimal ratio between global and local attention varies across tasks, configurations close to the training allocation generally yield better results. 
Under the same budget, fine-tuning around the training setting often improves performance, whereas simply enlarging the overall budget does not consistently bring further gains. 
Moreover, across most benchmarks, increasing global attention (enlarging the block count) tends to outperform increasing local attention (enlarging the sliding window).
Remarkably,~\model~achieves leading performance using only $3.6\%$ of the full attention budget. More results are in Appendix~\ref{attention_scaling}.


 
\finding{4}{What roles do compression, selection, and sliding-window gates play in~\model?}

\begin{figure}[!ht]
    \centering
    \begin{minipage}[b]{0.52\textwidth}
        \centering
        \includegraphics[width=\textwidth]{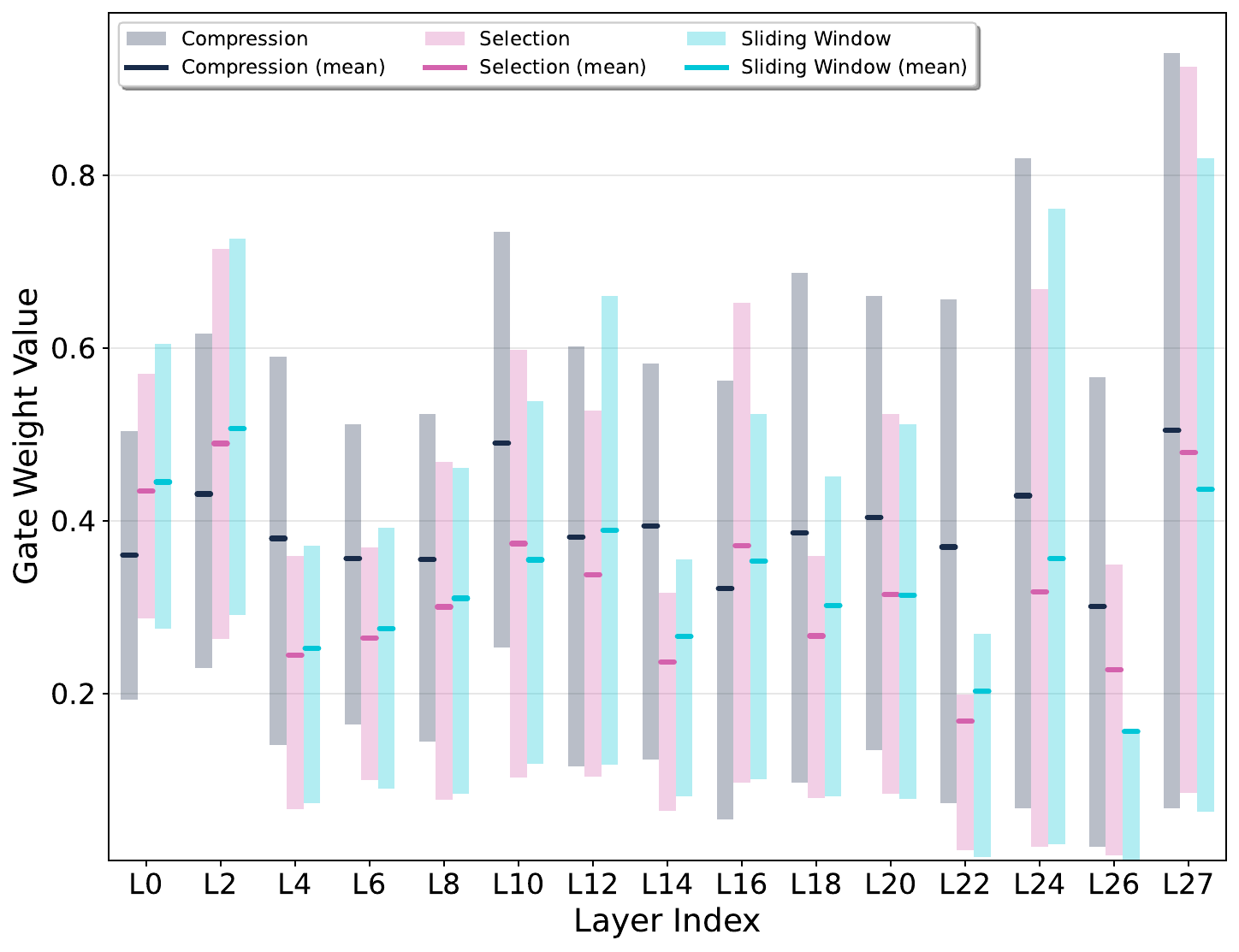}
        \caption{Gate weights across layers in~\model. Compression remains dominant, while selection and sliding-window weaken in later layers.}
        \label{fig:gate_dis}
    \end{minipage}
    \hfill
    \begin{minipage}[b]{0.45\textwidth}
        \centering
        \includegraphics[width=0.95\textwidth]{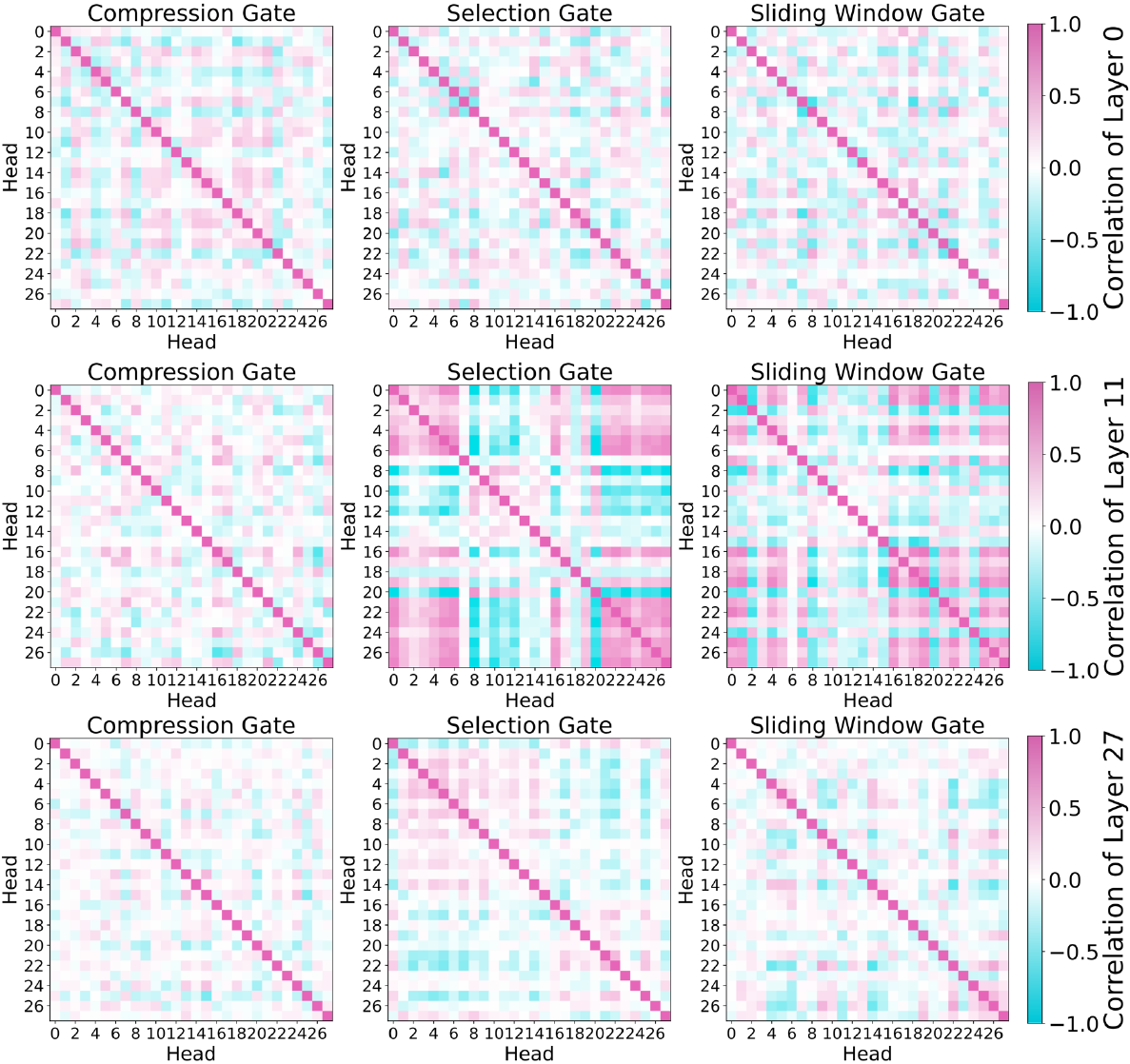}
        \caption{Inter-head similarities of gates in~\model. Selection and sliding-window gates show high similarity in middle layers.}
        \label{fig:inter_head}
    \end{minipage}
\end{figure}

We analyze the gating distribution of~\model~across Tomato~\cite{shangguan2024tomato}, VSIBench~\cite{yang2025thinkingspacemultimodallarge}, and LongVideoBench~\cite{wu2024longvideobench}, and aggregate the average routing gate weights over 100 examples from each. 
As illustrated in Figure~\ref{fig:gate_dis}, where shaded bars denote the interquartile range and horizontal lines represent mean values, each head in~\model~exhibits distinct and diverse preferences across branches throughout its full depth. The diversity allows different layers to specialize in distinct modes of the context-dependent information flow. The compression branch maintains relatively high average weights across most layers, underscoring its primary role in reducing redundancy while preserving salient features. The selection and sliding window gates fluctuate more strongly, occasionally surpassing the compression branch in early and middle layers. However, their contributions diminish in the final layers (e.g., L22–L26), demonstrating that the focus shifts towards aggregating high-level features. We also note strange behavior in the last layer, where all three branches are fully active despite selection and sliding window being inactive in the layers before. Full gate values distribution in Appendix~\ref{fullgate_dis}.

We further dive into the inter-head gate similarity of each layer in Figure~\ref{fig:inter_head}. In the middle layers, both selection and sliding window gates exhibit pronounced increases in inter-head similarity. This indicates that multiple mid-layer heads converge to highly consistent gating behaviors when the model performs block selection and local temporal integration. However, the compression gate shows consistently low inter-head similarity, indicating that it operates largely in a head-independent manner. At both the initial and final layers of~\model, inter-head similarity remains weak across all gates, reflecting the need to maintain diversity in early representations and to support mixing information in higher-level abstractions. More inter-head gate similarites visualization in Appendix~\ref{all_gate_corr}.

\finding{5}{Where does the efficiency bottleneck come from?}




\begin{wrapfigure}{r}{0.35\textwidth} 
  \centering
  \vspace{-8pt}
  \includegraphics[width=\linewidth]{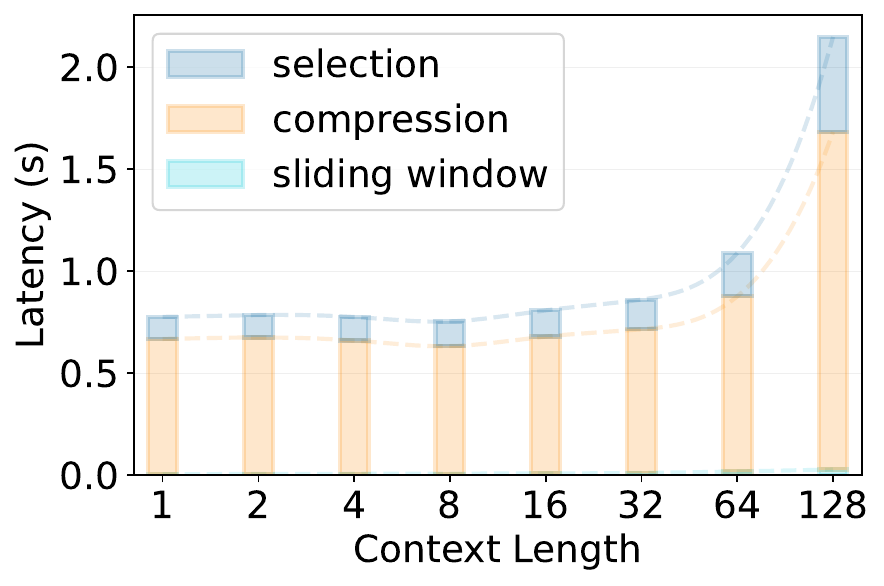} 
  \vspace{-8pt}
  \caption{Inference latency of each branch in~\model.}
  \label{fig:bottleneck}
  \vspace{-5pt}
\end{wrapfigure}


We measure the inference latency of each branch in~\model~using wall-clock time across varying context lengths from $1K$ to $128K$. The compression branch dominates runtime as the context grows, while the selection and sliding window branches contribute relatively little at longer contexts. Ideally, the compression branch grows approximately linearly with $L$, and the sliding window branch has a complexity of $O(L \cdot w)$, which results in linear scaling for a fixed window size $w$. The selection branch requires computing importance scores over all $L/b$ blocks per query, leading to a computational complexity of $O(L^2/b)$. However, wall-clock latency deviates from these estimates due to hardware parallelism, memory access patterns, and kernel launch overheads. Overall, the compression branch emerges as the primary bottleneck, highlighting the need for further optimization of its kernel design and memory efficiency.

\finding{6}{Do learnable sparse mechanisms induce dynamic attention sinks?}

In decoder-only transformers, a disproportionate amount of attention is often allocated to the first few tokens, which act as attention sinks and absorb excessive attention mass as a byproduct of softmax normalization. 
Prior studies~\cite{gu2024attention,xiao2023efficient} show that attention sinks arise from massive activations and unusually small key and value norms, so attention directed to these tokens contributes little to the residual state.
This raises an important question in learnable sparse attention: whether sparsity patterns amplify or mitigate such sinks. 

\begin{figure}[!ht]
    \centering
    \includegraphics[width=\linewidth]{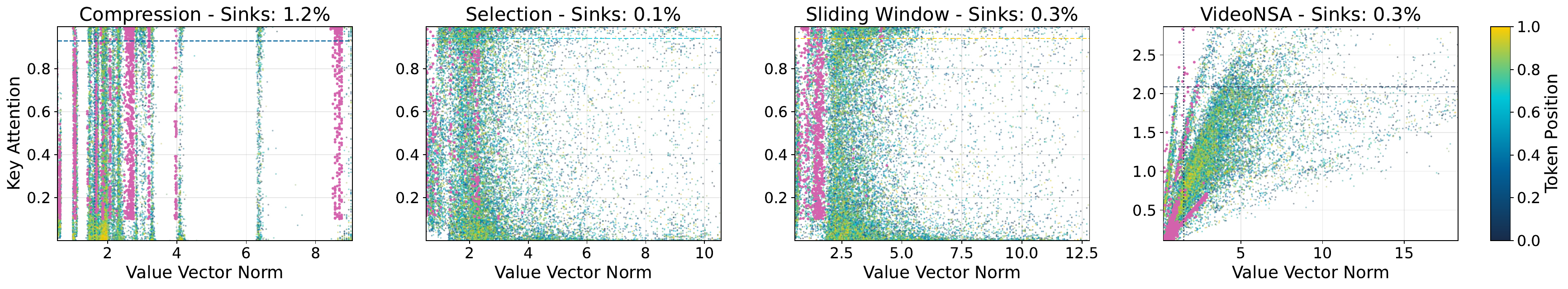}
    \caption{Attention sinks distribution of different branches. ~\model~maintains a low overall sink ratio, with pink points indicating identified sinks.}
    \label{fig:nsa_sink}
\end{figure}

\begin{figure}[!ht]
    \centering
    \begin{minipage}[b]{0.49\textwidth}
        \centering
        \includegraphics[width=\textwidth]{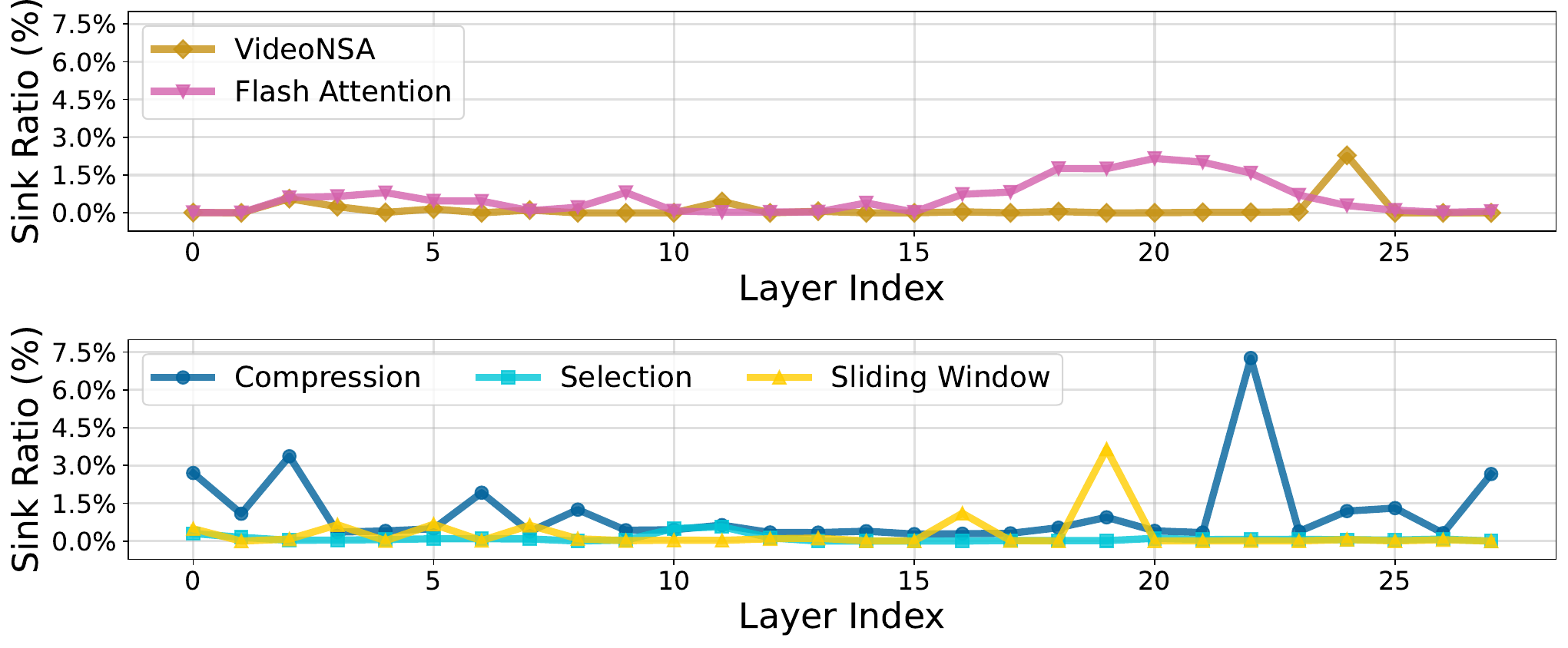}
        \caption{Layer-wise attention sink ratio distribution in different branches and Flash Attention.}
        \label{fig:sink_layer}
    \end{minipage}
    \hfill
    \begin{minipage}[b]{0.49\textwidth}
        \centering
        \includegraphics[width=\textwidth]{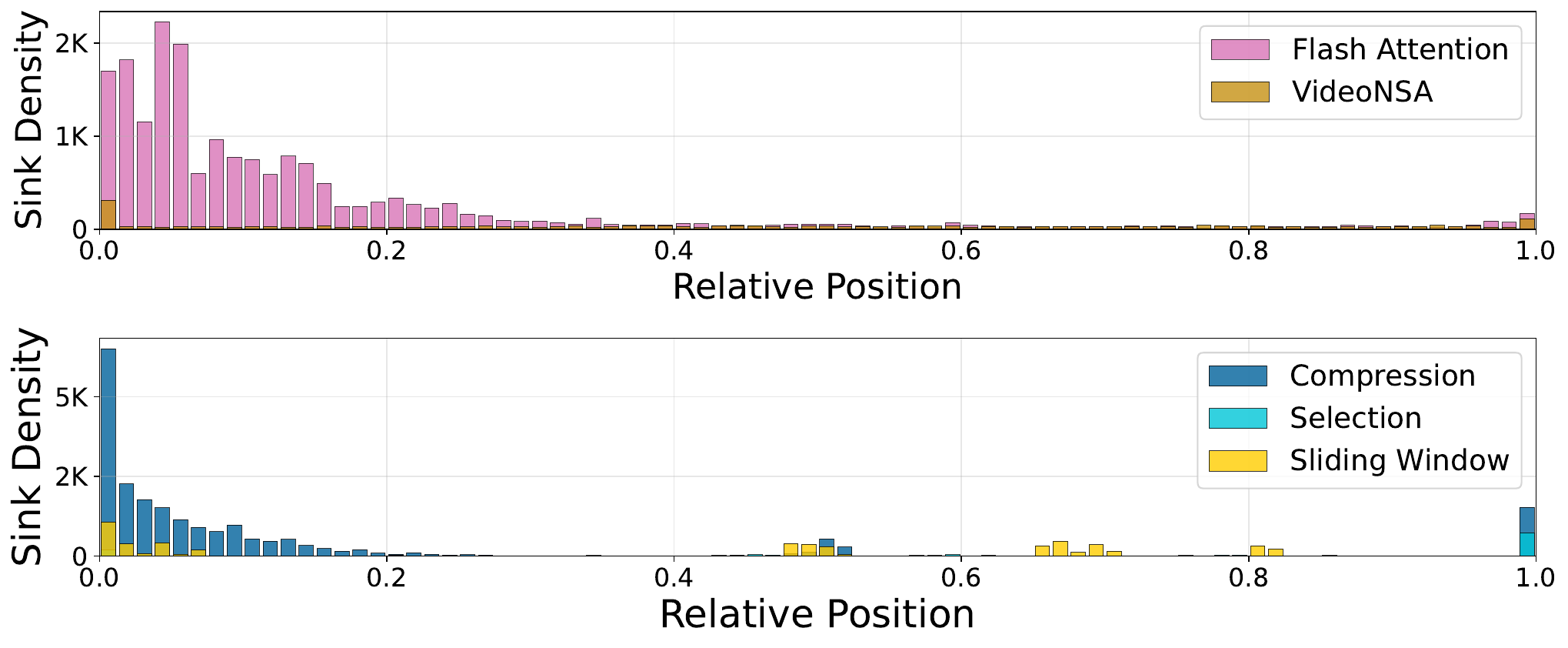}
        \caption{Relative positions of attention sinks in different branches and Flash Attention.}
        \label{fig:sink_position}
    \end{minipage}
\end{figure}

We follow the attention sink defination in~\cite{Pai2025SparsityIsCool}:
\[
\text{Attention Sink} = 
\mathbf{1}\Big\{ \alpha > 0.1 \ \land \ \|v\| < \mathrm{median}(\|v\|) - 2 \cdot \mathrm{IQR}(\|v\|) \Big\},
\]
where $\alpha$ is the average attention score received by the key, 
and $\|v\|$ is the value norm of the token.

Figure~\ref{fig:nsa_sink} illustrates the average distribution of attention sinks across the three branches of~\model. Each frame is encoded into 256 tokens, and we adopt the same sparse attention configuration as used during training. The three branches exhibit markedly different sink behaviors. 
The compression branch produces the most sinks, with distinct banded concentrations along the value norm axis caused by token merging that amplifies some token norms while suppressing others. Conversely, the selection branch yields almost no sinks, as its top-$k$ block filtering mechanism enforces a smoother value norm distribution. Notably, the sliding window branch demonstrates a clearer separation between sink and non-sink tokens along the value norm axis. Critically, dynamic gating allows~\model~to counteract the negative effects of the compression branch, achieving a stable model with a low overall sink ratio of 0.3\%. 

Figure~\ref{fig:sink_layer} indicates that~\model~maintains low sink ratios overall, with only minor fluctuations across layers. However, Flash Attention exhibits a gradual increase in sink ratios toward deeper layers. The compression branch maintains relatively high sink levels across most layers. The selection branch remains consistently close to zero, while the sliding window branch occasionally shows higher peaks in the middle-to-late layers, indicating that locality constraints may still introduce bias in long-sequence settings. 
From the perspective of positional distribution in Figure~\ref{fig:sink_position}, Flash Attention produces sinks that are uniformly spread across the entire sequence due to its fully connected dense attention. Under dynamic gating,\model~achieves smoother temporal coverage, alleviating over-reliance on early positions while avoiding the global diffusion characteristic of dense attention.
In contrast, the compression branch exhibits strong accumulation at the beginning with an even steeper decay, indicating that token merging exerts its strongest impact on early-stage representations. The selection branch yields very few sinks across the sequence, while the sliding window branch produces sparse peaks at periodic boundaries of local neighborhoods. More analysis about attention sinks on various sparse attention settings can be found in Appendix~\ref{block_sink}.



\section{Conclusion}


In this work, we present~\model, a hybrid hardware-aware sparse attention model that significantly advances video understanding across various tasks. By dynamically fusing block-wise compression, salient block selection, and a sliding window,~\model~effectively preserves critical information while achieving near-linear scalability in efficiency and memory. Our experiments demonstrate that~\model~consistently outperforms existing methods on key tasks including long video understanding, temporal reasoning, and spatial understanding. While the prefill stage remains the primary bottleneck, our findings confirm that this hybrid sparse approach provides a powerful and scalable framework, paving the way for more capable video foundation models.
\section{Acknowledgement}
This work is supported by NSF award IIS-2127544 and NSF award IIS-2433768.
We thank Lambda, Inc. for their compute resource help on this project.

\section{Ethics Statement}

This research on video understanding utilizes publicly available datasets, ensuring that all data complies with privacy regulations. We acknowledge the potential biases that can arise in automatic answer generation, particularly concerning gender, race, or other characteristics. We have taken measures to evaluate and minimize such biases, while remaining committed to further improvements. Additionally, we recognize the potential risks of misuse, such as generating misleading answers, and have checked the training dataset with safeguards against such applications.

\section{Reproducibility Statement}

We have made several efforts to ensure the reproducibility of our work. All the key implementation details, including the architecture of our model, the training procedures, and hyperparameter settings, are described in supplementary meterial Section~\ref{sec:detailed_training_settings}. The settings of the used evaluation benchmarks are in Section~\ref{sec:evaluation_datasets_and_settings} to further support reproducibility.

\section{The Use of Large Language Models}

Large language models (LLMs) were used only for light editorial purposes, such as minor grammar checking and language polishing. They were not used for generating scientific content, research ideation, experiment design, or analysis. The authors take full responsibility for the entirety of the paper, and LLMs are not considered contributors or eligible for authorship.

\bibliography{iclr2026_conference}
\bibliographystyle{iclr2026_conference}

\newpage
\appendix
\begin{center}
     \Large\textbf{Appendix}
\end{center}

The supplementary material is structured as follows:

\begin{itemize}
    \item Iiterature review about the related works in Section~\ref{related}.
    \item The training settings for~\model~in Section~\ref{train_detail}.
    \item The introduction of the used evaluation benchmarks and settings in Section~\ref{sec:evaluation_datasets_and_settings}.
    \item More results on long-form video benchmarks in Section~\ref{base_long_results}.
    \item More results on temporal reasoning benchmarks in Section~\ref{base_reasoning_results}.
    \item More results on spatial understanding benchmarks in Section~\ref{base_spatial_results}.
    \item  Results on additional video understanding benchmarks in Section~\ref{add_ben}.
    \item Visualization of attention pattern in each branch in Section~\ref{vis_attn_branch}.
    \item More results on branch combination in Section~\ref{branch_comb}.
    \item More results on information scaling study in Section~\ref{information_scaling}.
    \item  Additional context-length scaling results of Qwen2.5-VL in Section~\ref{qwen25_context}.
    \item More results on attention scaling study in Section~\ref{attention_scaling}.
    \item Theoretical analysis of context length and attention budget scaling in Section~\ref{theory}.
    \item Full gate values distribution in Section~\ref{fullgate_dis}.
    \item More inter-head gate similarites visualization in Section~\ref{all_gate_corr}.
    \item Benchmark-level gating analysis and PCA visualization in Section~\ref{pca}.
    \item Additional analysis of training and inference efficiency in Section~\ref{flops}.
    \item Additional analysis on CMP latency bottleneck in Section~\ref{cmp_latency}.
    \item More analysis about attention sinks on various sparse attention settings can be found in Section~\ref{block_sink}.
    \item Comparison of sparsity patterns between text-only NSA and VideoNSA in Section~\ref{text_only}.
    \item Visualization of attention sinks in dense attention in Section~\ref{dense_attn_sink}.
\end{itemize}

\section{Related Work}
\label{related}

\subsection{Efficient Video Understanding}
Video understanding systems typically convert videos into long sequences of vision tokens, which can easily exceed GPU memory and slow down inference as the video length grows. To address this, 
existing work mainly address this by \textbf{token compression}, \textbf{alternative sequence modeling}, and \textbf{KV-cache compression}. One important line of work emphasizes token compression. Spatial or temporal token merging methods~\cite{Wang2025LVC,Zhang2025NeuralDiscrete,Li2025ImprovingLLMVideo16fps,Jiang2025STORM,Li2025BreakingEncoder,Shao2025TokenCompressionSurvey,song2024moviechat,chai2024auroracap,Liao2025AreWU,Wu2025OnTG,Wu2024ZeroShotLV,Wu_2025_CVPR,Wu2024VideoRF,Cao2024ReframeAL,Li2025VEUBenchTC,Li2025UnveilingEI,Yang2023ExploringDI} progressively discard redundant content, while question-/task-aware strategies~\cite{Jiang2025VISA,Dong2025MMTok,Yao2025TimeChatOnline,song2025moviechat+} tailor retained tokens to the query. These approaches substantially lower FLOPs but still rely on dense attention once tokens are merged. Beyond pure self-attention, Mamba-based or hybrid architectures~\cite{Jiang2025STORM,Ren2025Vamba,xu2025auroralong,Tao2025InfiniteVLSL} inject state-space or recurrent modules to approach linear-time inference while preserving long-range dependencies. Also, there exists approach to  design data efficient systems for further fine-tuning~\cite{li2025reinforcement}. Another direction targets the key--value cache during decoding via task-aware sparsification and streaming-friendly memory~\cite{Qin2025VideoXL2,Ning2025LiveVLM,Kim2025InfiniPotV,Yang2025StreamMem,Liu2025TimeScopeTT} reduce memory and improve throughput, yet prefill still scales quadratically with sequence length. In contrast to methods that mostly decide \emph{where} to drop or compress tokens, our approach systematically probe the effectiveness of \emph{native sparse attention}~\cite{Yuan2025NativeSparseAttention} that restructures attention itself to be learnable and sparse from the ground up.~\model~attains near-linear scalability up to 128K tokens and processes over 10{,}000 frames on a single GPU, outperforming compression-only pipelines on long-video understanding, temporal reasoning, and spatial understanding tasks.

\subsection{Sparse Attention Mechanism}
Sparse attention is a central strategy for efficient long-context modeling in language and multimodal systems. Surveys \citep{zhang2025efficient} categorize approaches into \emph{pattern-based} vs.\ \emph{dynamic/learned}. \textbf{Pattern-based sparsity.} Methods such as Longformer \citep{beltagy2020longformer}, StreamingLLM \citep{xiaoefficient}, and TriangleMix \citep{He2025TriangleMix} prescribe fixed local/strided patterns that can be applied training-free; recent multimodal works \citep{Zhang2025DAM,Yang2025LessIsMore} follow similar principles, while hardware-efficient kernels like Flash Sparse Attention \citep{Yan2025FlashSparse} further reduce prefill latency. InfLLM-V2~\cite{zhao2025infllm} uses switchable dense sparse attention to smoothly adapt models from short to long sequences while maintaining consistency and achieving efficient acceleration with high performance. ProxyAttn~\cite{wang2025proxyattn} uses representative heads for fine-grained block importance estimation, enabling faster sparse attention with minimal performance loss. \textbf{Dynamic and trainable sparsity.} Content- or gradient-adaptive mechanisms select important connections (e.g., diagonal selection \citep{Tyagi2025DynamicDiagonal} or lag-relative strategies \citep{Liang2025LagRelative}); trainable sparse attention improves long-context reasoning \citep{Gao2025SeerAttentionR,Vasylenko2025LongContext,gao2024seerattention}, diffusion-based video generation \citep{Zhang2025VSA}, and state-space models \citep{Zhan2025ContextDependent}. SLA~\cite{zhang2025sla} decomposes attention weights into critical, marginal, and negligible parts, combining sparse and low-rank acceleration to greatly reduce computation while preserving generation quality. Hybrid approaches such as RocketKV \citep{Behnam2025RocketKV} combine token/cache compression with learned sparsity, and MMInference \citep{Li2025MMInference} accelerates modality-aware sparse prefill for VLMs. Despite these advances, most techniques are optimized for text or short multimodal contexts and do not directly address the ultra-long, highly redundant spatio-temporal structure of videos. VideoNSA unifies \emph{block-wise compression}, \emph{salient block selection}, and a \emph{sliding-window} branch under learnable gates that dynamically allocate computation across three native sparse branches \citep{Yuan2025NativeSparseAttention}. This end-to-end, data-driven design preserves critical global/local dependencies while scaling nearly linearly in both time and memory.
\section{Detailed Training Settings}
\label{train_detail}
\label{sec:detailed_training_settings}

Training hyperparameters for~\model~are shown in Table~\ref{tab:training_hyperparam}. We filter a subset of LLaVA-Video-178K~\cite{zhang2024videoinstructiontuningsynthetic} as the training data. For each video, we uniformly sample at 4 frames per second and retain only those with 350–550 frames, resulting in 216K video question–answer pairs from the original 961K pairs in LLaVA-Video-178K~\cite{zhang2024videoinstructiontuningsynthetic}.

\begin{table*}[h]
\caption{Training hyper-parameters for \model.}
\centering
\scriptsize
\resizebox{0.8\textwidth}{!}{\begin{tabular}{l | c }
\toprule
Hyper-parameters & Fine-tuning \\
\midrule
trainable parameters & ViT + MLP + LLM \\
\rowcolor{gray!15}
warmup schedule & linear \\
warmup start factor & 1e-5 \\
\rowcolor{gray!15}
warmup ratio & 0.1 \\
learning rate schedule & cosine \\
\rowcolor{gray!15}
optimizer & AdamW~\cite{loshchilov2017decoupled} \\
optimizer hyper-parameters & $\beta_1, \beta_2=(0.9, 0.999)$\\
\rowcolor{gray!15}
weight decay & 0.01 \\
max norm & 1 \\
\rowcolor{gray!15}
epoch & 1 \\
peak learning rate & 1e-6 \\
\rowcolor{gray!15}
total equivalent batch size & 32 \\
\bottomrule
\end{tabular}}
\label{tab:training_hyperparam}
\end{table*}
\section{Evaluation Benchmarks and Settings}
\label{sec:evaluation_datasets_and_settings}

We list all the hyper-parameters and prompt used for evaluation as shown in Table~\ref{tab:evaluation_settings}. 

\begin{table*}[h]
\caption{Evaluation settings summary for each benchmarks. For all benchmarks we set temperature, top p, number of beams to 0, 0, 1 respectively. \# TPF stands for the vision tokens per frame, and \# F stands for the number of sampling frames.}
\centering
\scriptsize
\resizebox{0.75\textwidth}{!}{\begin{tabular}{l c c c}
\toprule
Benchmark & \# TPF & \# F & \# Max New Tokens  \\
\midrule
LongVideoBench~\cite{wu2024longvideobench} & 512 & 256 & 32 \\
LongTimeScope~\cite{zohar2025apollo2} & 128 & 512 & 16  \\
TimeScope~\cite{zohar2025apollo2} & 64 & 2048 & 16  \\
MLVU$_{test}$~\cite{zhou2024mlvu} & 128 & 512 & 16 \\
Tomato~\cite{shangguan2024tomato} & 4FPS & 256 & 1024 \\
VSIBench~\cite{yang2025thinkingspacemultimodallarge} & 256 & 128 & 16 \\
\bottomrule
\end{tabular}}
\label{tab:evaluation_settings}
\end{table*}

\section{More Results on Long-form Video Benchmarks}
\label{base_long_results}

\setlength{\tabcolsep}{3pt}
\begin{table*}[ht]
\centering
\caption{LongTimeScope results across baselines. Metrics include overall accuracy and task-specific scores across different steps. Flash Attn stands for Qwen2.5-VL-7B~\cite{qwen2025qwen25technicalreport} accelerated by Flash Infer, and Flash Attn + SFT stands for our fine-tuning version.}
\resizebox{0.98\textwidth}{!}{
\begin{tabular}{lcccccccccc}
\toprule
\multirow{2}{*}{Method} & \multirow{2}{*}{Overall} 
    & \multicolumn{3}{c}{18000} 
    & \multicolumn{3}{c}{28800} 
    & \multicolumn{3}{c}{36000} \\
\cmidrule(lr){3-5}\cmidrule(lr){6-8}\cmidrule(lr){9-11}
 &  & OCR & QA & Temporal & OCR & QA & Temporal & OCR & QA & Temporal \\
\midrule
Flash Attn      & 40.7 & 54.0 & 42.0 & 22.0 & 48.0 & 60.0 & 24.0 & 48.0 & 58.0 & 10.0 \\
Flash Attn + SFT& 40.2 & 46.0 & 30.0 & 34.0 & 46.0 & 44.0 & 36.0 & 52.0 & 44.0 & 20.0 \\
AWQ             & --   & --   & --   & --   & --   & --   & --   & --   & --   & --   \\
XAttn           & 41.1 & 52.0 & 56.0 & 30.0 & 54.0 & 52.0 & 6.0  & 52.0 & 64.0 & 4.0  \\
MInference      & 44.4 & 64.0 & 56.0 & 26.0 & 58.0 & 60.0 & 8.0  & 56.0 & 66.0 & 6.0  \\
tri-shape       & 28.4 & 34.0 & 36.0 & 12.0 & 48.0 & 48.0 & 0.0  & 44.0 & 32.0 & 2.0  \\
FlexPrefill     & 39.1 & 52.0 & 46.0 & 24.0 & 46.0 & 56.0 & 14.0 & 46.0 & 66.0 & 2.0  \\
FastV           & 35.6 & 36.0 & 50.0 & 16.0 & 44.0 & 50.0 & 4.0  & 44.0 & 64.0 & 12.0 \\
VisionZip       & 31.1 & 38.0 & 32.0 & 14.0 & 56.0 & 46.0 & 0.0  & 44.0 & 46.0 & 4.0  \\
VScan           & 40.4 & 48.0 & 52.0 & 24.0 & 50.0 & 52.0 & 22.0 & 46.0 & 64.0 & 6.0  \\
VideoNSA        & 44.4 & 50.0 & 54.0 & 30.0 & 54.0 & 72.0 & 0.0  & 48.0 & 76.0 & 16.0 \\
\bottomrule
\end{tabular}
}
\label{tab:base_longtime}
\end{table*}

\setlength{\tabcolsep}{3pt}

\begin{table*}[ht]
\centering
\caption{LongVideoBench results across baselines. Metrics include overall accuracy and task-specific scores across different steps. Flash Attn stands for Qwen2.5-VL-7B~\cite{qwen2025qwen25technicalreport} accelerated by Flash Infer, and Flash Attn + SFT stands for our fine-tuning version.}
\resizebox{0.98\textwidth}{!}{
\begin{tabular}{lcccccccccccccccccccccc}
\toprule
Method & Overall & 600 & TOS & S2E & E3E & S2A & SAA & O3O & T3O & T3E & O2E & T2O & S2O & TAA & T2E & E2O & SSS & T2A & 60 & SOS & 15 & 3600 \\
\midrule
Flash Attn      & 58.7 & 58.5 & 38.4 & 69.9 & 66.0 & 70.5 & 56.9 & 57.6 & 58.1 & 53.4 & 63.2 & 63.2 & 55.6 & 52.4 & 63.1 & 63.1 & 39.2 & 64.6 & 72.7 & 61.7 & 65.6 & 52.3 \\
Flash Attn + SFT& 57.8 & 55.8 & 38.4 & 65.6 & 61.7 & 73.9 & 56.9 & 65.2 & 55.4 & 57.5 & 57.5 & 56.6 & 62.5 & 51.2 & 55.4 & 66.2 & 39.2 & 63.3 & 74.4 & 58.0 & 64.6 & 52.0 \\
AWQ             & 59.0 & 60.0 & 34.2 & 72.0 & 64.9 & 68.2 & 59.7 & 57.6 & 52.7 & 50.7 & 69.0 & 52.6 & 63.9 & 57.3 & 61.5 & 67.7 & 43.3 & 62.0 & 73.8 & 63.0 & 67.7 & 50.9 \\
XAttn           & 59.1 & 59.4 & 36.0 & 70.0 & 66.0 & 67.2 & 57.3 & 58.1 & 55.8 & 53.8 & 64.5 & 62.2 & 64.3 & 56.3 & 65.2 & 66.7 & 41.3 & 58.5 & 75.2 & 60.7 & 68.8 & 50.6 \\
MInference      & 59.2 & 60.6 & 34.6 & 74.3 & 66.0 & 68.3 & 58.7 & 56.6 & 53.1 & 52.4 & 68.0 & 56.9 & 60.1 & 58.8 & 60.5 & 65.2 & 39.2 & 63.6 & 74.6 & 66.9 & 67.3 & 50.8 \\
tri-shape       & 59.5 & 60.9 & 34.6 & 73.2 & 66.0 & 69.5 & 58.7 & 58.1 & 55.8 & 52.4 & 68.0 & 55.6 & 61.5 & 60.0 & 60.5 & 66.7 & 38.2 & 63.6 & 74.6 & 66.9 & 67.8 & 51.1 \\
FlexPrefill     & 58.4 & 61.7 & 31.5 & 65.6 & 62.8 & 71.6 & 59.7 & 59.1 & 58.1 & 52.1 & 65.5 & 51.3 & 62.5 & 48.8 & 61.5 & 72.3 & 42.3 & 63.3 & 71.5 & 65.4 & 58.2 & 52.1 \\
FastV           & 57.3 & 57.3 & 43.8 & 64.5 & 60.6 & 70.5 & 52.8 & 56.1 & 52.7 & 48.0 & 59.8 & 67.1 & 56.9 & 48.8 & 67.7 & 66.2 & 40.2 & 58.2 & 69.8 & 61.7 & 70.9 & 48.9 \\
VisionZip       & 52.4 & 53.2 & 32.9 & 63.4 & 66.0 & 58.0 & 54.2 & 50.0 & 51.4 & 42.5 & 57.5 & 47.4 & 58.3 & 45.1 & 56.9 & 61.5 & 30.9 & 51.9 & 62.2 & 61.7 & 58.2 & 46.8 \\
VScan           & 58.7 & 57.0 & 29.5 & 69.0 & 65.0 & 69.6 & 56.3 & 54.1 & 56.1 & 55.5 & 61.2 & 58.5 & 61.9 & 60.2 & 58.0 & 73.4 & 41.4 & 61.3 & 74.2 & 65.9 & 73.7 & 50.3 \\
VideoNSA & 60.2	& 59.9	& 48.1	& 65.1	& 67.6 & 74.1 &	55.6 &	55.5 &	58.4 &	56.3 &	62.2	& 57.0 &	63.9 &	53.3 &	56.2 &	71.6 &	35.9 &	62.7	& 67.5	& 72.4 &	66.3	& 55.1 \\
\bottomrule
\end{tabular}
}
\label{tab:base_longvid}
\end{table*}

\setlength{\tabcolsep}{3pt}

\begin{table*}[ht]
\centering
\caption{MLVU results across baselines. Metrics include overall accuracy and task-specific scores across different steps. Flash Attn stands for Qwen2.5-VL-7B~\cite{qwen2025qwen25technicalreport} accelerated by Flash Infer, and Flash Attn + SFT stands for our fine-tuning version.}
\resizebox{0.98\textwidth}{!}{
\begin{tabular}{lcccccccccc}
\toprule
Method & Overall & PlotQA & Needle & Ego & Count & Order & Anomaly Reco & Topic Reason. & SportsQA & TutorialQA \\
\midrule
Flash Attn       & 51.2 & 58.0 & 68.3 & 52.8 & 31.7 & 25.7 & 46.2 & 79.1 & 38.9 & 48.8 \\
Flash Attn + SFT & 51.2 & 58.0 & 58.3 & 58.5 & 23.3 & 40.0 & 43.6 & 81.3 & 36.1 & 37.2 \\
AWQ              & 46.0 & 42.7 & 53.0 & 40.9 & 27.2 & 50.2 & 57.0 & 65.0 & 38.3 & 39.2 \\
XAttn            & 50.2 & 60.0 & 64.7 & 56.5 & 28.0 & 29.4 & 41.6 & 74.9 & 39.7 & 39.9 \\
MInference       & 49.2 & 56.0 & 64.7 & 48.9 & 29.7 & 26.6 & 41.6 & 77.1 & 39.7 & 39.9 \\
tri-shape        & 49.2 & 56.0 & 64.7 & 48.9 & 29.7 & 26.6 & 41.6 & 77.1 & 39.7 & 39.9 \\
FlexPrefill      & 46.0 & 54.0 & 54.7 & 42.7 & 24.7 & 40.9 & 36.6 & 72.6 & 29.7 & 32.2 \\
FastV            & 41.8 & 44.0 & 45.0 & 47.2 & 18.3 & 30.0 & 46.2 & 84.6 & 28.6 & 32.2 \\
VisionZip        & 33.1 & 30.0 & 26.7 & 30.2 & 6.7  & 22.9 & 41.0 & 68.1 & 19.7 & 26.4 \\
VScan            & 48.1 & 58.0 & 63.3 & 50.9 & 28.3 & 24.3 & 43.6 & 78.0 & 47.2 & 39.5 \\
VideoNSA   & 51.8	& 48.0	& 69.3	& 51.3	& 27.7	& 34.6	& 44.5	& 86.2	& 47.7	& 31.6  \\
\bottomrule
\end{tabular}
}
\label{tab:base_mlvu}
\end{table*}

\setlength{\tabcolsep}{3pt}
\begin{table*}[ht]
\centering
\caption{TimeScope results across baselines. Metrics include overall accuracy and task-specific scores across different steps. Flash Attn stands for Qwen2.5-VL-7B~\cite{qwen2025qwen25technicalreport} accelerated by Flash Infer, and Flash Attn + SFT stands for our fine-tuning version.}
\resizebox{0.98\textwidth}{!}{
\begin{tabular}{lccccccccccc}
\toprule
Method & Overall & 60 & 120 & 180 & 300 & 600 & 1200 & 1800 & 3600 & 7200 & 10800 \\
\midrule
Flash Attn      & 81.0 & 96.7 & 96.0 & 96.0 & 94.7 & 94.0 & 88.0 & 82.0 & 68.7 & 52.7 & 41.3 \\
Flash Attn + SFT& 76.8 & 96.7 & 96.7 & 96.0 & 95.3 & 90.7 & 78.0 & 78.0 & 54.7 & 41.3 & 40.7 \\
AWQ             & --   & --   & --   & --   & --   & --   & --   & --   & --   & --   & --   \\
XAttn           & 83.1 & 94.0 & 93.4 & 93.4 & 92.0 & 92.7 & 89.4 & 82.7 & 72.7 & 70.7 & 50.7 \\
MInference      & 82.7 & 93.4 & 94.0 & 93.4 & 92.0 & 92.7 & 87.4 & 80.0 & 74.0 & 70.0 & 50.0 \\
tri-shape       & 82.7 & 93.4 & 94.0 & 93.4 & 92.0 & 92.7 & 87.4 & 80.0 & 74.0 & 70.0 & 50.0 \\
FlexPrefill     & 83.0 & 96.7 & 96.0 & 96.7 & 95.3 & 96.0 & 95.3 & 86.0 & 77.3 & 55.3 & 35.3 \\
FastV           & 46.5 & 82.7 & 76.0 & 74.0 & 54.0 & 32.7 & 32.7 & 29.3 & 29.3 & 34.0 & 20.0 \\
VisionZip       & 43.5 & 92.0 & 66.7 & 60.0 & 43.3 & 35.3 & 26.0 & 30.7 & 29.3 & 28.0 & 23.3 \\
VScan           & 80.3 & 96.7 & 96.7 & 96.0 & 93.3 & 92.7 & 89.3 & 81.3 & 60.0 & 55.3 & 41.3 \\
VideoNSA        & 83.7 & 96.7 & 96.0 & 97.4 & 92.0 & 85.4 & 91.6 & 89.3 & 73.3 & 63.3 & 52.0 \\
\bottomrule
\end{tabular}
}
\label{tab:base_time}
\end{table*}

We take LongVideoBench~\cite{wu2024longvideobench}, LongTimeScope~\cite{zohar2025apollo2}, MLVU~\cite{zhou2024mlvu}, and TimeScope~\cite{zohar2025apollo2} as representative long-video benchmarks and compare against existing token compression and sparse attention methods. As shown in Table~\ref{tab:base_longtime}, Table~\ref{tab:base_longvid}, Table~\ref{tab:base_mlvu}, and Table~\ref{tab:base_time},~\model~achieves comparable performance without specialized designs. Moreover, we observe that~\model~significantly outperforms the baselines on subtasks related to temporal reasoning and on videos of extended length.
\section{More Results on Temporal Reasoning Benchmarks}
\label{base_reasoning_results}

\setlength{\tabcolsep}{3pt}

\begin{table*}[ht]
\centering
\caption{Tomato results across baselines. Metrics include overall accuracy and task-specific scores across different steps. Flash Attn stands for Qwen2.5-VL-7B~\cite{qwen2025qwen25technicalreport} accelerated by Flash Infer, and Flash Attn + SFT stands for our fine-tuning version.}
\resizebox{0.98\textwidth}{!}{
\begin{tabular}{lcccccccccc}
\toprule
Method & Overall & Direction & Count & Rotation & Shape \& Trend & Vel. \& Freq. & Visual Cues & Human & Simulated & Object \\
\midrule
Flash Attn      & 22.6 & 23.6 & 23.3 & 16.1 & 22.9 & 21.9 & 42.9 & 18.0 & 19.7 & 27.9 \\
Flash Attn + SFT& 21.7 & 19.6 & 23.3 & 18.2 & 26.0 & 18.1 & 38.6 & 18.8 & 18.0 & 25.6 \\
XAttn           & 21.4 & 22.1 & 22.9 & 19.6 & 17.9 & 17.1 & 42.9 & 15.5 & 21.5 & 26.8 \\
MInference      & 23.0 & 22.6 & 27.1 & 18.9 & 22.0 & 20.0 & 37.1 & 16.6 & 20.6 & 29.6 \\
FlexPrefill     & 23.7 & 23.3 & 25.0 & 22.7 & 22.0 & 21.4 & 35.7 & 17.1 & 22.7 & 29.9 \\
FastV           & 21.6 & 20.6 & 26.0 & 20.3 & 23.3 & 12.7 & --   & 17.1 & 24.2 & 25.6 \\
VisionZip       & 19.1 & 17.6 & 16.8 & 21.0 & 19.3 & 19.0 & 30.0 & 14.8 & 21.5 & 22.3 \\
VScan           & 23.6 & 25.3 & 21.9 & 19.9 & 24.2 & 20.5 & 42.9 & 18.7 & 21.9 & 28.7 \\
VideoNSA & 26.5	& 21.6	& 31.5	& 22.0	& 25.6	& 23.3	& 40.0	& 21.7	& 23.6	& 29.3 \\
\bottomrule
\end{tabular}
}
\label{tab:base_tomato}
\end{table*}

We take Tomato~\cite{shangguan2024tomato} as the representative temporal reasoning benchmark and compare against existing token compression and sparse attention methods. As shown in Table~\ref{tab:base_tomato},~\model achieves comparable performance without specialized designs. Moreover, we observe that~\model~significantly outperforms the baselines on subtasks including object counting, shape description, and human actions.

\section{More Results on Spatial Understanding Benchmarks}
\label{base_spatial_results}

We take VSIBench~\cite{yang2025thinkingspacemultimodallarge} as the representative spatial understanding benchmark and compare against existing token compression and sparse attention methods. As shown in Table~\ref{tab:base_vsi},~\model~achieves comparable performance without specialized designs. Moreover, we observe that~\model~significantly outperforms the baselines on subtasks including object relative direction, route planning, and object size estimation.

\setlength{\tabcolsep}{3pt}

\begin{table*}[ht]
\centering
\caption{VSIBench results across baselines. Metrics include overall accuracy and task-specific scores across different steps. Flash Attn stands for Qwen2.5-VL-7B~\cite{qwen2025qwen25technicalreport} accelerated by Flash Infer, and Flash Attn + SFT stands for our fine-tuning version.}
\resizebox{0.98\textwidth}{!}{
\begin{tabular}{lccccccccc}
\toprule
Method & Overall & Obj. Order & Abs. Dist. & Counting & Rel. Dist. & Size Est. & Room Est. & Route Plan. & Rel. Dir. \\
\midrule
Flash Attn      & 29.7 & 25.7 & 16.0 & 20.5 & 34.7 & 49.5 & 22.5 & 30.4 & 38.5 \\
Flash Attn + SFT& 30.6 & 31.9 & 14.2 & 12.3 & 40.4 & 46.6 & 30.4 & 30.9 & 37.8 \\
AWQ             & --   & --   & --   & --   & --   & --   & --   & --   & --   \\
XAttn           & 35.0 & 32.7 & 18.1 & 39.7 & 37.6 & 52.1 & 30.0 & 32.5 & 37.4 \\
MInference      & 36.6 & 36.5 & 18.2 & 43.9 & 39.4 & 48.5 & 38.8 & 30.0 & 37.7 \\
tri-shape       & 36.5 & 35.7 & 18.2 & 44.3 & 39.8 & 48.6 & 38.8 & 29.0 & 37.7 \\
FlexPrefill     & 34.9 & 34.1 & 21.6 & 35.1 & 39.3 & 51.8 & 29.7 & 30.4 & 36.8 \\
FastV           & 34.0 & 31.7 & 21.7 & 26.1 & 36.2 & 47.8 & 35.0 & 33.5 & 40.1 \\
VisionZip       & 32.1 & 28.8 & 17.9 & 28.8 & 36.5 & 48.9 & 26.9 & 29.4 & 39.3 \\
VScan           & 34.4 & 33.0 & 21.9 & 33.0 & 40.0 & 51.9 & 28.5 & 30.4 & 36.6 \\
VideoNSA & 36.0 &	25.5	& 19.0	& 42.5	& 35.4	& 54.0	& 30.1	& 37.5	& 43.6 \\
\bottomrule
\end{tabular}
}
\label{tab:base_vsi}
\end{table*}

\section{Results on Additional Video Understanding Benchmarks}
\label{add_ben}
\label{sec:add_ben}

We conduct additional experiments on LSDBench~\cite{qu2025does} and VideoEvalPro~\cite{ma2025videoeval} to compare~\model~and other training-free sparse attention baselines, demonstrating the consistent advantage of~\model~in multiple video understanding tasks.




\begin{table*}[t]
\centering

\begin{minipage}{0.58\linewidth}
\renewcommand{\arraystretch}{1.05}
\caption{Results on LSDBench~\cite{qu2025does}.}
\resizebox{\linewidth}{!}{%

\begin{tabular}{lc}
\toprule
Model & Accuracy \\
\midrule
LongVA~\cite{zhang2024long} & 32.5 \\
LongVila~\cite{chen2024longvila} & 49.8 \\
InternVL2.5~\cite{chen2024expanding} & 50.1  \\
\midrule
Qwen2.5-VL-7B~\cite{qwen2025qwen25technicalreport} & 52.2 \\
Qwen2.5-VL-7B-SFT &  52.5 \\
\multicolumn{2}{l}{\textit{Sparse Attention Methods}} \\ 
+ Tri-Shape~\cite{li2024scbench} &  49.5 \\
+ MInference~\cite{jiang2024minference} & 49.5 \\
+ FlexPrefill~\cite{lai2025flexprefill} & 52.3 \\
+ XAttention~\cite{xu2025xattention}& 51.3 \\
\midrule
\textbf{\model} & \textbf{55.2} \\ 
\bottomrule
\end{tabular}
}
\label{tab:app_lsd} 
\end{minipage}


\begin{minipage}{0.96\linewidth}
\renewcommand{\arraystretch}{1.05}
\caption{Results on VideoEvalPro~\cite{ma2025videoeval}. HP stands for Holistic Perception, HR stands for Holistic Reasoning, LR stands for Local Reasoning, LP stands for Local Perception.}
\addtolength\tabcolsep{+2.4pt} 
\resizebox{\linewidth}{!}{%
\begin{tabular}{lccccc}
\toprule
Model & HP & HR & LR & LP & Overall \\
\midrule
LongVA~\cite{zhang2024long} & 20.5 & 6.8 & 19.0 & 9.5 & 16.5 \\
Video-XL~\cite{shu2025video} & 22.3 & 15.0 & 18.2 & 10.2 & 18.6 \\
InternVL2.5~\cite{chen2024expanding} &  28.8 & 19.7 & 21.5 & 16.7 & 24.6 \\
\midrule
Qwen2.5-VL-7B~\cite{qwen2025qwen25technicalreport} &  33.9 & 15.6 & 24.8 & 17.8 & 27.7 \\
Qwen2.5-VL-7B-SFT & 34.5 & 15.8 & 25.3 & 18.2 & 28.3  \\
\multicolumn{2}{l}{\textit{Sparse Attention Methods}} \\ 
+ Tri-Shape~\cite{li2024scbench} & 34.1 & 16.3 & 25.1 & 20.0 & 28.4 \\
+ MInference~\cite{jiang2024minference} & 32.3 & \textbf{17.1} & \textbf{27.7} & 16.7 & 26.0 \\
+ FlexPrefill~\cite{lai2025flexprefill} & 33.0 & 15.9 & 26.3 & 19.8 & 28.3 \\
+ XAttention~\cite{xu2025xattention} & 34.5 & 16.6 & 25.6 & \textbf{20.5} & 28.9 \\
\midrule
\textbf{\model} & \textbf{35.4} & 16.9 & 26.3 & 19.1 & \textbf{29.4}  \\ 
\bottomrule
\end{tabular}
}
\label{tab:app_xxxx} 
\end{minipage}

\end{table*}

\arrayrulecolor{black}
\color{black}

\section{Visualization of Attention Pattern in Each Branch}
\label{vis_attn_branch}

\begin{figure}[t]
    \centering
    \includegraphics[width=\linewidth]{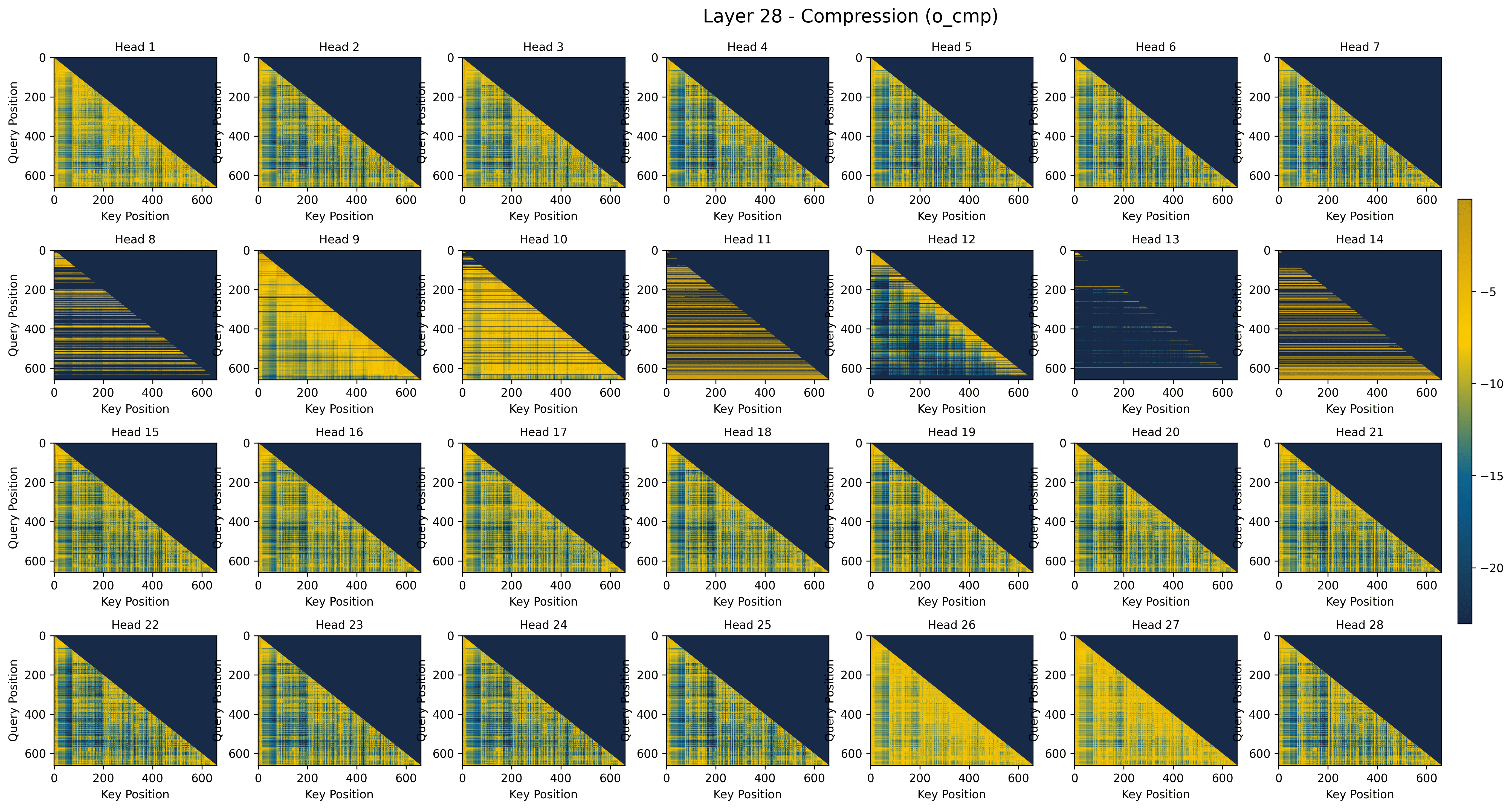}
    \caption{Attention pattern of the compression branch in the final layer of~\model.}
    \label{fig:cmp_28}
\end{figure}

\begin{figure}[t]
    \centering
    \includegraphics[width=\linewidth]{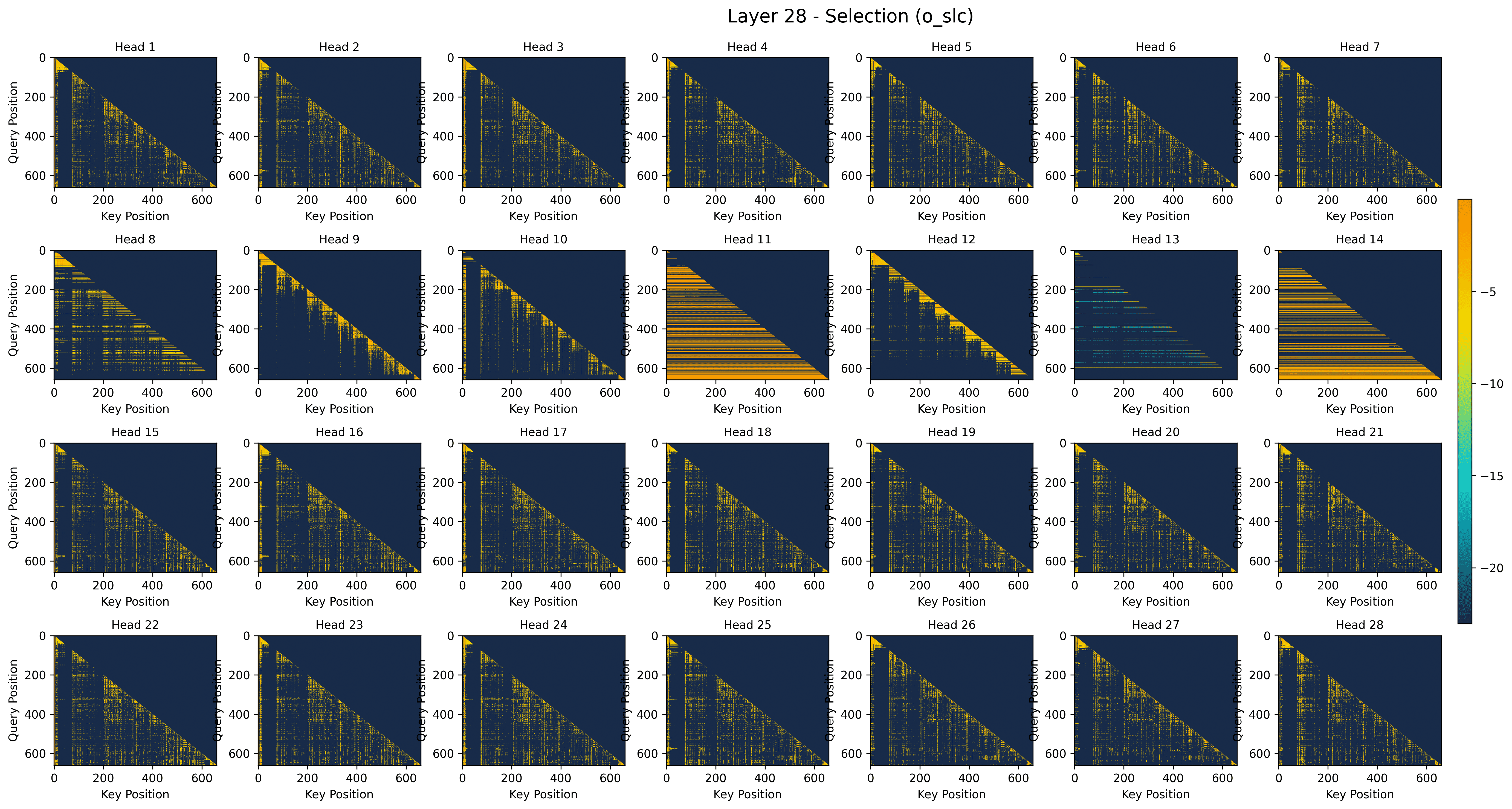}
    \caption{Attention pattern of the selection branch in the final layer of~\model.}
    \label{fig:slc_28}
\end{figure}

\begin{figure}[t]
    \centering
    \includegraphics[width=\linewidth]{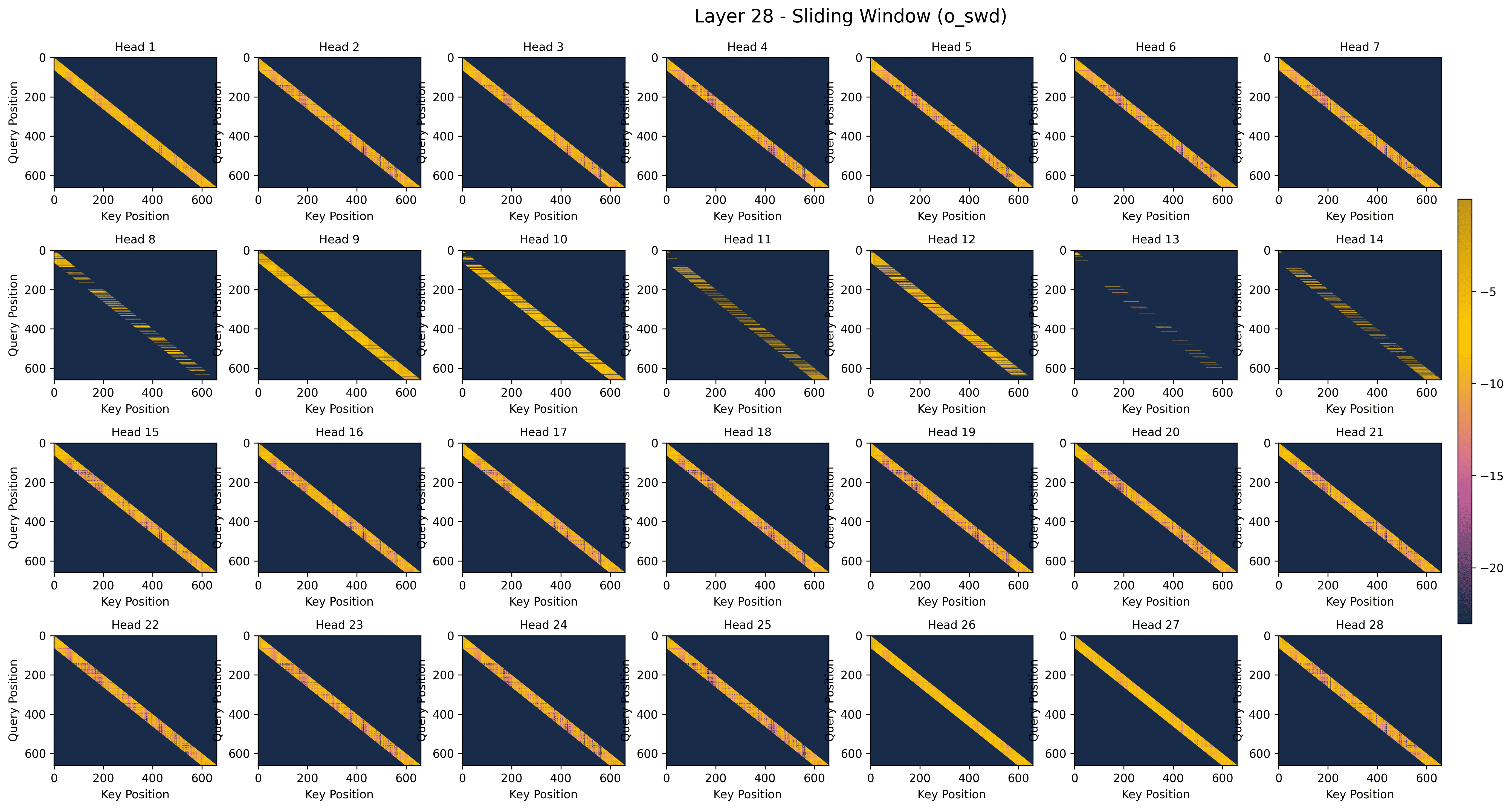}
    \caption{Attention pattern of the sliding window branch in the final layer of~\model.}
    \label{fig:swa_28}
\end{figure}

\begin{figure}[t]
    \centering
    \includegraphics[width=\linewidth]{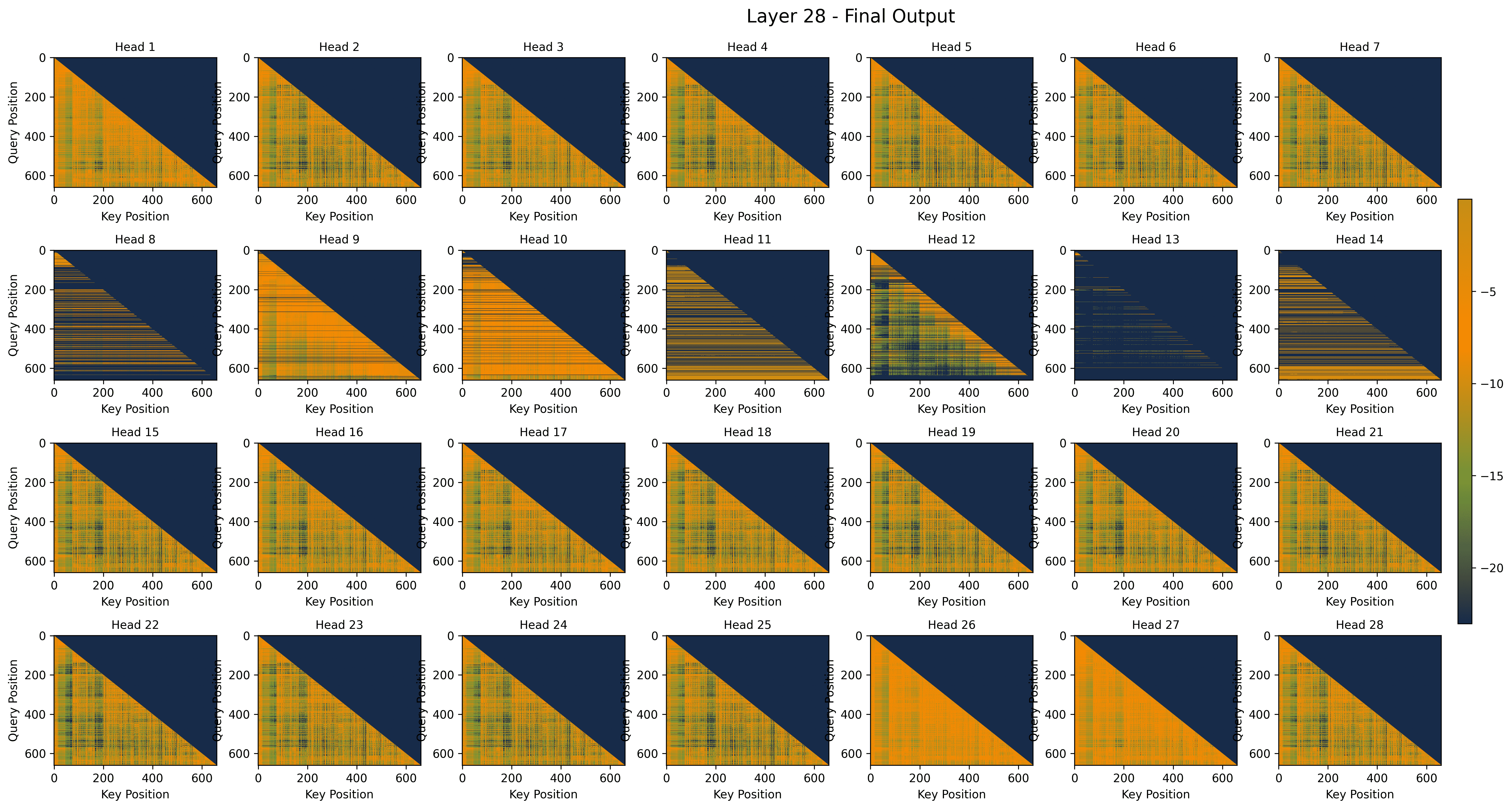}
    \caption{Attention pattern of the final vision attention output in the final layer of~\model.}
    \label{fig:final_28}
\end{figure}

We visualize the attention patterns of the last layer across the three branches in Figure~\ref{fig:cmp_28}, Figure~\ref{fig:slc_28}, Figure~\ref{fig:swa_28}, and Figure~\ref{fig:final_28}, together with the final attention output, as representative examples. The compression branch reduces redundancy to preserve salient information, the selection branch highlights task-relevant regions with sparse activations, and the sliding window branch enforces local temporal coverage by focusing on short-range dependencies. These complementary roles collectively shape the final attention output.





\section{More Results on Branch Combination}
\label{branch_comb}

In this section, we report detailed results of different branch combinations across three domains, including long video understanding (Table~\ref{tab:branch_longvid}, Tavke~\ref{tab:branch_longtime}, Table~\ref{tab:branch_time}, and Table~\ref{tab:branch_mlvu}), temporal reasoning (Table~\ref{tab:branch_tomato}), and spatial understanding (Table~\ref{tab:branch_vsi}). The corresponding performances are summarized in the table, which highlights how the use of individual branches or their combinations affects downstream tasks.

\begin{table*}[ht]
\centering
\caption{LongVideoBranch results across different branch selection strategy. Metrics include overall accuracy and task-specific scores across different steps.}
\resizebox{0.98\textwidth}{!}{
\begin{tabular}{lcccccccccccccccccccccc}
\toprule
Method & Overall & 600 & TOS & S2E & E3E & S2A & SAA & O3O & T3O & T3E & O2E & T2O & S2O & TAA & T2E & E2O & SSS & T2A & 60 & SOS & 15 & 3600 \\
\midrule
\model~ + Test SFT  & 56.1 & 57.0 & 46.6 & 59.1 & 61.7 & 69.3 & 56.9 & 63.6 & 52.7 & 50.7 & 56.3 & 59.2 & 59.7 & 43.9 & 55.4 & 64.6 & 38.1 & 58.2 & 70.4 & 60.5 & 65.1 & 48.1 \\
NSA-CMP          & 48.1 & 50.5 & 38.4 & 51.6 & 56.4 & 53.4 & 50.0 & 45.5 & 51.4 & 41.1 & 54.0 & 42.1 & 43.1 & 45.1 & 47.7 & 53.9 & 26.8 & 53.2 & 55.2 & 64.2 & 47.6 & 44.3 \\
NSA-SLC          &48.4 & 49.0 & 32.9 & 61.3 & 59.6 & 58.0 & 52.8 & 48.5 & 46.0 & 43.8 & 52.9 & 36.8 & 47.2 & 42.7 & 44.6 & 55.4 & 33.0 & 48.1 & 53.5 & 55.6 & 50.3 & 45.7 \\
NSA-SWA          &49.1 & 50.7 & 37.0 & 52.7 & 56.4 & 59.1 & 51.4 & 48.5 & 43.2 & 45.2 & 55.2 & 42.1 & 48.6 & 45.1 & 46.2 & 61.5 & 30.9 & 45.6 & 54.1 & 65.4 & 48.7 & 46.5 \\
NSA-CMPSLC       &49.4 & 49.5 & 34.3 & 55.9 & 61.7 & 58.0 & 56.9 & 48.5 & 47.3 & 41.1 & 56.3 & 35.5 & 52.8 & 47.6 & 46.2 & 55.4 & 34.0 & 41.8 & 54.1 & 63.0 & 48.2 & 48.2 \\
NSA-SLCSWA       & 49.3 & 48.8 & 32.9 & 58.1 & 61.7 & 55.7 & 52.8 & 47.0 & 46.0 & 46.6 & 54.0 & 34.2 & 48.6 & 47.6 & 47.7 & 54.0 & 35.1 & 48.1 & 54.1 & 64.2 & 49.2 & 48.2 \\
NSA-CMPSWA       & 48.8 & 49.3 & 34.3 & 53.8 & 59.6 & 54.6 & 52.8 & 50.0 & 48.7 & 42.5 & 57.5 & 40.8 & 51.4 & 42.7 & 46.2 & 55.4 & 29.9 & 45.6 & 57.6 & 64.2 & 48.7 & 45.9 \\

\bottomrule
\end{tabular}
}
\label{tab:branch_longvid}
\end{table*}

\setlength{\tabcolsep}{3pt}
\begin{table*}[ht]
\centering
\caption{LongTimeScope results across different branch selection strategy. Metrics include overall accuracy and task-specific scores across different steps.}
\resizebox{0.98\textwidth}{!}{
\begin{tabular}{lcccccccccc}
\toprule
\multirow{2}{*}{Method} & \multirow{2}{*}{Overall} 
    & \multicolumn{3}{c}{18000} 
    & \multicolumn{3}{c}{28800} 
    & \multicolumn{3}{c}{36000} \\
\cmidrule(lr){3-5}\cmidrule(lr){6-8}\cmidrule(lr){9-11}
 &  & OCR & QA & Temporal & OCR & QA & Temporal & OCR & QA & Temporal \\
\midrule
\model~ + Test SFT  & 40.9 & 52.0 & 42.0 & 42.0 & 48.0 & 62.0 & 18.0 & 42.0 & 50.0 & 12.0 \\
NSA-CMP         & 25.1 & 20.0 & 22.0 & 24.0 & 38.0 & 40.0 & 0.0  & 34.0 & 34.0 & 14.0 \\
NSA-SLC         & 37.1 & 30.0 & 38.0 & 40.0 & 50.0 & 58.0 & 12.0 & 42.0 & 44.0 & 20.0 \\
NSA-SWA         & 29.8 & 34.0 & 34.0 & 22.0 & 36.0 & 46.0 & 4.0  & 34.0 & 46.0 & 12.0 \\
NSA-CMPSLC      & 32.4 & 36.0 & 34.0 & 24.0 & 46.0 & 54.0 & 8.0  & 42.0 & 36.0 & 12.0 \\
NSA-SLCSWA      & 34.4 & 38.0 & 36.0 & 36.0 & 46.0 & 56.0 & 8.0  & 38.0 & 36.0 & 16.0 \\
NSA-CMPSWA      & 31.6 & 30.0 & 38.0 & 20.0 & 40.0 & 52.0 & 16.0 & 36.0 & 36.0 & 16.0 \\
VideoNSA        & 44.4 & 50.0 & 54.0 & 30.0 & 54.0 & 72.0 & 0.0  & 48.0 & 76.0 & 16.0 \\
\bottomrule
\end{tabular}
}
\label{tab:branch_longtime}
\end{table*}

\begin{table*}[ht]
\centering
\caption{TimeScope results across different branch selection strategy. Metrics include overall accuracy and task-specific scores across different steps.}
\resizebox{0.98\textwidth}{!}{
\begin{tabular}{lccccccccccc}
\toprule
Method & Overall & 60 & 120 & 180 & 300 & 600 & 1200 & 1800 & 3600 & 7200 & 10800 \\ 

\midrule
Full Attn       & 81.0 & 96.7 & 96.0 & 96.0 & 94.7 & 94.0 & 88.0 & 82.0 & 68.7 & 52.7 & 41.3 \\
Flash Attn      & 81.0 & 96.7 & 96.0 & 96.0 & 94.7 & 94.0 & 88.0 & 82.0 & 68.7 & 52.7 & 41.3 \\
Flash Attn + SFT& 76.8 & 96.7 & 96.7 & 96.0 & 95.3 & 90.7 & 78.0 & 78.0 & 54.7 & 41.3 & 40.7 \\
AWQ             & --   & --   & --   & --   & --   & --   & --   & --   & --   & --   & --   \\
XAttn           & 83.1 & 94.0 & 93.4 & 93.4 & 92.0 & 92.7 & 89.4 & 82.7 & 72.7 & 70.7 & 50.7 \\
MInference      & 82.7 & 93.4 & 94.0 & 93.4 & 92.0 & 92.7 & 87.4 & 80.0 & 74.0 & 70.0 & 50.0 \\
tri-shape       & 82.7 & 93.4 & 94.0 & 93.4 & 92.0 & 92.7 & 87.4 & 80.0 & 74.0 & 70.0 & 50.0 \\
FlexPrefill     & 83.0 & 96.7 & 96.0 & 96.7 & 95.3 & 96.0 & 95.3 & 86.0 & 77.3 & 55.3 & 35.3 \\
FastV           & 46.5 & 82.7 & 76.0 & 74.0 & 54.0 & 32.7 & 32.7 & 29.3 & 29.3 & 34.0 & 20.0 \\
VisionZip       & 43.5 & 92.0 & 66.7 & 60.0 & 43.3 & 35.3 & 26.0 & 30.7 & 29.3 & 28.0 & 23.3 \\
VScan           & 80.3 & 96.7 & 96.7 & 96.0 & 93.3 & 92.7 & 89.3 & 81.3 & 60.0 & 55.3 & 41.3 \\
Retake          & --   & --   & --   & --   & --   & --   & --   & --   & --   & --   & --   \\
AdaRetake       & --   & --   & --   & --   & --   & --   & --   & --   & --   & --   & --   \\
SFT + Test NSA  & 81.0 & 96.7 & 96.0 & 96.0 & 94.7 & 94.0 & 88.0 & 82.0 & 68.7 & 52.7 & 41.3 \\
NSA + Test SFT  & 83.0 & 96.7 & 95.3 & 94.0 & 93.3 & 94.0 & 90.7 & 87.3 & 76.7 & 54.7 & 47.3 \\
NSA-CMP         & 41.5 & 82.0 & 74.0 & 65.3 & 59.3 & 17.3 & 25.3 & 19.3 & 26.7 & 27.3 & 18.0 \\
NSA-SLC         & 63.7 & 92.0 & 86.0 & 86.7 & 78.0 & 66.7 & 57.3 & 51.3 & 40.7 & 38.0 & 40.0 \\
NSA-SWA         & 59.3 & --   & --   & --   & --   & --   & --   & --   & --   & --   & --   \\
NSA-CMPSLC      & 57.3 & 88.7 & 80.0 & 73.3 & 73.3 & 46.7 & 44.7 & 48.7 & 42.7 & 43.3 & 32.0 \\
NSA-SLCSWA      & 65.2 & 92.0 & 89.3 & 89.3 & 79.3 & 66.0 & 59.3 & 50.0 & 41.3 & 40.7 & 44.7 \\
NSA-CMPSWA      & 57.3 & 88.7 & 80.0 & 73.3 & 73.3 & 46.7 & 44.7 & 48.7 & 42.7 & 43.3 & 32.0 \\
VideoNSA        & 83.7 & 96.7 & 96.0 & 97.4 & 92.0 & 85.4 & 91.6 & 89.3 & 73.3 & 63.3 & 52.0 \\
\bottomrule
\end{tabular}
}
\label{tab:branch_time}
\end{table*}
\begin{table*}[ht]
\centering
\caption{MLVU results across different branch selection strategy. Metrics include overall accuracy and task-specific scores across different steps.}
\resizebox{0.98\textwidth}{!}{
\begin{tabular}{lcccccccccc}
\toprule
Method & Overall & PlotQA & Needle & Ego & Count & Order & Anomaly Reco & Topic Reason. & SportsQA & TutorialQA \\
\midrule
NSA + Test SFT   & 51.6 & 56.0 & 61.7 & 66.0 & 31.7 & 28.6 & 51.3 & 80.2 & 36.1 & 32.6 \\
NSA-CMP          & 43.9 & 36.0 & 35.0 & 42.9 & --   & 24.3 & 30.8 & 80.2 & 30.6 & --   \\
NSA-SLC          & 47.7 & 50.0 & 50.0 & 52.4 & --   & 22.9 & 33.3 & 74.7 & 33.3 & --   \\
NSA-SWA          & 40.2 & 40.0 & 40.0 & 41.5 & 15.0 & 24.3 & 30.8 & 76.9 & 36.1 & 34.9 \\
NSA-SLCSWA       & 42.4 & 42.0 & 48.3 & 45.3 & 16.7 & 25.7 & 38.5 & 75.8 & 33.3 & 34.9 \\
NSA-CMPSWA       & 43.4 & 46.0 & 40.0 & 43.4 & 18.3 & 35.7 & 33.3 & 82.4 & 27.8 & 32.6 \\
\bottomrule
\end{tabular}
}
\label{tab:branch_mlvu}
\end{table*}

\begin{table*}[ht]
\centering
\caption{Tomato results across different branch selection strategy. Metrics include overall accuracy and task-specific scores across different steps.}
\resizebox{0.98\textwidth}{!}{
\begin{tabular}{lcccccccccc}
\toprule
Method & Overall & Direction & Count & Rotation & Shape \& Trend & Vel. \& Freq. & Visual Cues & Human & Simulated & Object \\
\midrule
NSA + Test SFT  & 23.4 & 21.3 & 29.1 & 17.5 & 25.1 & 20.0 & 40.0 & 19.3 & 19.3 & 28.5 \\
NSA-CMP         & 23.3 & 22.1 & 29.5 & 17.1 & 24.7 & 20.5 & 34.3 & 19.2 & 22.7 & 27.3 \\
NSA-SLC         & 24.0 & 21.3 & 32.2 & 16.4 & 26.0 & 22.9 & 32.9 & 19.8 & 21.5 & 28.7 \\
NSA-SWA         & 24.0 & 21.3 & 32.2 & 16.4 & 26.0 & 22.9 & 32.9 & 19.8 & 21.5 & 28.7 \\
NSA-CMPSLC      & 23.5 & 20.8 & 29.8 & 18.5 & 22.9 & 23.8 & 34.3 & 19.0 & 26.8 & 25.8 \\
NSA-SLCSWA      & 23.0 & 20.6 & 27.4 & 18.5 & 22.4 & 24.8 & 32.9 & 19.5 & 21.0 & 26.8 \\
NSA-CMPSWA      & 24.5 & 23.1 & 30.8 & 18.9 & 25.1 & 22.9 & 32.9 & 21.0 & 23.6 & 28.0 \\
\bottomrule
\end{tabular}
}
\label{tab:branch_tomato}
\end{table*}
\begin{table*}[ht]
\centering
\caption{VSIBench results across different branch selection strategy. Metrics include overall accuracy and task-specific scores across different steps.}
\resizebox{0.98\textwidth}{!}{
\begin{tabular}{lccccccccc}
\toprule
Method & Overall & Obj. Order & Abs. Dist. & Counting & Rel. Dist. & Size Est. & Room Est. & Route Plan. & Rel. Dir. \\
\midrule
NSA + Test SFT  & 33.1 & 24.3 & 19.8 & 31.2 & 38.0 & 49.8 & 32.2 & 32.5 & 37.2 \\
NSA-CMP         & 29.2 & 19.9 & 16.3 & 12.6 & 29.3 & 48.7 & 26.7 & 38.1 & 41.7 \\
NSA-SLC         & 27.6 & 18.0 & 10.9 & 17.3 & 32.0 & 47.8 & 24.8 & 32.0 & 38.1 \\
NSA-SWA         & 29.8 & 22.8 & 15.6 & 17.4 & 32.3 & 49.8 & 27.2 & 33.5 & 39.4 \\
NSA-CMPSLC      & 29.4 & 19.9 & 16.3 & 15.1 & 31.0 & 51.1 & 25.5 & 33.5 & 42.6 \\
NSA-SLCSWA      & 29.1 & 19.9 & 12.2 & 18.5 & 31.4 & 49.6 & 26.5 & 34.0 & 40.4 \\
NSA-CMPSWA      & 30.3 & 22.5 & 15.6 & 15.3 & 31.1 & 52.5 & 26.7 & 35.1 & 43.3 \\
\bottomrule
\end{tabular}
}
\label{tab:branch_vsi}
\end{table*}
\section{More results on information scaling study}
\label{information_scaling}

Figure~\ref{app:attention_budget} shows the scaling performance of~\model~under different context allocation strategies on LongTimeScope and MLVU. Both benchmarks were trained with a maximum context length of 32K tokens, yet their performance consistently improves when scaled to 64K, beyond the training budget. On LongTimeScope~\cite{zohar2025apollo2}, the best results emerge around 512 frames with 128 TPF at 64K tokens, underscoring the dataset’s reliance on extended temporal coverage for long-horizon reasoning. In contrast, MLVU~\cite{zhou2024mlvu} also peaks at 64K with the same allocation, but its contours are smoother, and competitive performance extends across a broader range of frame–token trade-offs. This suggests that while LongTimeScope demands aggressive temporal scaling, MLVU benefits from a more balanced distribution of temporal and spatial information.

\begin{figure}[!ht]
    \captionsetup{skip=2pt}
    \centering
    \begin{tabular}{ccc}
         \subfloat[Information Scaling of LongTimeScope]{
             \includegraphics[width=0.48\textwidth]{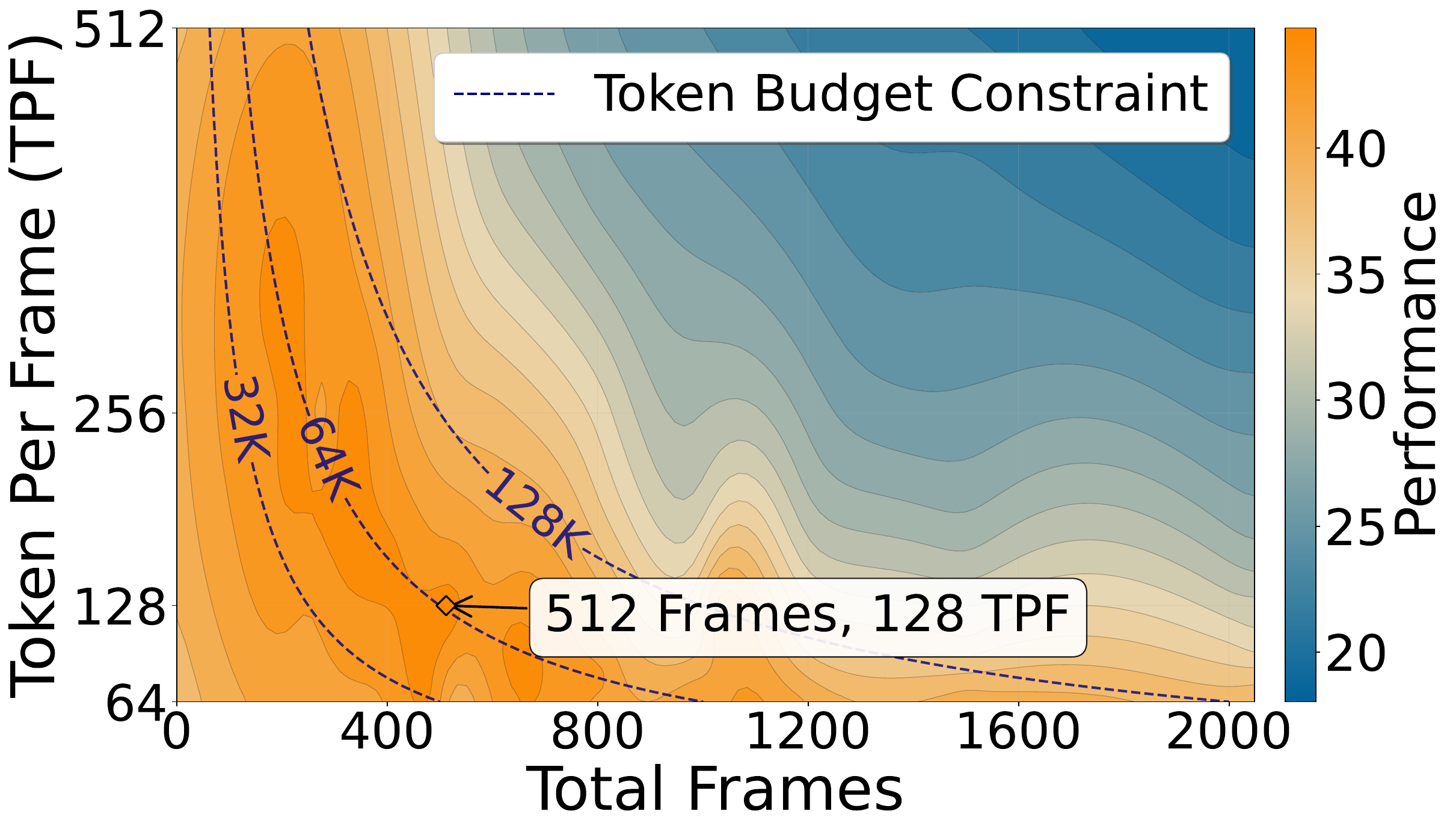}
         } &
         \subfloat[Information Scaling of MLVU]{\includegraphics[width=0.48\textwidth]{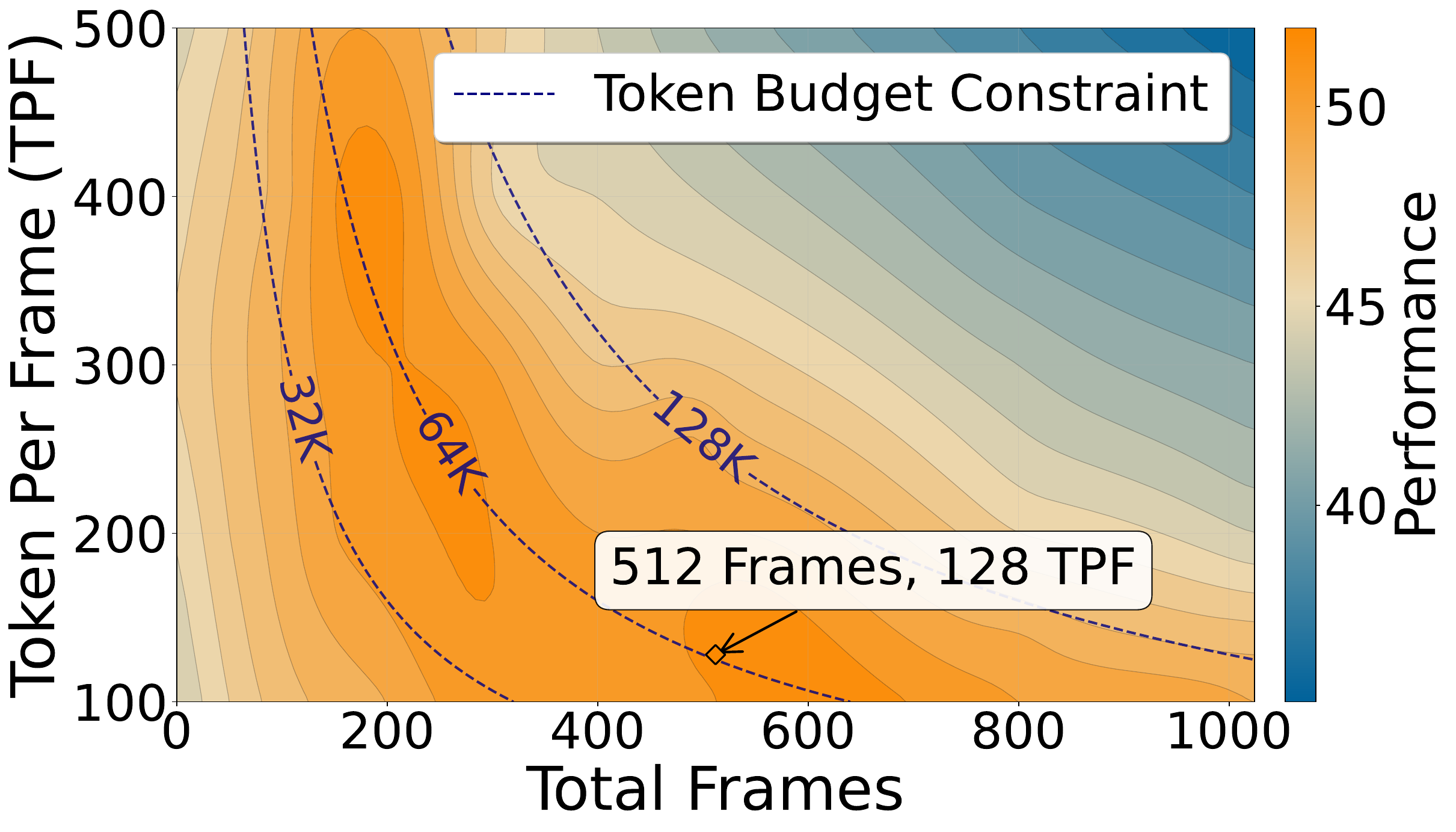}} 
         \\
    \end{tabular}
    \caption{Scaling Performance of VideoNSA under Different Context Allocation Strategies. 
We highlight the token budget constraint to indicate settings with equal context length, 
and annotate the best-performing configuration under each benchmark. }
\label{app:attention_budget}
\end{figure}

In addition to the overall scaling trends, we further report detailed subtask-level results under different allocation settings in Table~\ref{tab:frame_longtime}, Table~\ref{tab:frame_time}, Table~\ref{tab:frame_longvid}, Table~\ref{tab:frame_mlvu}, Table~\ref{tab:frame_tomato}, and Table~\ref{tab:frame_vsi}.

\setlength{\tabcolsep}{3pt}
\begin{table*}[ht]
\centering
\caption{Ablation study results on information scaling of LongTimeScope~\cite{zohar2025apollo2}. Metrics include overall accuracy and task-specific scores across different steps. \# TPF stands for token per frame, and \# F stands for sampling frame number.}
\resizebox{0.98\textwidth}{!}{
\begin{tabular}{lccccccccccc}
\toprule
\multirow{2}{*}{\# TPF} & \multirow{2}{*}{\# F} & \multirow{2}{*}{Overall} 
  & \multicolumn{3}{c}{18000} 
  & \multicolumn{3}{c}{28800} 
  & \multicolumn{3}{c}{36000} \\
\cmidrule(lr){4-6} \cmidrule(lr){7-9} \cmidrule(lr){10-12}
 & & & OCR & QA & Temporal & OCR & QA & Temporal & OCR & QA & Temporal \\
\midrule
256 & 128 & 42.9 & 54.0 & 48.0 & 36.0 & 46.0 & 62.0 & 6.0 & 40.0 & 80.0 & 14.0 \\
512 & 128 & 41.1 & 54.0 & 60.0 & 28.0 & 42.0 & 62.0 & 4.0 & 40.0 & 78.0 & 2.0 \\
128 & 256 & 42.0 & 58.0 & 56.0 & 26.0 & 46.0 & 62.0 & 2.0 & 40.0 & 78.0 & 10.0 \\
256 & 256 & 41.3 & 58.0 & 52.0 & 36.0 & 48.0 & 62.0 & 0.0 & 40.0 & 70.0 & 6.0 \\
512 & 256 & 41.6 & 54.0 & 56.0 & 32.0 & 46.0 & 60.0 & 2.0 & 40.0 & 78.0 & 6.0 \\
64  & 512 & 40.2 & 52.0 & 52.0 & 26.0 & 44.0 & 64.0 & 2.0 & 44.0 & 76.0 & 2.0 \\
128 & 512 & 44.4 & 50.0 & 54.0 & 30.0 & 54.0 & 72.0 & 0.0 & 48.0 & 76.0 & 16.0 \\
256 & 512 & 38.7 & 48.0 & 50.0 & 30.0 & 52.0 & 56.0 & 8.0 & 36.0 & 60.0 & 8.0 \\
64  & 1024 & 41.6 & 54.0 & 56.0 & 22.0 & 46.0 & 66.0 & 4.0 & 36.0 & 72.0 & 18.0 \\
128 & 1024 & 41.1 & 50.0 & 46.0 & 32.0 & 46.0 & 62.0 & 14.0 & 38.0 & 54.0 & 28.0 \\
64  & 2048 & 38.4 & 50.0 & 62.0 & 26.0 & 40.0 & 60.0 & 2.0 & 38.0 & 42.0 & 26.0 \\
\bottomrule
\end{tabular}
}
\label{tab:frame_longtime}
\end{table*}

\begin{table*}[ht]
\centering
\caption{Ablation study results on information scaling of TimeScope~\cite{zohar2025apollo2}. Metrics include overall accuracy and task-specific scores across different steps. \# TPF stands for token per frame, and \# F stands for sampling frame number.}
\resizebox{0.98\textwidth}{!}{
\begin{tabular}{lcccccccccccc}
\toprule
\# TPF & \# F & Overall & 60 & 120 & 180 & 300 & 600 & 1200 & 1800 & 3600 & 7200 & 10800 \\ 
\midrule
256 & 128 & 73.1 & 96.7 & 94.7 & 93.4 & 85.4 & 72.0 & 62.6 & 57.3 & 56.6 & 54.0 & 58.6 \\
512 & 128 & 72.5 & 95.4 & 94.0 & 92.7 & 82.0 & 72.7 & 64.6 & 57.3 & 56.3 & 53.3 & 56.6 \\
128 & 256 & 76.5 & 98.0 & 97.4 & 96.0 & 86.7 & 78.7 & 78.0 & 63.3 & 58.6 & 54.0 & 54.0 \\
256 & 256 & 76.1 & 96.7 & 96.7 & 91.4 & 86.7 & 76.0 & 74.6 & 64.0 & 62.0 & 56.0 & 56.6 \\
512 & 256 & 75.8 & 95.4 & 94.7 & 90.7 & 86.7 & 76.0 & 75.3 & 66.0 & 62.6 & 53.3 & 57.3 \\
64  & 512 & 78.5 & 96.7 & 95.4 & 94.7 & 88.0 & 80.7 & 82.6 & 71.3 & 62.0 & 58.0 & 55.3 \\
128 & 512 & 76.5 & 98.0 & 97.4 & 96.0 & 86.7 & 78.7 & 78.0 & 63.3 & 58.6 & 54.0 & 54.0 \\
256 & 512 & 77.3 & 96.7 & 96.7 & 90.7 & 83.4 & 76.7 & 78.0 & 72.0 & 66.6 & 59.3 & 52.6 \\
64  & 1024 & 81.7 & 96.7 & 95.4 & 94.7 & 90.0 & 84.7 & 88.0 & 78.0 & 69.3 & 64.0 & 56.6 \\
128 & 1024 & 81.8 & 98.0 & 97.4 & 94.0 & 85.4 & 80.7 & 92.0 & 78.0 & 72.6 & 66.0 & 54.0 \\
64  & 2048 & 82.7 & 96.7 & 95.4 & 94.7 & 90.0 & 82.0 & 91.3 & 88.0 & 73.3 & 63.3 & 52.0 \\
\bottomrule
\end{tabular}
}
\label{tab:frame_time}
\end{table*}
\begin{table*}[ht]
\centering
\caption{Ablation study results on information scaling of LongVideoBench~\cite{wu2024longvideobench}. Metrics include overall accuracy and task-specific scores across different steps. \# TPF stands for token per frame, and \# F stands for sampling frame number.}
\resizebox{0.98\textwidth}{!}{
\begin{tabular}{cc|cccccccccccccccccccccc}
\toprule
\# TPF & \# F & Overall & 600.0 & TOS & S2E & E3E & S2A & SAA & O3O & T3O & T3E & O2E & T2O & S2O & TAA & T2E & E2O & SSS & T2A & 60.0 & SOS & 15.0 & 3600.0 \\
\midrule
512 & 64   & 58.3 & 55.5 & 44.0 & 67.2 & 66.6 & 71.8 & 57.0 & 54.0 & 48.9 & 50.8 & 59.9 & 55.7 & 63.9 & 53.3 & 57.8 & 70.1 & 37.9 & 60.2 & 68.1 & 71.2 & 65.7 & 54.9 \\
128 & 128  & 57.4 & 56.8 & 45.4 & 66.1 & 66.6 & 69.5 & 61.2 & 55.5 & 53.0 & 49.5 & 59.9 & 51.7 & 54.2 & 47.2 & 57.8 & 68.5 & 40.0 & 58.9 & 68.6 & 70.0 & 63.6 & 52.4 \\
256 & 128  & 57.9 & 58.5 & 48.1 & 69.4 & 66.6 & 70.6 & 58.4 & 55.5 & 53.0 & 48.1 & 57.6 & 47.8 & 54.2 & 49.7 & 59.3 & 68.5 & 40.0 & 65.2 & 68.1 & 70.0 & 63.1 & 52.6 \\
512 & 128  & 59.0 & 59.4 & 49.5 & 68.3 & 66.6 & 72.9 & 63.9 & 54.0 & 53.0 & 49.5 & 61.0 & 51.7 & 61.2 & 50.9 & 56.2 & 68.5 & 39.0 & 64.0 & 68.6 & 71.2 & 63.6 & 54.2 \\
128 & 256  & 58.7 & 52.7 & 46.7 & 68.3 & 65.5 & 69.5 & 57.0 & 52.5 & 53.0 & 48.1 & 56.4 & 46.5 & 51.4 & 48.5 & 57.8 & 67.0 & 37.9 & 65.2 & 63.4 & 70.0 & 58.3 & 52.7 \\
256 & 256  & 58.2 & 58.7 & 39.9 & 64.0 & 66.6 & 71.8 & 58.4 & 54.0 & 59.7 & 52.2 & 59.9 & 55.7 & 58.4 & 50.9 & 56.2 & 70.1 & 40.0 & 61.4 & 66.9 & 68.7 & 63.1 & 53.5 \\
512 & 256  & 59.4 & 60.4 & 52.2 & 67.2 & 65.5 & 75.2 & 61.2 & 54.0 & 55.7 & 52.2 & 62.2 & 53.1 & 62.6 & 49.7 & 56.2 & 68.5 & 35.9 & 65.2 & 67.5 & 72.4 & 65.7 & 54.0 \\
64  & 512  & 57.7 & 58.2 & 41.3 & 67.2 & 68.7 & 65.0 & 58.4 & 58.5 & 54.3 & 52.2 & 62.2 & 49.1 & 58.4 & 53.3 & 62.4 & 71.6 & 35.9 & 55.1 & 66.3 & 68.7 & 61.5 & 53.5 \\
128 & 512  & 58.5 & 59.4 & 42.6 & 68.3 & 66.6 & 69.5 & 59.8 & 60.0 & 57.0 & 52.2 & 64.5 & 50.4 & 59.8 & 52.1 & 59.3 & 68.5 & 35.9 & 60.2 & 65.7 & 68.7 & 63.6 & 54.0 \\
256 & 512  & 58.3 & 59.2 & 44.0 & 64.0 & 64.5 & 71.8 & 65.3 & 52.5 & 55.7 & 55.0 & 61.0 & 54.4 & 62.6 & 49.7 & 59.3 & 71.6 & 37.9 & 56.4 & 66.9 & 67.5 & 63.1 & 53.5 \\
64  & 1024 & 58.4 & 59.4 & 42.6 & 65.1 & 68.7 & 66.1 & 62.6 & 55.5 & 58.4 & 49.5 & 64.5 & 54.4 & 58.4 & 50.9 & 59.3 & 74.7 & 36.9 & 57.6 & 66.3 & 68.7 & 61.5 & 54.2 \\
128 & 1024 & 58.7 & 58.5 & 41.3 & 67.2 & 68.7 & 71.8 & 65.3 & 60.0 & 59.7 & 52.2 & 59.9 & 53.1 & 62.6 & 47.2 & 59.3 & 71.6 & 32.8 & 60.2 & 65.7 & 67.5 & 63.6 & 55.1 \\
\bottomrule
\end{tabular}
}
\label{tab:frame_longvid}
\end{table*}
\begin{table*}[ht]
\centering
\caption{Ablation study results on information scaling of MLVU~\cite{zhou2024mlvu}. Metrics include overall accuracy and task-specific scores across different steps. \# TPF stands for token per frame, and \# F stands for sampling frame number.}
\resizebox{0.98\textwidth}{!}{
\begin{tabular}{lc|cccccccccc}
\toprule
\# TPF & \# F & Overall & PlotQA & Needle & Ego & Count & Order & Anomaly Reco & Topic Reason. & SportsQA & TutorialQA \\
\midrule
256	& 128 &	49.6 &	46.0 &	52.7 &	53.2 &	24.3 &	36.0 &	47.0 &	87.3 &	36.6 &	36.2 \\
512 &	128 &	49.2 &	52.0 &	51.0 &	47.5 &	24.3	& 37.4	& 39.3 &	87.3 &	42.1 &	33.9 \\
128 &	256	& 50.6	& 50.0	& 57.7 &	60.7 &	24.3 &	33.1 &	39.3 &	86.2 &	42.1 &	36.2 \\
256 &	256	& 51.2 &	50.0 &	56.0	& 56.9	& 27.7 &	38.9 &	41.9	& 85.1	& 42.1	& 36.2 \\
512 &	256	& 48.0	& 54.0	& 49.3	& 49.4 &	22.7 &	37.4 &	39.3	& 86.2	& 33.8	& 29.3 \\
64 &	512 &	51.2 &	50.0 &	62.7 &	55.1 &	24.3	& 34.6 &	47.0 &	84.0 &	42.1 &	38.6 \\
128 &	512	& 51.8 &	48.0 &	69.3 &	51.3 &	27.7 &	34.6	& 44.5 &	86.2 &	47.7 &	31.6 \\
256	& 512 &	48.6 &	50.0 &	51.0 &	47.5 &	24.3	& 33.1 &	52.2 &	84.0 &	47.7 &	26.9 \\
64 &	1024 &	51.8 &	56.0 &	66.0 &	53.2 &	26.0 &	36.0 &	47.0 &	84.0 &	42.1 &	31.6 \\
128 &	1024	& 48.0 &	52.0 &	51.0	& 49.4	& 29.3 &	33.1 &	44.5 &	80.7 &	44.9 &	24.6 \\
\bottomrule
\end{tabular}
}
\label{tab:frame_mlvu}
\end{table*}
\begin{table*}[ht]
\centering
\caption{Ablation study results on information scaling of Tomato~\cite{shangguan2024tomato}. Metrics include overall accuracy and task-specific scores across different steps. \# TPF stands for token per frame, and \# F stands for sampling frame number.}
\resizebox{0.98\textwidth}{!}{
\begin{tabular}{cc|ccccccccccc}
\toprule
FPS & TPF & Overall & Direction & Count & Rotation & Shape\&Trend & Velocity\&Freq. & Visual Cues & Human & Simulated & Object \\
\midrule
1 & 64  & 24.7 & 22.8 & 26.7 & 20.6 & 25.1 & 27.1 & 34.3 & 21.2 & 23.2 & 28.4 \\
1 & 128 & 23.9 & 20.6 & 29.8 & 19.9 & 24.7 & 23.8 & 32.9 & 19.8 & 24.0 & 27.6 \\
1 & 256 & 24.7 & 22.3 & 29.5 & 19.6 & 25.1 & 25.2 & 35.7 & 20.8 & 23.3 & 28.7 \\
1 & 512 & 23.9 & 20.6 & 29.8 & 19.9 & 24.7 & 22.9 & 34.3 & 20.3 & 21.5 & 27.9 \\
2 & 64  & 24.5 & 21.1 & 31.8 & 19.6 & 22.4 & 25.7 & 35.7 & 20.7 & 21.9 & 28.8 \\
2 & 128 & 24.3 & 20.6 & 30.5 & 20.3 & 24.7 & 23.3 & 38.6 & 20.7 & 22.3 & 28.4 \\
2 & 256 & 24.4 & 21.3 & 29.5 & 18.5 & 26.5 & 24.8 & 37.1 & 20.3 & 24.0 & 28.2 \\
2 & 512 & 24.7 & 19.4 & 32.2 & 21.3 & 25.6 & 23.8 & 37.1 & 20.0 & 24.0 & 29.1 \\
4 & 64  & 25.1 & 22.1 & 31.5 & 19.6 & 26.5 & 23.3 & 38.6 & 21.2 & 25.0 & 29.0 \\
4 & 128 & 25.8 & 21.8 & 33.2 & 21.3 & 25.6 & 25.7 & 37.1 & 21.5 & 25.3 & 29.9 \\
4 & 256 & 26.2 & 23.1 & 32.5 & 20.6 & 26.9 & 26.7 & 37.1 & 21.8 & 25.3 & 30.5 \\
4 & 512 & 26.5 & 21.6 & 31.5 & 22.0 & 25.6 & 23.3 & 40.0 & 21.7 & 23.6 & 29.3 \\
\bottomrule
\end{tabular}
}
\label{tab:frame_tomato}
\end{table*}
\begin{table*}[ht]
\centering
\caption{Ablation study results on information scaling of VSIBench~\cite{yang2025thinkingspacemultimodallarge}. Metrics include overall accuracy and task-specific scores across different steps. \# TPF stands for token per frame, and \# F stands for sampling frame number.}
\resizebox{0.98\textwidth}{!}{
\begin{tabular}{cc|ccccccccc}
\toprule
TPF & \# Max Frames & Overall & Obj. Order & Abs. Dist. & Counting & Rel. Dist. & Size Est. & Room Est. & Route Plan. & Rel. Dir. \\
\midrule
512 & 32   & 34.9 & 27.6 & 16.2 & 31.4 & 35.1 & 52.2 & 31.5 & 40.1 & 44.6 \\
512 & 64   & 34.8 & 29.1 & 17.5 & 34.9 & 33.0 & 52.2 & 31.0 & 36.5 & 43.9 \\
256 & 128  & 36.0 & 24.7 & 17.6 & 41.3 & 37.5 & 53.9 & 30.7 & 39.1 & 43.3 \\
512 & 128  & 34.6 & 27.4 & 17.2 & 37.3 & 34.4 & 50.3 & 30.6 & 35.5 & 43.8 \\
128 & 256  & 35.6 & 26.8 & 17.0 & 42.0 & 36.8 & 51.8 & 31.2 & 35.0 & 44.2 \\
256 & 256  & 35.5 & 27.8 & 17.0 & 42.4 & 33.7 & 51.3 & 31.6 & 35.5 & 44.5 \\
512 & 256  & 34.8 & 28.2 & 16.5 & 40.3 & 33.9 & 48.8 & 30.7 & 36.5 & 43.3 \\
64  & 512  & 34.2 & 29.1 & 15.8 & 42.1 & 33.9 & 45.5 & 27.7 & 37.0 & 42.9 \\
128 & 512  & 36.0 & 25.5 & 19.0 & 42.5 & 35.4 & 54.0 & 30.1 & 37.5 & 43.6 \\
256 & 512  & 33.9 & 28.2 & 15.8 & 42.9 & 31.3 & 43.6 & 29.9 & 36.5 & 43.1 \\
64  & 1024 & 35.8 & 24.4 & 18.5 & 46.4 & 34.4 & 52.4 & 29.7 & 37.5 & 43.0 \\
128 & 1024 & 35.7 & 26.6 & 18.4 & 45.3 & 32.7 & 50.3 & 31.7 & 37.0 & 43.7 \\
\bottomrule
\end{tabular}
}
\label{tab:frame_vsi}
\end{table*}
\section{Additional Context-length Scaling Results of Qwen2.5-VL}
\label{qwen25_context}
\label{sec:qwen25_context}

\begin{figure}[!ht]
    \captionsetup{skip=2pt}
    \centering
    \begin{tabular}{ccc}
         \subfloat[LongTimeScope]{
             \includegraphics[width=0.32\textwidth]{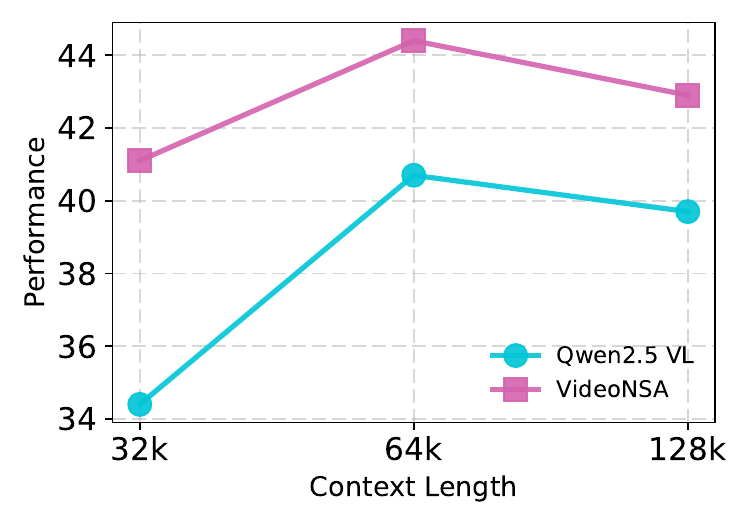}
         } &
         \subfloat[TimeScope]{\includegraphics[width=0.32\textwidth]{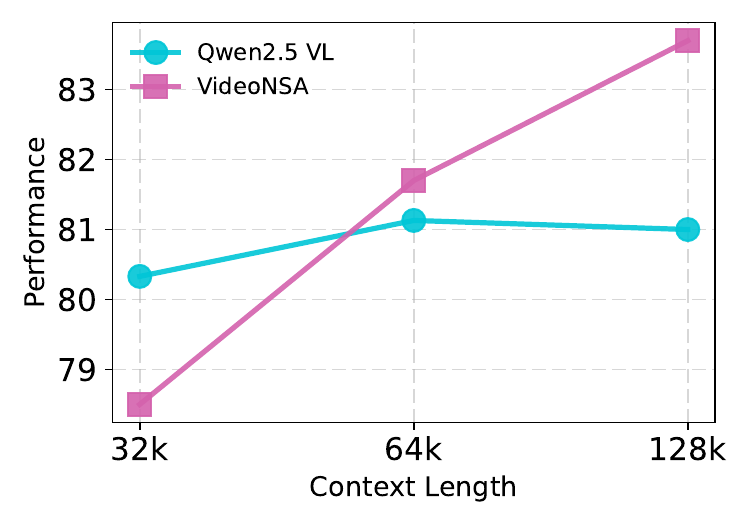}} & 
         \subfloat[LongVideoBench]{\includegraphics[width=0.32\textwidth]{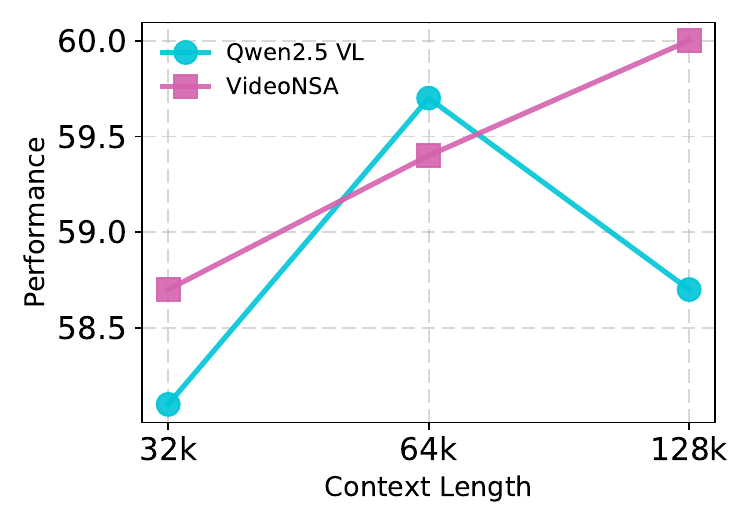}}
         \\
         \subfloat[MLVU]{\includegraphics[width=0.32\textwidth]{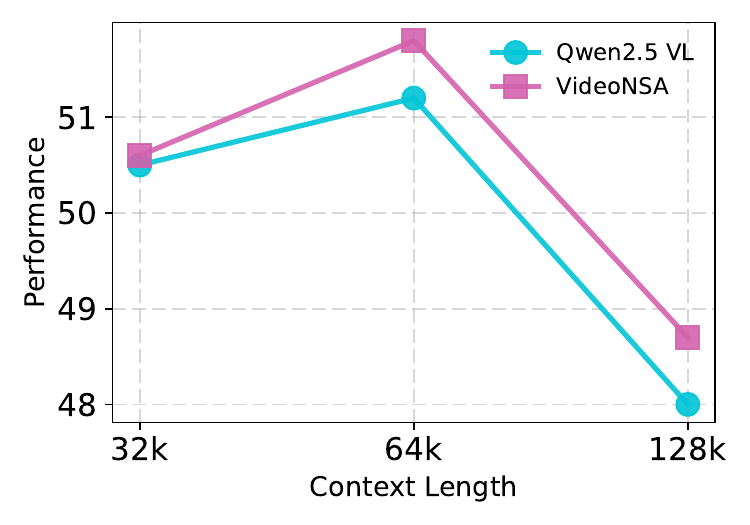}} & 
         \subfloat[VSIBench]{\includegraphics[width=0.32\textwidth]{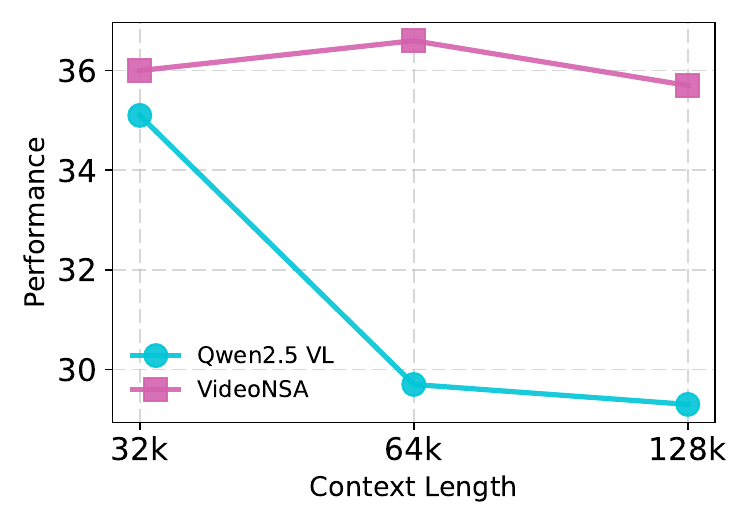}}
         \\
    \end{tabular}
    \caption{Performance comparison of Qwen2.5-VL and VideoNSA under different context lengths.}
\label{app:qwen25_context}
\end{figure}

\begin{table}[t]
\centering
\caption{Performance of Qwen2.5-VL 7B under different context lengths.}
\label{tab:qwen_long_context}
\begin{tabular}{lccccc}
\toprule
Context & LVB & MLVU & TimeScope & LTS & VSIBench\\
\midrule
32k  & 58.1 & 50.5 & 80.33 & 34.4 & 35.1 \\
64k  & 59.7 & 51.2 & 81.13 & 40.7 & 29.7 \\
128k & 58.7 & 48.0 & 81.00 & 39.7 & 29.3 \\
\bottomrule
\end{tabular}
\end{table}

\begin{table}[t]
\centering
\caption{Performance of~\model~under different context lengths.}
\label{tab:videonsa_long_context}
\begin{tabular}{lccccc}
\toprule
Context & LVB & MLVU & TimeScope & LTS & VSIBench\\
\midrule
32k  & 58.7 & 50.6 & 78.5 & 41.1 & 36.0 \\
64k  & 59.4 & 51.8 & 81.7 & 44.4 & 36.6 \\
128k & 60.0 & 48.7 & 83.7 & 42.9 & 35.7 \\
\bottomrule
\end{tabular}
\end{table}

We include Table~\ref{tab:qwen_long_context} and Table~\ref{tab:videonsa_long_context} to further illustrate the long-context behavior of the base model. Since Qwen2.5-VL 7B~\cite{qwen2025qwen25technicalreport} has a maximum context window of 128k, its modeling ability tends to become less stable when approaching this upper bound. As shown in Figure~\ref{qwen25_context}, Qwen2.5-VL~\cite{qwen2025qwen25technicalreport} often peaks at 64k and slightly declines at 128k across several benchmarks. In contrast,~\model~maintains stable or stronger performance at 128k, demonstrating that the observed $64\text{k} > 128\text{k}$ phenomenon arises from backbone limitations rather than the proposed sparse architecture.

\section{More results on attention scaling study}
\label{attention_scaling}


Figure~\ref{app:attention_budget} evaluates the scaling behavior of VideoNSA under different attention allocation strategies, where the x-axis denotes the sliding window size (log scale), the y-axis shows the block count, and the size and color of each marker reflect performance, with the dashed blue curve indicating configurations of equal attention budget and arrows marking the training setting as well as reduced-budget configurations (3.6\% and 1.8\%); on LongVideoBench, performance peaks near the training configuration and degrades when allocating excessive budget to local attention through larger sliding windows, while the best configuration achieves strong results with only 3.6\% of the full budget, and on TimeScope, performance is even more sensitive, with larger sliding windows quickly reducing accuracy whereas maintaining more global blocks yields superior outcomes, and overall the results confirm that training allocations are well balanced, that prioritizing global attention is consistently more effective than enlarging local windows under equal budget, and that VideoNSA sustains leading performance with as little as 3.6\% or less of the full attention cost, demonstrating both efficiency and hardware awareness.

\begin{figure}[!ht]
    \captionsetup{skip=2pt}
    \centering
    \begin{tabular}{ccc}
         \subfloat[Attention Scaling of LongVideoBench]{
             \includegraphics[width=0.48\textwidth]{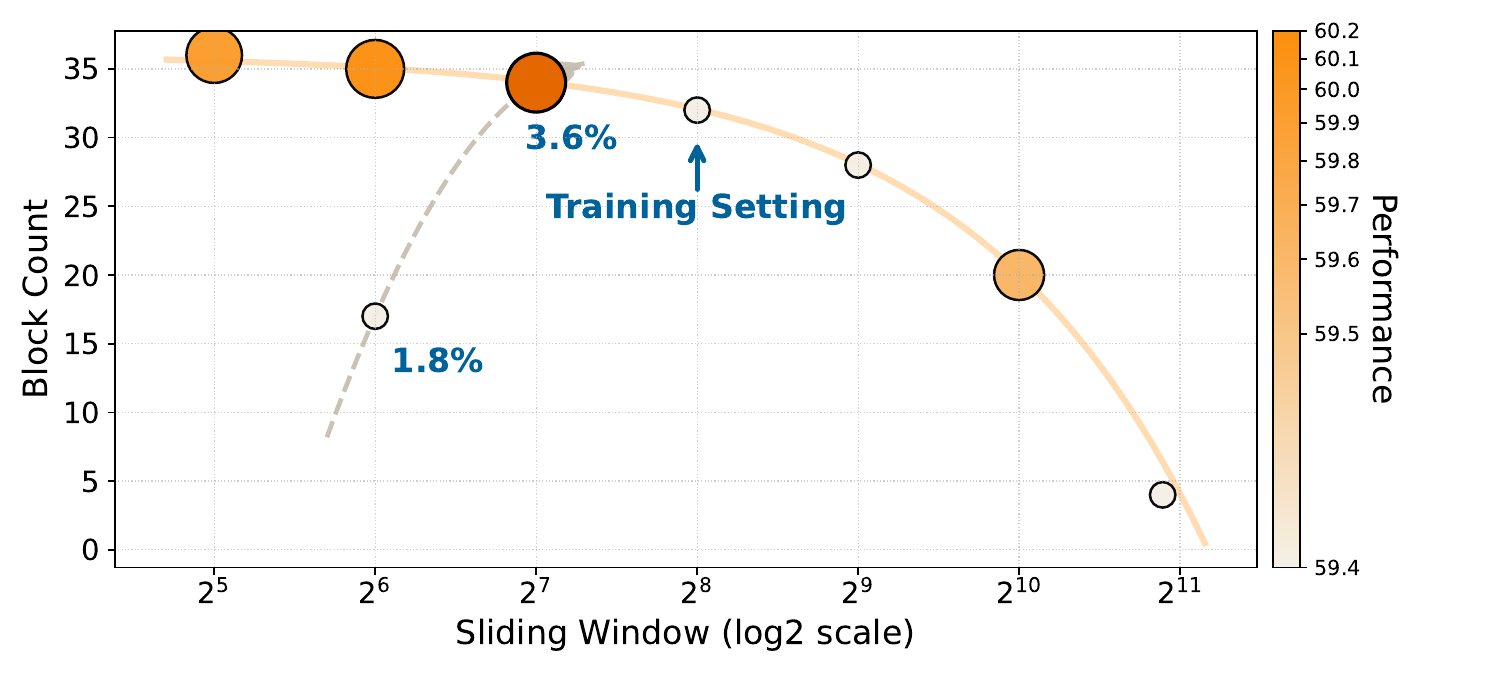}
         } &
         \subfloat[Attention Scaling of TimeScope]{\includegraphics[width=0.48\textwidth]{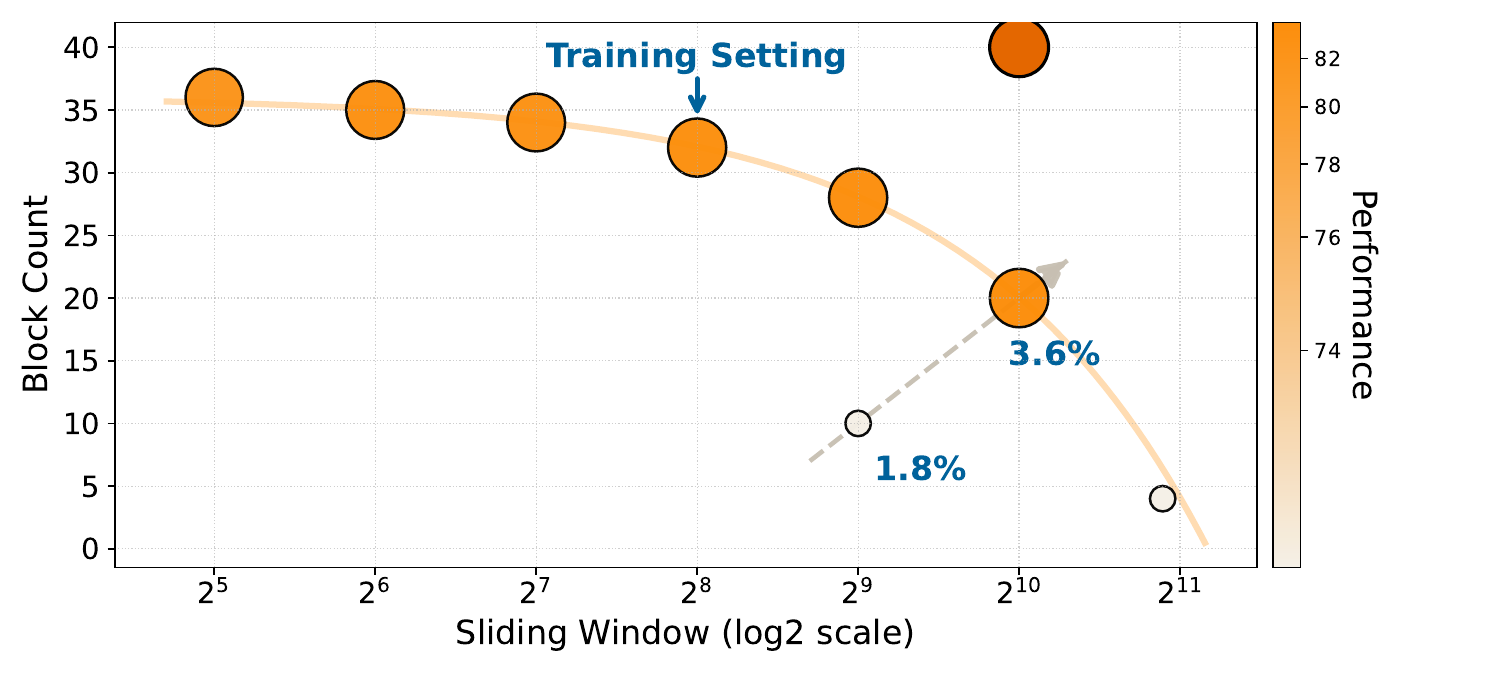}} 
         \\
    \end{tabular}
    \caption{Scaling Performance of VideoNSA under Different Attention Allocation Strategies. 
We highlight the attention budget constraint to indicate settings with equal attention budget, 
and annotate the best-performing configuration under each benchmark. }
\label{app:attention_budget}
\end{figure}

In addition to the overall scaling trends, we further report detailed subtask-level results under different allocation settings in Table~\ref{tab:frame_longtime}, Table~\ref{tab:frame_time}, Table~\ref{tab:frame_longvid}, Table~\ref{tab:frame_mlvu}, Table~\ref{tab:frame_tomato}, and Table~\ref{tab:frame_vsi}.

\setlength{\tabcolsep}{3pt}
\arrayrulecolor{black}
\color{black}
\begin{table*}[ht]
\centering
\caption{LongTimeScope~\cite{zohar2025apollo2} results across different attention budget strategy. Metrics include overall accuracy and task-specific scores across different steps.}
\resizebox{0.98\textwidth}{!}{
\begin{tabular}{lccccccccccc}
\toprule
\multirow{2}{*}{Block Count} & \multirow{2}{*}{Window Size} & \multirow{2}{*}{Overall} 
  & \multicolumn{3}{c}{18000} 
  & \multicolumn{3}{c}{28800} 
  & \multicolumn{3}{c}{36000} \\
\cmidrule(lr){4-6} \cmidrule(lr){7-9} \cmidrule(lr){10-12}
 & & & OCR & QA & Temporal & OCR & QA & Temporal & OCR & QA & Temporal \\
\midrule
36 & 32 & 44.0 & 56.0 & 50.0 & 28.0 & 46.0 & 66.0 & 16.0 & 46.0 & 72.0 & 16.0 \\
35 & 64   & 44.0 & 54.0 & 58.0 & 26.0 & 46.0 & 68.0 & 12.0 & 46.0 & 74.0 & 12.0 \\
34 & 128  & 41.8 & 50.0 & 56.0 & 28.0 & 44.0 & 64.0 & 6.0  & 46.0 & 74.0 & 8.0  \\
28 & 512  & 42.0 & 50.0 & 56.0 & 28.0 & 48.0 & 64.0 & 6.0  & 46.0 & 76.0 & 4.0  \\
20 & 1024 & 40.9 & 52.0 & 56.0 & 28.0 & 48.0 & 64.0 & 0.0  & 44.0 & 76.0 & 0.0  \\
4  & 1900 & 0.0  & 0.0  & 0.0  & 0.0  & 0.0  & 0.0  & 0.0  & 0.0  & 0.0  & 0.0  \\
16 & 128  & 41.6 & 52.0 & 56.0 & 26.0 & 46.0 & 62.0 & 10.0 & 42.0 & 76.0 & 4.0  \\
64 & 512  & 42.4 & 52.0 & 56.0 & 28.0 & 48.0 & 64.0 & 8.0  & 46.0 & 76.0 & 4.0  \\
\bottomrule
\end{tabular}
}
\label{tab:attention_longtime}
\end{table*}

\begin{table*}[ht]
\centering
\caption{TimeScope~\cite{zohar2025apollo2} results across different attention budget strategy. Metrics include overall accuracy and task-specific scores across different steps.}
\resizebox{0.98\textwidth}{!}{
\begin{tabular}{lcccccccccccc}
\toprule
Block Count & Window Size & Overall & 60 & 120 & 180 & 300 & 600 & 1200 & 1800 & 3600 & 7200 & 10800 \\ 
\midrule
36 & 32   & 81.6 & 97.4 & 97.4 & 96.7 & 85.4 & 82.7 & 86.3	& 85.3	& 71.3	& 60.6	& 52.7 \\
35 & 64   & 82.4 & 92.7 & 91.3 & 92.7 & 91.4 & 83.4 & 91.6	& 91.3	& 74.6	& 63.3	& 52.0 \\
34 & 128  & 82.2 & 96.7 & 96.7 & 94.7 & 89.4 & 82.0 & 88.3	& 85.3	& 76.6	& 60.0	& 52.7 \\
28 & 512  & 82.8 & 94.7 & 97.4 & 94.7 & 90.7 & 82.7 & 91.6	& 88.6	& 74.0	& 63.3	& 50.0 \\
20 & 1024 & 83.2 & 96.7 & 97.4 & 97.4 & 88.7 & 85.4 & 89.0	& 84.6	& 78.6	& 58.6	& 55.3 \\
4  & 1900 & 8.6  & 4.7  & 4.7  & 4.7  & 4.7  & 4.7  & 12.3	& 13.3	& 13.3	& 13.3	& 10.0 \\
10 & 512  & 59.9 & 4.7  & 4.7  & 94.7 & 88.7 & 79.4 & 85.6	& 80.0	& 68.0	& 58.0	& 35.3 \\
40 & 1024 & 83.7 & 96.7 & 96.0 & 97.4 & 92.0 & 85.4 & 91.6	& 89.3	& 73.3	& 63.3	& 52.0 \\
\bottomrule
\end{tabular}
}
\label{tab:attention_time}
\end{table*}

\begin{table*}[ht]
\centering
\caption{LongVideoBench~\cite{wu2024longvideobench} results across different attention budget strategy. Metrics include overall accuracy and task-specific scores across different steps.}
\resizebox{0.98\textwidth}{!}{
\begin{tabular}{cc|cccccccccccccccccccccc}
\toprule
Block Count & Window Size & Overall & 600.0 & TOS & S2E & E3E & S2A & SAA & O3O & T3O & T3E & O2E & T2O & S2O & TAA & T2E & E2O & SSS & T2A & 60.0 & SOS & 15.0 & 3600.0 \\
\midrule
36 & 32   & 59.9 & 57.7 & 48.1 & 64.0 & 66.6 & 72.9 & 55.6 & 52.5 & 58.4 & 57.7 & 59.9 & 54.4 & 70.9 & 52.1 & 56.2 & 70.1 & 32.8 & 66.5 & 67.5 & 72.4 & 65.2 & 56.1 \\
35 & 64   & 60.1 & 58.7 & 48.1 & 64.0 & 65.5 & 75.2 & 57.0 & 55.5 & 59.7 & 59.1 & 58.7 & 55.7 & 66.7 & 49.7 & 56.2 & 70.1 & 34.8 & 65.2 & 66.9 & 73.7 & 65.7 & 55.9 \\
34 & 128  & 60.2 & 59.9 & 48.1 & 65.1 & 67.6 & 74.1 & 55.6 & 55.5 & 58.4 & 56.3 & 62.2 & 57.0 & 63.9 & 53.3 & 56.2 & 71.6 & 35.9 & 62.7 & 67.5 & 72.4 & 66.3 & 55.1 \\
28 & 512  & 59.4 & 60.4 & 46.7 & 62.9 & 66.6 & 74.1 & 58.4 & 52.5 & 54.3 & 56.3 & 64.5 & 55.7 & 59.8 & 49.7 & 56.2 & 79.3 & 34.8 & 61.4 & 67.5 & 68.7 & 64.1 & 53.3 \\
20 & 1024 & 59.6 & 60.6 & 45.4 & 66.1 & 66.6 & 72.9 & 59.8 & 54.0 & 54.3 & 55.0 & 64.5 & 57.0 & 58.4 & 50.9 & 56.2 & 80.9 & 32.8 & 62.7 & 69.2 & 67.5 & 63.6 & 53.1 \\
4  & 1900 & 28.3 & 27.6 & 23.4 & 24.2 & 35.7 & 25.2 & 32.0 & 29.7 & 27.3 & 24.8 & 31.1 & 30.7 & 33.4 & 35.1 & 27.0 & 28.5 & 26.6 & 19.7 & 27.4 & 28.0 & 27.1 & 29.7 \\
17 & 64   & 59.4 & 58.0 & 46.7 & 62.9 & 65.5 & 74.1 & 55.6 & 52.5 & 58.4 & 57.7 & 58.7 & 57.0 & 66.7 & 50.9 & 56.2 & 71.6 & 30.7 & 64.0 & 68.1 & 72.4 & 65.7 & 54.2 \\
\bottomrule
\end{tabular}
}
\label{tab:attention_longvid}
\end{table*}
\begin{table*}[ht]
\centering
\caption{MLVU~\cite{zhou2024mlvu} results across different attention budget strategy. Metrics include overall accuracy and task-specific scores across different steps.}
\resizebox{0.98\textwidth}{!}{
\begin{tabular}{cc|cccccccccc}
\toprule
Block Count & Window Size & Overall & Direction & Count & Rotation & Shape\&Trend & Velocity\&Freq. & Visual Cues & Human & Simulated & Object \\
\midrule
32 & 256  & 26.5 & 21.6 & 31.5 & 22.0 & 25.6 & 23.3 & 40.0 & 21.7 & 23.6 & 29.3 \\
36 & 32   & 25.9 & 21.6 & 32.5 & 19.2 & 25.1 & 25.5 & 37.1 & 21.4 & 21.5 & 29.2 \\
35 & 64   & 27.1 & 23.8 & 33.9 & 20.6 & 25.6 & 25.5 & 37.1 & 22.1 & 24.2 & 30.8 \\
34 & 128  & 27.2 & 23.8 & 34.2 & 20.3 & 25.1 & 25.5 & 38.6 & 21.9 & 24.2 & 30.8 \\
28 & 512  & 26.1 & 21.8 & 32.2 & 19.2 & 24.7 & 27.5 & 37.1 & 22.1 & 22.4 & 28.2 \\
20 & 1024 & 25.1 & 20.6 & 30.8 & 17.5 & 23.3 & 29.4 & 34.3 & 21.4 & 23.3 & 25.6 \\
64 & 512  & 25.3 & 21.3 & 30.5 & 19.6 & 24.2 & 27.5 & 32.9 & 21.4 & 22.9 & 27.4 \\
4  & 2048 & 26.4 & 21.8 & 33.6 & 20.3 & 25.6 & 27.5 & 32.9 & 21.8 & 24.7 & 29.4 \\
16 & 128  & 21.4 & 19.5 & 17.5 & 20.2 & 21.0 & 30.0 & 28.8 & 17.8 & 17.6 & 20.6 \\
\bottomrule
\end{tabular}
}
\label{tab:attention_mlvu}
\end{table*}
\begin{table*}[ht]
\centering
\caption{Tomato~\cite{shangguan2024tomato} results across different attention budget strategy. Metrics include overall accuracy and task-specific scores across different steps.}
\resizebox{0.98\textwidth}{!}{
\begin{tabular}{cc|ccccccccccc}
\toprule
Block Count & Window Size & Overall & Direction & Count & Rotation & Shape\&Trend & Velocity\&Freq. & Visual Cues & Human & Simulated & Object \\
\midrule
36 & 32   & 25.9 & 21.6 & 32.5 & 19.2 & 25.1 & 25.5 & 37.1 & 21.4 & 21.5 & 29.2 \\
35 & 64   & 27.1 & 23.8 & 33.9 & 20.6 & 25.6 & 25.5 & 37.1 & 22.1 & 24.2 & 30.8 \\
34 & 128  & 27.2 & 23.8 & 34.2 & 20.3 & 25.1 & 25.5 & 38.6 & 21.9 & 24.2 & 30.8 \\
28 & 512  & 26.1 & 21.8 & 32.2 & 19.2 & 24.7 & 27.5 & 37.1 & 22.1 & 22.4 & 28.2 \\
20 & 1024 & 25.1 & 20.6 & 30.8 & 17.5 & 23.3 & 29.4 & 34.3 & 21.4 & 23.3 & 25.6 \\
64 & 512  & 25.3 & 21.3 & 30.5 & 19.6 & 24.2 & 27.5 & 32.9 & 21.4 & 22.9 & 27.4 \\
4  & 2048 & 26.4 & 21.8 & 33.6 & 20.3 & 25.6 & 27.5 & 32.9 & 21.8 & 24.7 & 29.4 \\
16 & 128  & 21.4 & 19.5 & 17.5 & 20.2 & 21.0 & 30.0 & 28.8 & 17.8 & 17.6 & 20.6 \\
\bottomrule
\end{tabular}
}
\label{tab:attention_tomato}
\end{table*}
\begin{table*}[ht]
\centering
\caption{VSIBench~\cite{yang2025thinkingspacemultimodallarge} results across different attention budget strategy. Metrics include overall accuracy and task-specific scores across different steps.}
\resizebox{0.98\textwidth}{!}{
\begin{tabular}{cc|ccccccccc}
\toprule
Block Count & Window Size & Overall & Appearance & Abs. Dist. & Counting & Rel. Dist. & Size Est. & Room Est. & Route Plan. & Rel. Dir. \\
\midrule
28 & 512  & 36.0 & 23.9 & 18.2 & 45.5 & 36.9 & 54.1 & 29.8 & 36.0 & 43.4 \\
20 & 1024 & 35.9 & 24.0 & 18.5 & 46.8 & 36.7 & 53.6 & 29.0 & 36.5 & 42.0 \\
4  & 1900 & 0.0  & 0.0  & 0.0  & 0.0  & 0.0  & 0.0  & 0.0  & 0.0  & 0.0  \\
34 & 128  & 35.5 & 24.7 & 18.7 & 37.8 & 36.2 & 53.9 & 31.7 & 36.0 & 45.2 \\
35 & 64   & 35.4 & 25.5 & 20.4 & 33.2 & 35.4 & 53.6 & 31.5 & 38.1 & 46.0 \\
36 & 32   & 34.9 & 25.3 & 20.6 & 28.4 & 36.0 & 54.4 & 30.9 & 38.1 & 45.9 \\
36 & 62   & 35.3 & 25.2 & 20.5 & 28.3 & 35.8 & 54.7 & 31.0 & 41.1 & 46.1 \\
16 & 128  & 0.0  & 0.0  & 0.0  & 0.0  & 0.0  & 0.0  & 0.0  & 0.0  & 0.0  \\
64 & 512  & 35.8 & 23.6 & 18.3 & 45.9 & 36.8 & 54.2 & 29.3 & 36.5 & 42.3 \\
\bottomrule
\end{tabular}
}
\label{tab:attention_vsi}
\end{table*}
\section{Theoretical Foundations of Scaling Behavior}
\label{theory}
\label{sec:theory}

In Section~\ref{sec:findings}, we perform two scaling experiments along context length and attention budget. We observe that~\model~exhibits strong extrapolation ability on context length: although trained with only 36K tokens, it can generalize to 128K at test time, achieving the best performance at 64K. In contrast, when scaling the attention budget, even a small reduction to \~3.6\% of attention computation already delivers outstanding performance, and further increasing the visible-token count does not yield additional gains. To clarify these phenomena, we provide theoretical interpretations from routing-path stability and the geometric structure of RoPE~\cite{su2024roformer}.

\paragraph{Routing-path Stability.}Recent work~\cite{huang2025transformers} indicates that a model’s ability to maintain performance on long sequences depends critically on the stability of its attention routing structure across positions. In the standard attention mechanism, the attention weight from the query vector \(Q_n\) at position \(n\) to the key vector \(Z_j\) at position \(j\) is defined as
\[
\text{Attn}_{n\to j}
=\frac{\exp(Z_j^\top Q_n)}{\sum_k \exp(Z_k^\top Q_n)}.
\]
Here, \(Q_n\) and \(Z_j\) denote the query and key representations at positions \(n\) and \(j\), respectively. If the model can consistently focus its attention on the task-relevant target token set \(\mathcal{T}\) during inference, then (i) \(\sum_{j\in\mathcal{T}}\text{Attn}_{n\to j}\) should dominate across different positions; and (ii) the attention assigned to the same key token \(j\) should remain nearly unchanged under positional shifts, i.e.,
\[
\Delta_i=\bigl|\text{Attn}_{n\to j}-\text{Attn}_{n+i\to j}\bigr|\approx 0,
\]
where \(\Delta_i\) measures the deviation of routing paths across positions in long sequences.

When we scale the context length using dense temporal and spatial sampling, the sparse-attention pattern and mask structure \(M\) remain unchanged, which means the model continues to use the routing structure learned during training while simply facing a larger pool of candidate evidence. Since denser sampling mainly introduces redundant or finer-grained details, the model treats these tokens as auxiliary evidence, leaving the core target tokens and their relative attention weights essentially unchanged. Consequently, the overall routing-path structure is preserved, \(\Delta_i\) remains small, and the model can maintain or even improve its performance at longer context lengths.

In contrast, attention-budget scaling explicitly modifies the set of visible tokens in the sparse-attention mechanism by replacing the original mask \(M\) with a new mask \(M'\). The effective query becomes
\[
Q_{\text{eff}} = Q \odot M',
\]
where \(\odot\) denotes elementwise multiplication, and the corresponding new attention weight is
\[
\text{Attn}'_{n\to j}\propto \exp\bigl(Z_j^\top(Q\odot M')\bigr).
\]
Even if the modification from \(M\) to \(M'\) appears small in proportion, it substantially changes the set of candidate evidence accessible to each query and alters the relative logits. First, the newly visible tokens reduce the relative weight allocated to the original key tokens, producing a dilution effect. Second, in video tasks, the added visible tokens often lie in similar visual-semantic clusters as the original key tokens and thus have non-negligible similarity scores \(Z_{j'}^\top Q\). These tokens directly compete with and divert attention away from the originally dominant key tokens. Under the combined influence of these effects, the stable routing-path structure learned during training is overwritten: formerly high-weight key tokens may be diluted or overshadowed by the new candidates, causing \(\Delta_i\) to increase significantly. As a result, the model is more likely to follow incorrect reasoning paths, leading to degraded performance.

\paragraph{Geometric Rotational of RoPE.} RoPE~\cite{su2024roformer} maps the representation at position \(i\) into a rotation in a two-dimensional subspace:
\[
q'(i)=R(i\omega)q,\qquad k'(j)=R(j\omega)k,
\]
where \(R(i\omega)\) is a rotation matrix and \(\omega\) denotes the frequency parameters. This yields an inner product that depends only on the relative distance between the two positions:
\[
\langle q'(i),k'(j)\rangle=\langle R((i-j)\omega)q,k\rangle.
\]
RoPE~\cite{su2024roformer} therefore establishes a structured geometric correspondence between relative distance and rotation phase. Under this geometry, when the context length is moderately increased (e.g., from 36K to 64K), the model only needs to resolve a larger phase difference \(d\omega\); within this range, the growth of the phase still lies in the extrapolation regime covered by the empirical distribution seen during training. As a result, the model can naturally generalize.

LM-Infinite~\cite{han2023lm} further proves that, in order to distinguish the growing clusters of relative distances \(\alpha(n)\), the attention logit must increase monotonically with sequence length:
\[
\sup_{q,k,d\le n} |w(q,k,d)|\ge
\left(\frac{\alpha(n)}{2}\right)^{1/(2r)}\frac{\varepsilon}{4e},
\]
where \(w(q,k,d)\) denotes the logit at relative distance \(d\), and \(\alpha(n)\) grows with \(n\). This ``logit growth'' is controlled and beneficial at moderate lengths, expanding the dynamic range of attention and enabling the model to maintain token separability over larger distances and consistenting with the strong performance we observe around 64K.

However, when the effective phase difference \(d\omega\) becomes excessively large, the rotation angle may approach or exceed the periodic range of multiple frequency dimensions, giving rise to \emph{phase aliasing}: tokens that should correspond to distinct relative distances collapse into similar or even indistinguishable phase regions. In such cases, although attention logits continue to grow with length, the high-frequency components of RoPE lose their discriminative resolution, reducing geometric separability among tokens, which aligns with existing analyses~\cite{press2021train, chen2023extending} showing the degradation of relative positional encoding at extreme distances.

\section{Full Gate Values Distribution}
\label{fullgate_dis}

\begin{figure}[t]
    \centering
    \includegraphics[width=\linewidth]{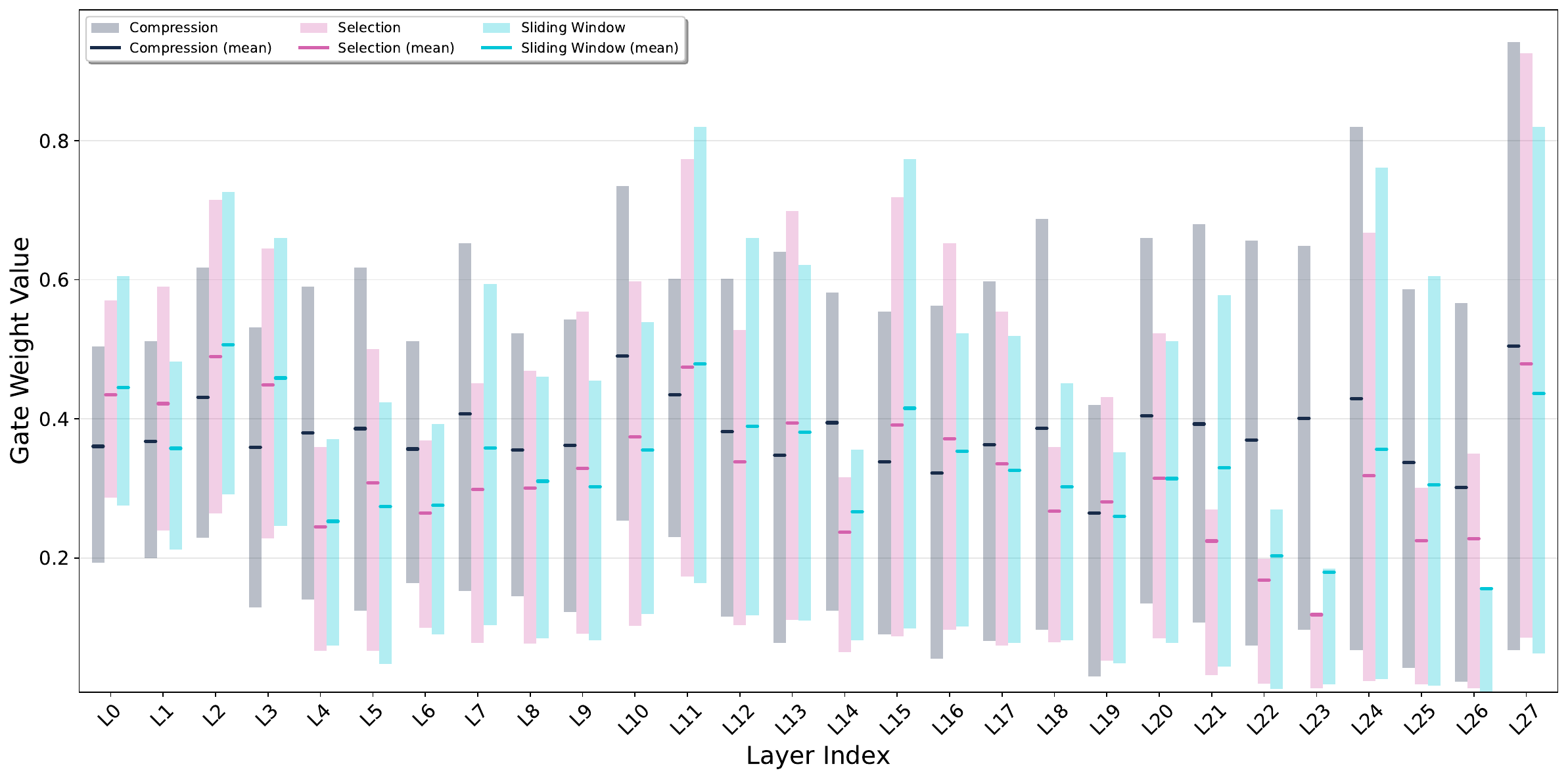}
    \caption{Gate weight distribution of each layer.}
    \label{fig:gate_all}
\end{figure}
\section{More Inter-head Gate Similarites Visualization}
\label{all_gate_corr}

\foreach \i in {0,4,...,24}{%
  \begin{figure}[H]
  \captionsetup{font=tiny}
    \centering
    \begin{minipage}{0.48\textwidth}
      \centering
      \caption*{L\i}
      \vspace{-1em}
      \includegraphics[width=\linewidth]{fig/gate_corr/gate_corr_L\i.pdf}
    \end{minipage}\hfill
    \begin{minipage}{0.48\textwidth}
      \centering
      \caption*{L\the\numexpr\i+1\relax}
      \vspace{-1em}
      \includegraphics[width=\linewidth]{fig/gate_corr/gate_corr_L\the\numexpr\i+1\relax.pdf}
    \end{minipage}
    \caption{More Inter-head Gate Similarites Visualization}
  \end{figure}
  \vspace{-2em}
}

\section{Benchmark-level Gating Analysis and PCA Visualization}
\label{pca}
\label{sec:pca}

\begin{figure}[!ht]
    \captionsetup{skip=2pt}
    \centering
    \begin{tabular}{cccc}
         \subfloat[Compression Branch]{
             \includegraphics[width=0.32\textwidth]{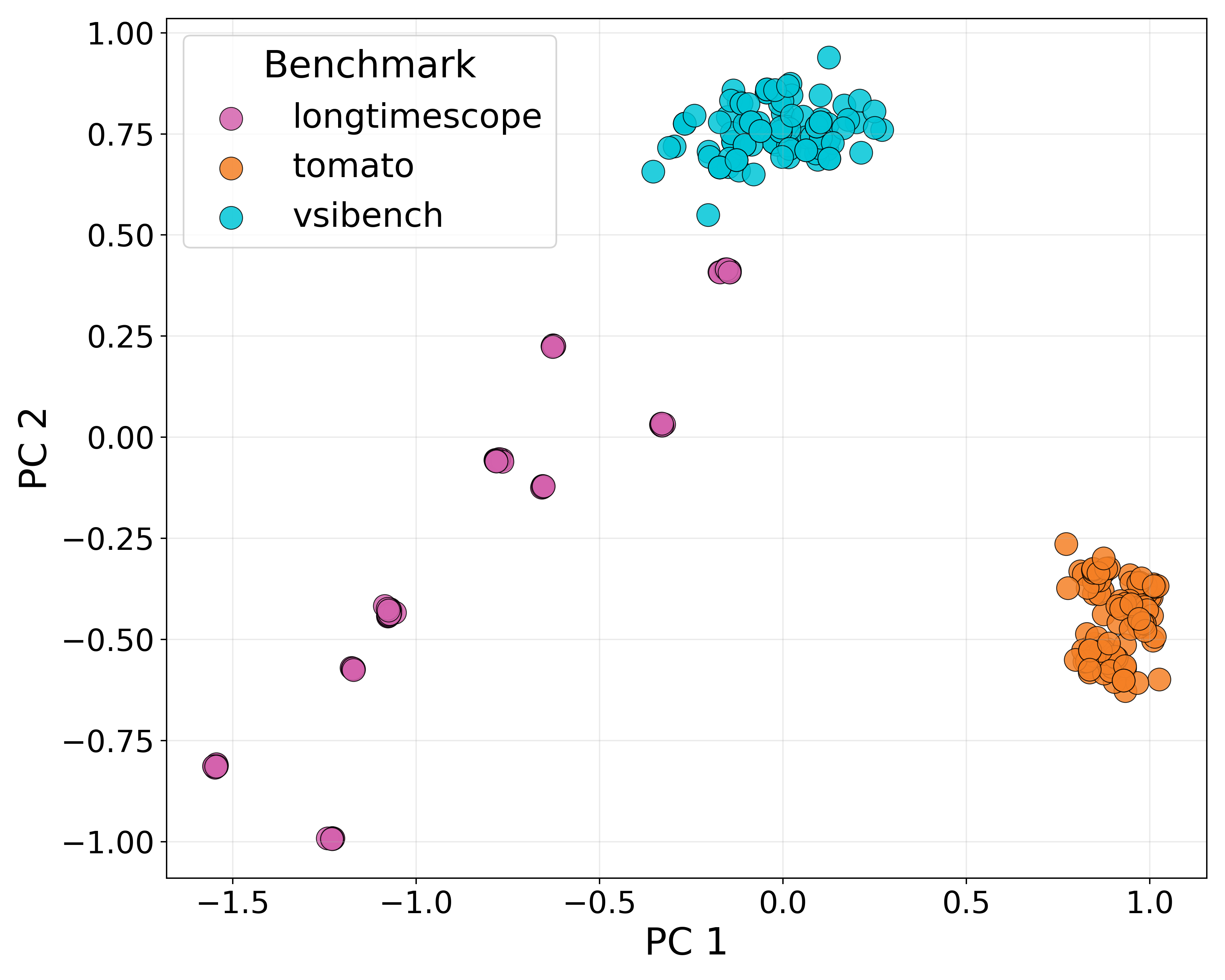}
         } &
         \subfloat[Selection Branch]{
             \includegraphics[width=0.32\textwidth]{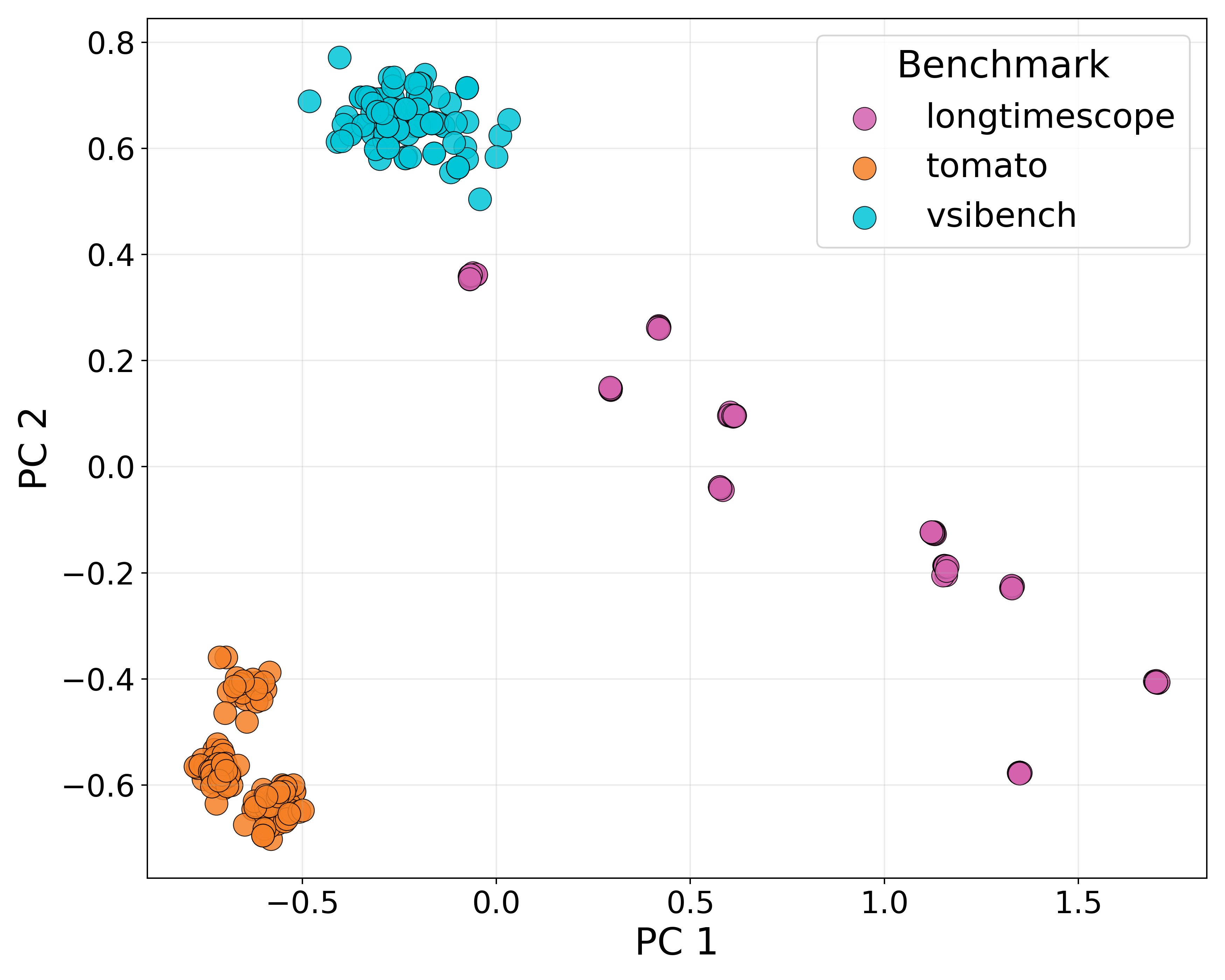}
         } &
         \subfloat[Sliding Window]{\includegraphics[width=0.32\textwidth]{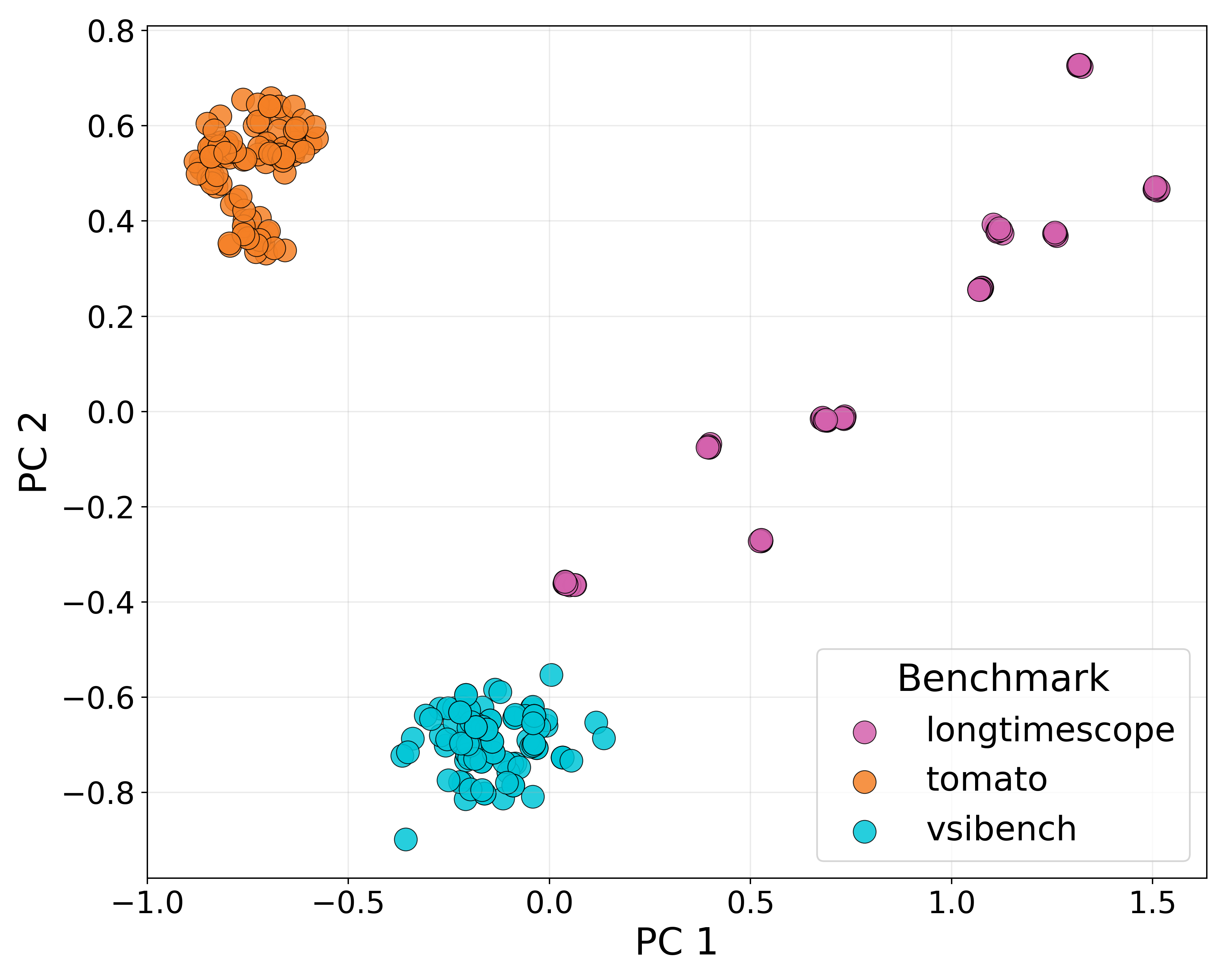}} 
         \\
    \end{tabular}
    \caption{PCA visualization of benchmark-level gate patterns.}
    \label{fig:pca}
\end{figure}

In this section, we provide additional evidence that~\model’s routing strategy depends on input video content rather than layer depth alone. We collect the layer–head gate vectors for representative videos from three benchmarks with distinct visual properties (LongTimeScope~\cite{zohar2025apollo2} for multi-shot transitions, Tomato~\cite{shangguan2024tomato} for high-frequency motion, and VSIBench~\cite{yang2025thinkingspacemultimodallarge} for complex spatial layouts) and project the gate vectors into a 2D space using PCA.

As shown in Figure~\ref{fig:pca}, the gate patterns form three clearly separated clusters, regardless of whether we use the compression branch, the selection branch, or the sliding-window branch, which indicates that~\model~learns benchmark-specific routing strategies conditioned on visual content, rather than following a fixed depth pattern.

\definecolor{sand}{HTML}{f5f0E6}   
\definecolor{blue}{HTML}{00629B}   

\begin{table*}[h]
\centering
\vspace{-10pt}
\caption{Ablation on static gates averaged over a 1K training subset.}
\addtolength\tabcolsep{-2.4pt} 
\resizebox{1\linewidth}{!}{
\begin{tabular}{ccccccc}
\toprule
\multicolumn{1}{c}{\multirow{2}{*}{Model}} & \multicolumn{4}{c}{Long Video Understanding}  & \multicolumn{1}{c}{Temporal Reasoning} & \multicolumn{1}{c}{Spatial Understanding} \\
\cmidrule(lr){2-5} \cmidrule(lr){6-6} \cmidrule(lr){7-7}
\multicolumn{1}{c}{}& LongVideoBench & MLVU$_{Test}$ & TimeScope & LongTimeScope & Tomato & VSIBench \\ 

\midrule
\textcolor{gray}{Qwen2.5-VL-7B} & \textcolor{gray}{58.7} & \textcolor{gray}{51.2} & \textcolor{gray}{81.0} & \textcolor{gray}{40.7} & \textcolor{gray}{22.6} & \textcolor{gray}{29.7}\\
\model & \textbf{59.4 (+1.1\%)} & \textbf{51.8 (+1.2\%)} & 82.7 (+2.1\%) & \textbf{44.4 (+9.1\%)} & \textbf{26.2 (+15.9\%)} & \textbf{36.1 (+20.3\%)} \\
\model+ Static Gate & \textbf{58.4 (-0.5\%)} & \textbf{51.2 (0.0\%)} & 81.2 (+0.2\%) & \textbf{41.5 (+2.0\%)} & \textbf{23.7 (+4.8\%)} & \textbf{31.8 (+7.1\%)} \\
					
\bottomrule
\end{tabular}}

\label{tab:static_gate} 
\end{table*}

\arrayrulecolor{black}
\color{black}

To further isolate the role of input-driven routing, we replace each layer’s gate with a static value averaged from a 1K training subset, forcing the model to depend only on layer depth.
As shown in Table~\ref{tab:static_gate}, the performance drops across all six benchmarks, especially on tasks requiring long-range temporal integration, confirming that dynamic gating is essential.

\section{Additional Analysis of Training and Inference Efficiency}
\label{flops}
\label{sec:flops}

\begin{table}[t]
\centering
\caption{Theoretical FLOPs comparison among different attention mechanisms.
``VideoNSA (ideal)'' denotes the theoretical FLOPs of NSA without query-head padding.}
\small
\begin{tabular}{lcc}
\toprule
\textbf{Method} & \textbf{FLOPs} & \textbf{Relative} \\
\midrule
Flash Attention     & 8.40 PF  & 1.00$\times$ \\
Tri-shape           & 7.07 PF  & 0.84$\times$ \\
MInference          & 4.13 PF  & 0.49$\times$ \\
Flexprefill         & 7.75 PF  & 0.92$\times$ \\
XAttention          & 1.94 PF  & 0.23$\times$ \\
VideoNSA (ideal)    & 2.05 PF  & 0.24$\times$ \\
VideoNSA            & 4.68 PF  & 0.56$\times$ \\
\bottomrule
\end{tabular}
\label{tab:flops_comparison}
\end{table}

\begin{figure}[!ht]
    \captionsetup{skip=2pt}
    \centering
    \begin{tabular}{ccc}
         \subfloat[Absolute prefill latency across attention mechanisms.]{
             \includegraphics[width=0.98\textwidth]{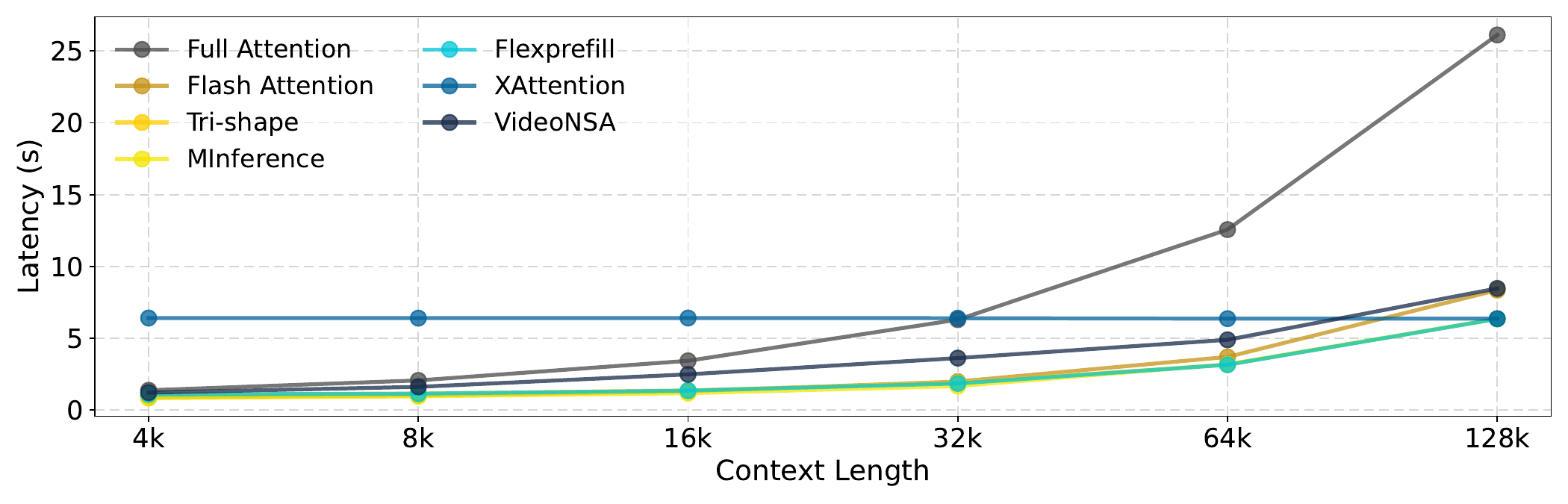}
         } \\
         \subfloat[Prefill-time speedup over full attention across context lengths.]{\includegraphics[width=0.98\textwidth]{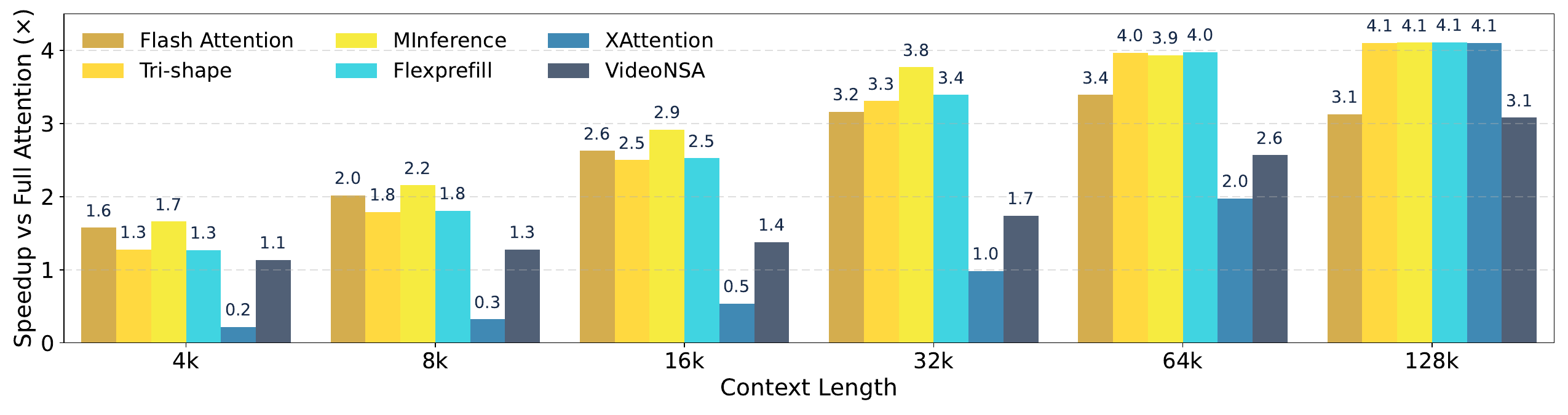}} 
         \\
    \end{tabular}
    \caption{Inference efficiency comparison across attention mechanisms.}
\label{app:efficiency_overall}
\end{figure}

To complement the efficiency discussion in the main paper, we provide additional analysis of both FLOPs and wall-clock latency across different attention mechanisms and context lengths.

\paragraph{Training Efficiency.}
Under identical optimization settings, training~\model~requires approximately $4600$ H100 GPU hours, while the dense baseline requires $5280$ H100 GPU hours. This corresponds to $0.87\times$ of the dense baseline, indicating that~\model~achieves slightly improved training efficiency despite using a more complex attention mechanism.

\paragraph{Inference Efficiency.}
Table~\ref{tab:flops_comparison} presents the theoretical FLOPs of different attention mechanisms. In the ideal case, NSA requires only $2.05$ PFLOPs, which is $0.24\times$ that of Flash Attention, demonstrating the theoretical computational efficiency of the sparse routing structure.
However, the actual FLOPs and wall-clock latency of~\model~are higher than this ideal value due to implementation constraints in the current NSA kernel. The Qwen2.5-VL 7B~\cite{qwen2025qwen25technicalreport} adopts an unusual head configuration of $4$ KV heads and $28$ query heads. To satisfy Triton kernel requirements, the query heads must be padded to $64$, which introduces additional computation and memory access overhead. As a result, the practical efficiency of~\model~deviates from its theoretical FLOPs advantage. As shown in Figure~\ref{app:efficiency_overall},~\model's latency grows much more slowly than dense attention, and compared with other sparse baselines, it delivers competitive inference speed while achieving stronger model performance.

\section{Additional Analysis on CMP Latency Bottleneck}
\label{cmp_latency}
\label{sec:cmp_latency}

\begin{table}[t]
\centering
\caption{CMP latency under different block sizes and context lengths.}
\label{tab:cmp_blocksize}
\vspace{4pt}

\begin{tabular}{c|cccccc}
\toprule
{\textbf{Block Size}} & \textbf{4k} & \textbf{8k} & \textbf{16k} & \textbf{32k} & \textbf{64k} & \textbf{128k} \\
\midrule
32  & 0.868 & 1.022 & 1.311 & 1.982 & 3.698 & 8.343 \\
64  & 0.882 & 1.036 & 1.322 & 1.997 & 3.687 & 8.323 \\
128 & 0.880 & 1.027 & 1.308 & 1.993 & 3.705 & 8.353 \\
\bottomrule
\end{tabular}

\end{table}

\begin{table}[t]
    \centering
    \caption{Latency comparison between different NSA implementations at 8k context length.}
    \label{tab:nsa_kernels}
    \vspace{4pt}
    \begin{tabular}{c|cc}
        \toprule
        \textbf{Implementation} & \textbf{forward} & \textbf{backward} \\
        \midrule
        nsa-impl~\cite{Pai2025SparsityIsCool} & 5.402 & 32.826 \\
        flash-nsa~\cite{mdy6662025ScalableFlashNativeSparseAttention} & 2.429 & 5.537 \\
        \bottomrule
    \end{tabular}
\end{table}

In this section, we provide additional analysis supporting the observation in findings that the CMP branch becomes the dominant source of latency as the context length increases.

Since the block size determines how many CMP operations are executed, we vary the block size and measure the resulting latency across multiple context lengths. 
As summarized in Table~\ref{tab:cmp_blocksize}, although increasing the block size reduces the number of CMP executions, the overall latency improvement remains small. Smaller blocks are dominated by memory-access overhead, while larger blocks incur higher computation per block. As a result, block-size scaling affects latency only moderately within a narrow range and does not change the overall scaling trend.

We also observe that the most significant acceleration comes from more efficient NSA implementations instead of architectural hyperparameters. As shown in Table~\ref{tab:nsa_kernels}, the flash-nsa~\cite{mdy6662025ScalableFlashNativeSparseAttention} implementation runs about twice as fast as our current nsa-impl~\cite{Pai2025SparsityIsCool} in the forward pass and up to six times faster in the backward pass. Other teams are also developing improved kernels such as optimizing NSA for TPUs~\cite{ko2025OptimizingNSAforTPUs}. These findings show that the dominant factor affecting CMP and overall NSA latency comes from kernel efficiency, including memory access patterns and kernel design.

\section{More Analysis about Attention Sinks on Various Sparse Attention Settings}
\label{block_sink}

\begin{figure}[!ht]
    \centering
    \begin{minipage}[b]{0.7\textwidth} 
      \centering
      \subfloat[Compression sinks across block size and counts.]{\includegraphics[width=\linewidth]{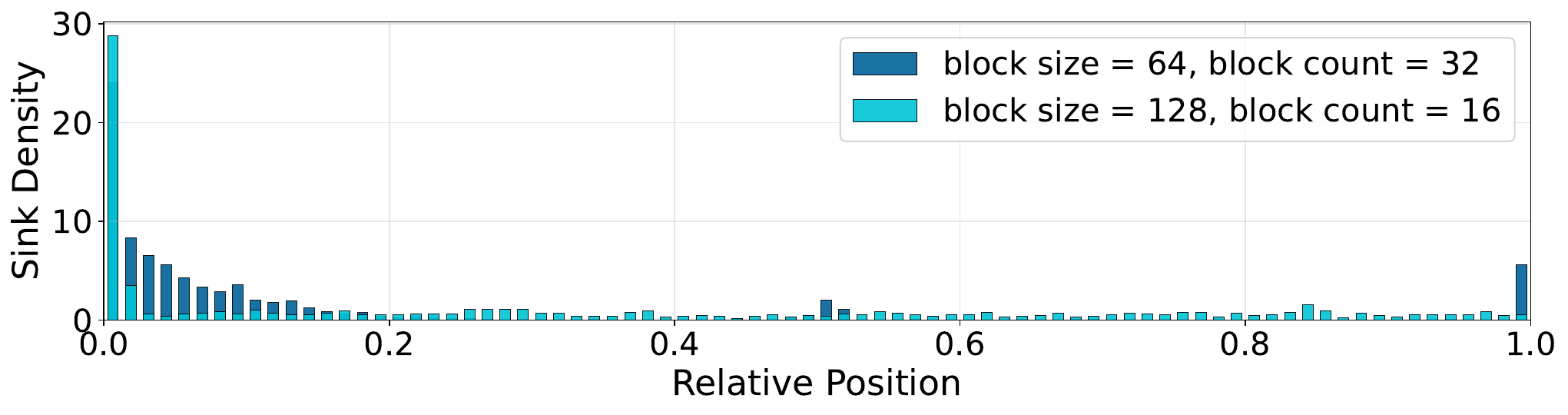}\label{fig:cmp_sink}}\\[0.3cm]
      \subfloat[Selection sinks across block size and counts.]{\includegraphics[width=\linewidth]{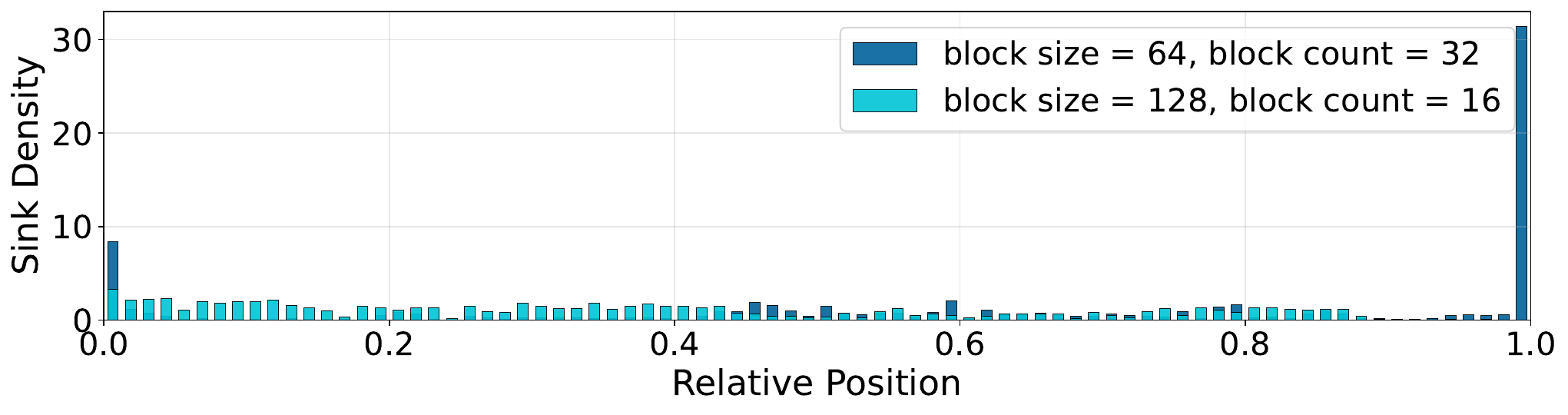}\label{fig:slc_sink}}\\[0.3cm]
      \subfloat[Sliding window sinks across window sizes.]{\includegraphics[width=\linewidth]{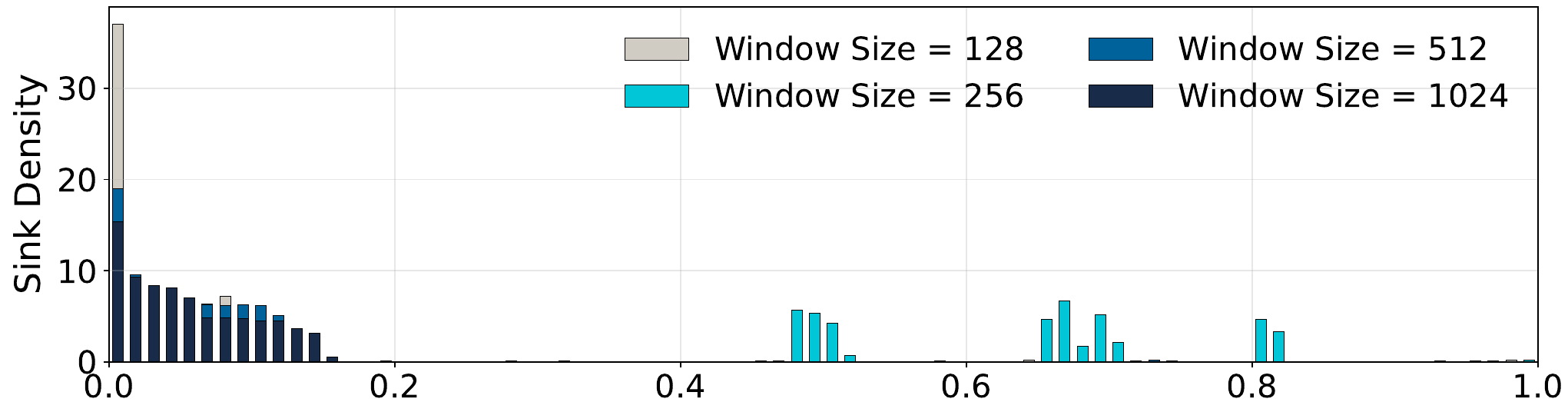}\label{fig:swa_sink}}
      \caption{Attention sink distributions across the three branches under different sparse settings.}
      \label{fig:branch_sink}
    \end{minipage}
\end{figure}

Figure~\ref{fig:cmp_sink} indicates that in the compression branch, smaller blocks produce sharper and higher sink peaks at the sequence start, while larger blocks used in training reduce the initial peak but introduce broader low-density diffusion with periodic boundary spikes.
The selection sinks in~\ref{fig:slc_sink} remain at consistently low densities under different configurations, suggesting that the top-k filtering mechanism robustly suppresses sink formation across different settings.
Figure~\ref{fig:branch_sink} shows the distribution of attention sinks under different sparse attention settings. When varying the window size, sinks are concentrated near the beginning and decay rapidly with position. Overall, larger windows yield lower sink density but broader coverage, while the training configuration ($w=256$) strikes a middle ground and exhibits sparse periodic clusters in the mid-to-late sequence, reflecting sensitivity to local boundaries learned during training.
\section{Discussion on Modality-Specific Sparsity}
\label{text_only}
\label{sec:text_only}

\begin{figure}[t]
    \centering
    \includegraphics[width=\linewidth]{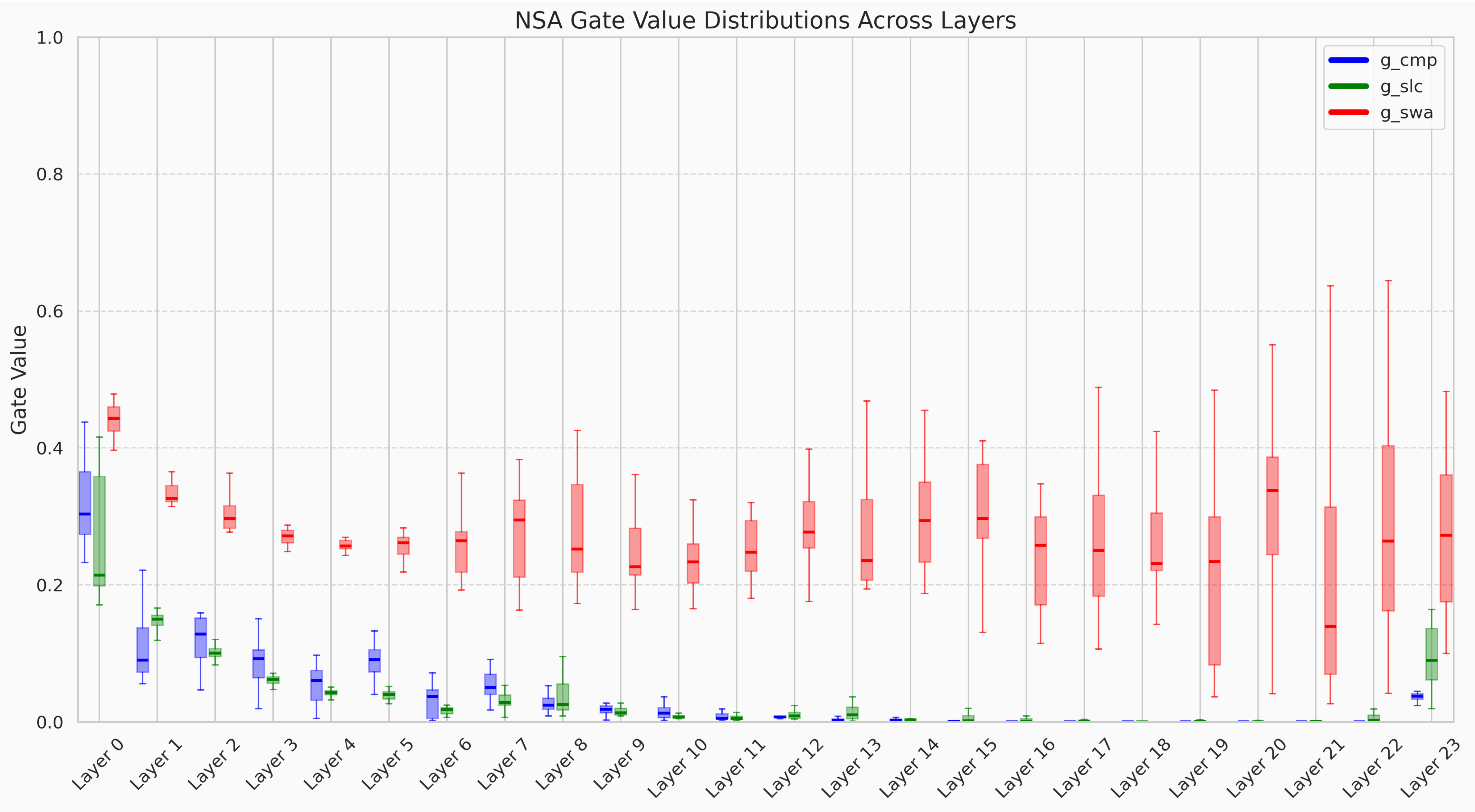}
    \caption{Layer-wise gate distributions of text-only NSA~\cite{Pai2025SparsityIsCool}.}
    \label{fig:text_only_layer}
\end{figure}

To further contextualize the modality-specific sparsity patterns exhibited by~\model, we compare its behavior with the text-only NSA~\cite{Pai2025SparsityIsCool} used in language models. As shown in Figure~\ref{fig:text_only_layer}, the text-only NSA~\cite{Pai2025SparsityIsCool} displays a distinct gating dynamic. The sliding-window branch gradually becomes dominant in deeper layers, while the compression and selection branches diminish rapidly and remain almost inactive throughout most of the network. This pattern reflects the one-dimensional and relatively uniform nature of textual sequences, where long-range interactions are sparse and stable, and the model tends to converge toward a single prevailing routing path. The text-only NSA~\cite{Pai2025SparsityIsCool} also presents a noticeable anomaly in the final layer, where all three branches suddenly become active again despite having remained largely inactive in previous layers. This behavior suggests a late-stage shift in inductive patterns that is characteristic of language modeling.
In contrast,~\model, as shown in Figure~\ref{fig:gate_all}, maintains active and balanced usage of all three branches across nearly the entire depth of the network, with the compression branch playing a consistently prominent role. 

\begin{figure}[t]
    \centering
    \includegraphics[width=\linewidth]{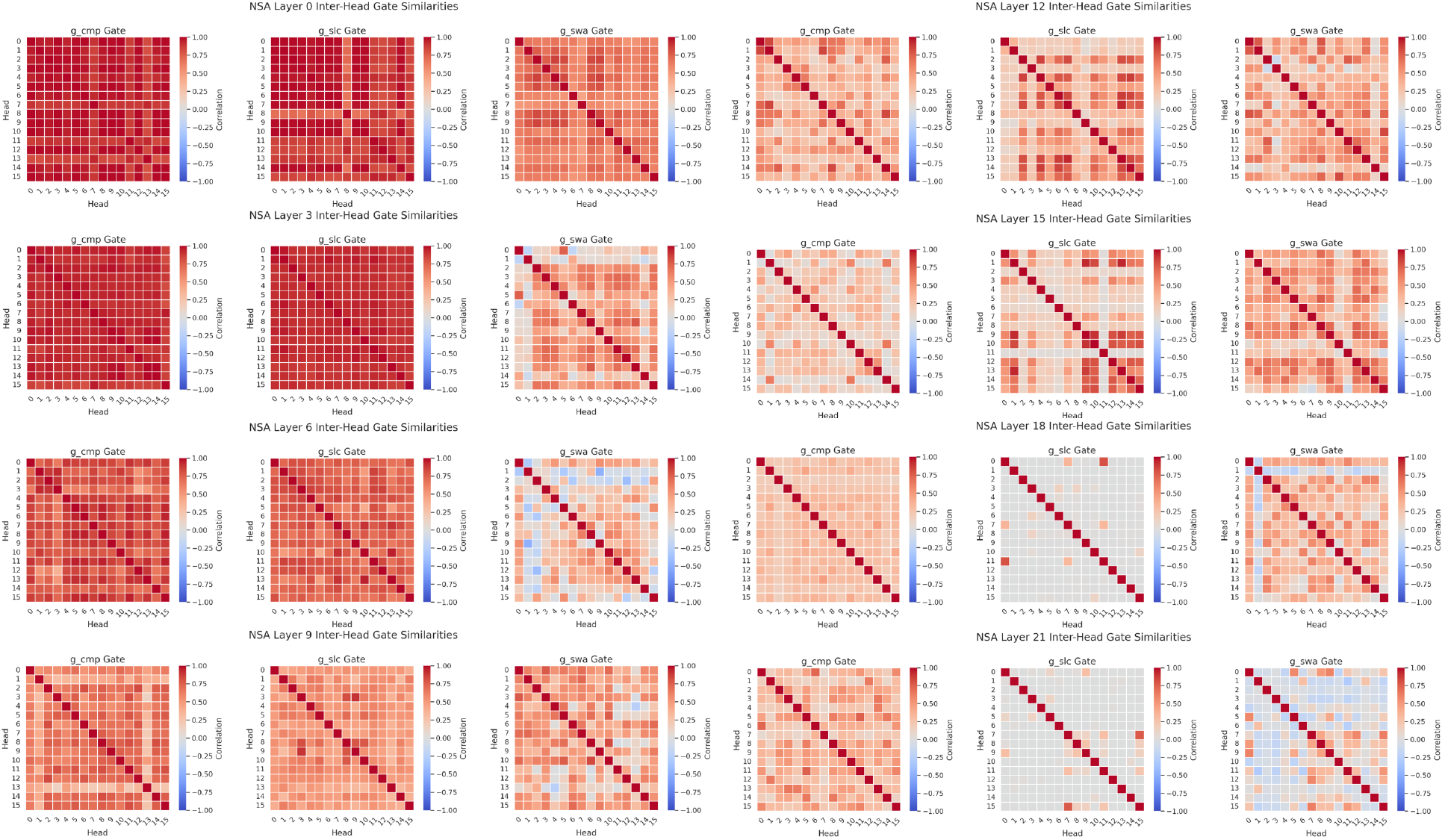}
    \caption{Inter-head gate similarities of text-only NSA~\cite{Pai2025SparsityIsCool}.}
    \label{fig:text_only_inter}
\end{figure}

The inter-head similarities further highlight the divergence between the two modalities. The text-only NSA~\cite{Pai2025SparsityIsCool} in Figure~\ref{fig:text_only_inter} exhibits strong correlations across heads in the early layers, which indicates a set of conserved induction-like operations. Later in the model, selection and sliding window gate values become decorrelated across heads.~\model, as shown in Section~\ref{all_gate_corr}, however, displays substantially weaker cross-head correlations overall, and only a few mid-layer clusters emerge in the selection and sliding-window branches. These findings imply that~\model~adjusts its sparse routing behavior to accommodate the rich spatiotemporal redundancy and multi-scale structure of video inputs, rather than collapsing into a single dominant pathway as observed in the text-only model.

\section{Dense Attention Sink Visualization}
\label{dense_attn_sink}

\foreach \i in {0,14,27}{
  \begin{center}
    \includegraphics[width=0.9\textwidth]{fig/dense_attn_sink/layer_\i\string_all_heads_attention_sink.pdf}
    \vspace{-2em}
  \end{center}
}

\end{document}